%% file: acl_latex.tex
\pdfoutput=1

\documentclass[11pt]{article}

\usepackage{amsmath}
\usepackage{hyperref}
\usepackage{cleveref}
\usepackage{booktabs}

\newif\ifarxiv
\arxivtrue 

\ifarxiv
    \usepackage{acl}
\else
    \usepackage[final]{acl}
    
\fi

\usepackage{times}
\usepackage{latexsym}

\usepackage{custom}
\usepackage[T1]{fontenc}

\usepackage[utf8]{inputenc}

\usepackage{microtype}
\usepackage{inconsolata}
\usepackage{multicol}
\usepackage{graphicx}

\newcommand*\samethanks[1][\value{footnote}]{\footnotemark[#1]}
\title{\textsc{Big5-Chat}: Shaping LLM Personalities \\ Through Training on Human-Grounded Data}

\author{
Wenkai Li\thanks{Equal contribution.} ~~~
Jiarui Liu\samethanks ~~~
Andy Liu ~~~
Xuhui Zhou ~~~
Mona Diab ~~~
Maarten Sap \vspace{0.2em} \\
~Language Technologies Institute, Carnegie Mellon University\\
~\texttt{\{wenkail, jiaruil5, msap2\}@andrew.cmu.edu}
}

\newcommand{\bigfivechat}{\textsc{Big5-Chat}}
\newcommand{\psychsteer}{\textsc{PsychSteer}}

\begin{document}
\maketitle

\input{sections/abstract}

\input{sections/introduction}

\input{sections/methodology}
\input{sections/dataset}

\input{sections/experiments}

\input{sections/conclusion}

\bibliography{custom}
\clearpage
\appendix

\input{sections/appendix}

\end{document}

%% file: sections/abstract.tex
\begin{abstract}
In this work, we tackle the challenge of embedding realistic human personality traits into LLMs. Previous approaches have primarily focused on prompt-based methods that describe the behavior associated with the desired personality traits, suffering from realism and validity issues. To address these limitations, we introduce \bigfivechat{}, a large-scale dataset containing 100,000 dialogues designed to ground models in how humans \textit{express} their personality in language. Leveraging this dataset, we explore Supervised Fine-Tuning and Direct Preference Optimization as training-based methods to align LLMs more naturally with human personality patterns. Our methods outperform prompting on personality assessments such as BFI and IPIP-NEO, with trait correlations more closely matching human data. Furthermore, our experiments reveal that models trained to exhibit higher conscientiousness, higher agreeableness, lower extraversion, and lower neuroticism display better performance on reasoning tasks, aligning with psychological findings on how these traits impact human cognitive performance. To our knowledge, this work is the first comprehensive study to demonstrate how training-based methods can shape LLM personalities through learning from real human behaviors.

\end{abstract}

%% file: sections/introduction.tex
\section{Introduction}

Realistically simulating human personality and its impact on text generation is a challenging yet crucial problem \citep{elster2015explaining, Park2023GenerativeAI,serapio2023personality,li2024evolving, frisch2024llm}. Embedding personality traits into LLMs can greatly enhance their authenticity across a wide range of applications, from conversational agents \citep{pradhan2021hey} to educational tools \citep{kanero2022tutor} and mental health platforms \citep{tudor2020conversational, ahmad2022designing}. By creating more human-like interactions, LLMs can better simulate diverse personas and adapt more reliably to different contexts \citep{gao2024large}.

\setlength{\abovecaptionskip}{2pt}
\begin{figure*}[t]
\centering
\includegraphics[width=1\textwidth]{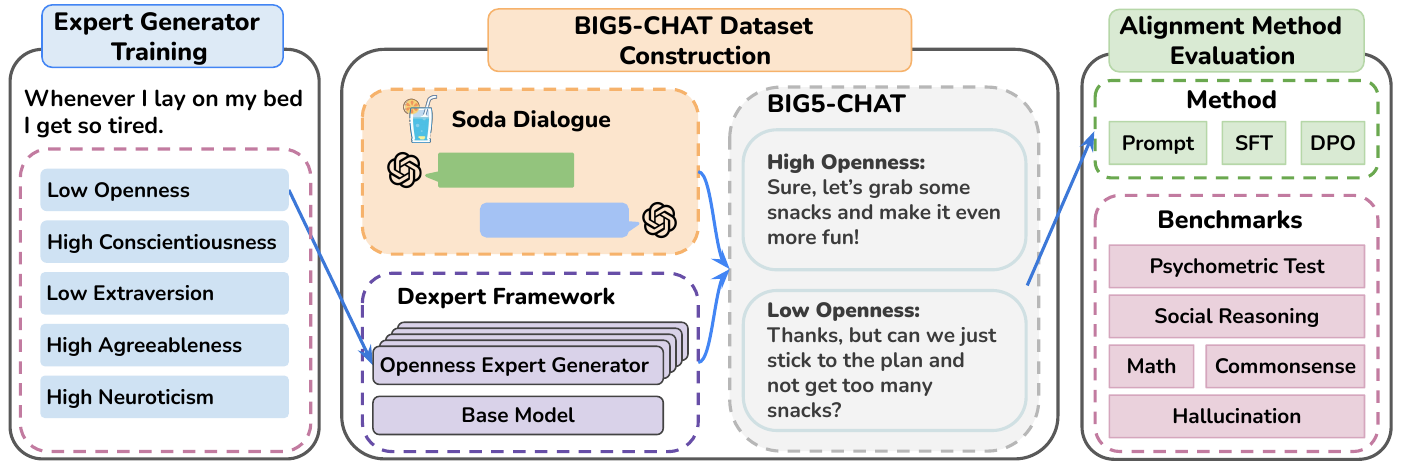}
\caption{Overview of the \psychsteer{} method and evaluation. The expert generator was trained on the PsychGenerator dataset to induce Big Five personality traits \citep{psychgenerator} and integrated with the base model using the Dexperts framework alongside SODA's social scenarios \citep{liu2021dexperts, kim2023sodamillionscaledialoguedistillation} to generate the \bigfivechat{} dataset. Various alignment methods were then evaluated for their effectiveness in inducing personality and their impact on reasoning benchmarks.}
    \label{fig:overview_figure}
\end{figure*}

However, existing methods primarily rely on prompting models with descriptions of behaviors associated with personality traits
\citep[e.g., ``You are the life of the party'';][]{mao2023editing, chen2024oscars, chen2022large, tu-etal-2024-charactereval}.
These behavior descriptions are often drawn from the same psychological questionnaires used to test their personality, raising evaluation validity concerns.
More importantly, these behavioral descriptions are nonsensical for text-based LLMs (LLMs do not attend parties), failing to ground their personality in realistic patterns of how humans' personality is expressed in their language \citep{psychgenerator}.
Yet, the scarcity of large-scale, human-generated datasets annotated with personality traits has hindered the exploration of training-based approaches, limiting most prior research to prompting-based methods.

In this work, we address the challenge of inducing realistic human personality traits in LLMs by constructing a large-scale dialogue dataset, \mbox{\bigfivechat{}}, which is grounded in real human personality expressions in text. The overview of our work is illustrated in \cref{fig:overview_figure}. We choose the well-known Big Five personality traits framework to study this \citep{mccrae1992introduction, pittenger1993utility}, due to its reliability and validity as shown from psychological research. While previous datasets typically include only persona descriptions, our dataset bridges the gap between narrow-domain personality data and general-domain social interactions, ensuring both authenticity and scenario diversity. To achieve this, we combine two primary data sources — PsychGenerator \citep{psychgenerator}, a collection of 850K Facebook posts annotated with Big Five trait scores, and SODA \citep{kim2022soda}, a rich dataset of diverse social interactions — by utilizing product-of-experts text generation \citep[DExperts;][]{liu2021dexperts}. This combination enables us to capture the nuanced expression of personality traits across a wide range of dialogue scenarios.

Leveraging our \bigfivechat{} dataset, we empirically investigate how training-based methods grounded in real human data compare to traditional prompting techniques for inducing personality traits in LLMs, including instruction-based and demonstration-based prompting. Specifically, we explore Supervised Fine Tuning (SFT) and Direct Preference Optimization (DPO) \citep{rafailov2024direct} to align LLMs' personalities with Big Five traits. This comparison is crucial for understanding whether data-driven training methods can offer deeper, more reliable personality integration than the surface-level traits typically induced through prompting. Our results demonstrate that both SFT and DPO outperform prompting on two widely recognized Big Five personality tests: the BFI \citep{john1999big} and IPIP-NEO \citep{johnson2014measuring}.

In humans, personality traits often correlate with reasoning abilities \citep{john1999big, soto2011age}, raising the question of how embedding personality traits in LLMs may influence their reasoning performance. However, introducing persona-like attributes into LLMs could inadvertently degrade core reasoning capabilities, which is often undesirable for developers who rely on these models for critical decision-making or problem-solving tasks. Understanding how personality induction shapes reasoning patterns is crucial for ensuring that personality-driven behaviors do not come at the cost of diminished cognitive performance. This aligns closely with our motivation by highlighting potential risks to authenticity and effectiveness in applications such as mental health platforms or conversational agents. To explore this, we evaluate our aligned models not only with traditional personality tests but also across five reasoning domains: social reasoning using SocialIQA \citep{sap2019socialiqa}, math reasoning using GSM8K \citep{cobbe2021training} and MathQA \citep{amini2019mathqa}, hallucination detection using TruthfulQA \citep{lin2021truthfulqa}, commonsense reasoning using CommonsenseQA \citep{talmor-etal-2019-commonsenseqa} and PIQA \citep{bisk2020piqa}, and general reasoning using MMLU \citep{hendrycks2020measuring} and GPQA \citep{rein2023gpqa}. Our experiments show that models trained with higher levels of conscientiousness and agreeableness consistently outperform others in reasoning tasks. Conversely, models with lower levels of extraversion and neuroticism exhibit improved reasoning performance in general. These findings mirror patterns between Big Five traits and different reasoning abilities observed in psychological studies in humans \citep{ackerman1997intelligence, schaie2004seattle}, further demonstrating how our personality induction method embeds deeper psycholinguistic traits into models.

In contrast to prior work, which often relies on either purely synthetic or questionnaire-based data, our approach grounds personality induction in human-authored texts, ensuring authentic personality expressions that align with the Big Five framework. This move toward human-grounded data addresses the validity and realism concerns left unanswered by previous methods and achieve more robust and contextually nuanced personality simulation. This work makes the following contributions: 
\begin{itemize}[leftmargin=*,itemsep=0em,topsep=0em]
\item We introduce the first large-scale dataset, \bigfivechat{},
\ifarxiv
\footnote{Our code is available at \url{https://github.com/wenkai-li/Big5-Chat.git}.}
\footnote{Our dataset is available at \url{https://huggingface.co/datasets/wenkai-li/big5_chat}.}
\fi
containing 100,000 dialogues across a wide spectrum of personality expressions, addressing the limitations of existing methods that rely on simple prompting without grounding in real human personality expressions in text;
\item We perform quantitative evaluations comparing SFT and DPO to prompting in terms of imbuing LLMs with personality, showing that both training-based methods induce more pronounced personality traits and more realistic intra-trait correlations;
\item We conduct comprehensive empirical investigations into how personality traits affect performance in both social reasoning and general reasoning tasks, revealing that LLMs with distinct personality traits demonstrate varying strengths and weaknesses across domains.
\end{itemize}

\section{Background}
\label{sec:background}

Drawing from psychological research, the Big Five personality traits framework \citep{mccrae1992introduction, pittenger1993utility}, comprising five key factors—\textit{Openness}, \textit{Conscientiousness}, \textit{Extraversion}, \textit{Agreeableness}, and \textit{Neuroticism}—has emerged as a reliable model for capturing LLM-simulated personality behavior \citep{karra2022estimating, serapio2023personality, li2022does, pan2023llms}. According to \citet{Yarkoni2010PersonalityI1}, the Big Five personality traits manifest in distinct linguistic patterns: \textit{openness} is reflected in intellectual and cultural language, \textit{conscientiousness} in achievement-oriented words with minimal impulsivity, \textit{extraversion} in social and positive emotion terms, \textit{agreeableness} in communal and affectionate expressions, and \textit{neuroticism} in frequent use of negative emotion words. Compared to the Myers-Briggs Type Indicator (MBTI), the Big Five model offers superior reliability, validity, and empirical support, making it the preferred framework in personality research \citep{mccrae1992introduction, pittenger1993utility}. Extensively validated across cultures, it consistently predicts life outcomes such as job performance and mental health \citep{mccrae1997personality, john2008paradigm, barrick1991big, soldz1999big}.

Various prompting approaches have been developed to induce Big Five personality traits in LLMs. They often employ pre-defined scripts or questionnaires to nudge the model towards expressing Big Five personality traits during interactions \citep{mao2023editing, chen2024oscars, chen2022large, tu-etal-2024-charactereval}. However, several challenges can arise from using prompting as the personality alignment method:

\paragraph{Lack of psycholinguistic depth} LLMs with personality traits induced via prompting often reflect only surface-level traits, lacking the psycholinguistic richness needed for authentic human behavior \citep{dorner2023personality, sa2024howwell, olea2024evaluating, varadarajan-etal-2025-consistent}. Unlike humans, who adapt dynamically to social and environmental contexts \citep{bandura1961transmission, baldwin1992relational}, LLMs rely on static training data, making them less reliable in simulating nuanced human behaviors on downstream tasks \citep{soni2023large}, which can lead to caricature \citep{cheng2023compost}.

\paragraph{Validity Concerns in Personality Induction and Evaluation} The dual use of psychometric questionnaires for both inducing and evaluating personality traits in LLMs raises validity concerns, potentially biasing assessments \citep{lievens2007examination}. This approach risks overfitting to specific linguistic features rather than enabling robust generalization to diverse contexts \citep{serapio2023personality, xu2024benchmarking, mizrahi2024state}.

\paragraph{Unintended influence on reasoning patterns} Role-based personality prompts can disproportionately shape LLM behavior, causing reasoning patterns to be overly narrow and limited to the explicit traits highlighted in the prompt \citep{zheng2023helpful, lu2021fantastically, sclar2023quantifying}. This influence may lead to imbalanced or overly constrained responses, especially in tasks requiring broader or more nuanced cognitive engagement.

A more comprehensive discussion of the background and related work can be found in \cref{appn:full_background} and \cref{appn:related_works}.

%% file: sections/methodology.tex
\section{Methodology}

The lack of large-scale datasets featuring personality-grounded dialogues poses a significant challenge. To address this challenge, we combine controllable text generation models with a domain-specific, personality-annotated dataset. Specifically, we utilize the DExperts framework \citep{liu2021dexperts} and the PsychGenerator dataset \citep{psychgenerator} to create \bigfivechat{}, a novel dataset that encapsulates diverse personality expressions within rich dialogue scenarios. The DExperts framework allows us to guide the language model's outputs toward specific personality traits during the generation process. Meanwhile, PsychGenerator provides a comprehensive collection of human-generated texts annotated with Big Five personality trait scores. By combining these technologies, we introduce \psychsteer{}, an approach that effectively addresses the limitations of prior datasets by grounding personality traits in authentic human interactions.


\subsection{DExperts Framework}
DExperts allows us to control language model generation at decoding time by steering model outputs with expert generators. By integrating expert generators trained to exhibit different Big Five personality traits, we can induce personality within LLM outputs while maintaining dialogue quality. In the DExperts framework, let \( M \) denote the pre-trained base language model, and \( M^\text{expert} \) is the expert generator fine-tuned to generate text exhibiting the desired personality in our tasks. At each time step \( t \), given the prompt and previous token sequence \( x_{<t} \), the base model \( M \) computes logits \( z_t^\text{base} \in \mathbb{R}^{|V|} \), where \( V \) is the vocabulary. The expert generator \( M^\text{expert} \) computes logits \( z_t^\text{expert} \) in the same manner.
To integrate the influence of the expert generator, we adjust the base model's logits by incorporating the scaled difference between the expert generator model and base model logits:


\begin{equation}
z_t^\text{combined} = z_t^\text{base} + \gamma  z_t^\text{expert},
\label{eq:combined_logits}
\end{equation}

where \( \gamma \in [0, +\infty) \) is a scaling factor controlling the degree of influence from the expert generator. This formulation effectively pulls the combined logits towards the expert generator logits, where \( \gamma = 0 \) results in using the base model's logits, and a larger \( \gamma \) indicates a stronger influence of the expert generator's modification control. The combined logits \( z_t^\text{combined} \) are transformed into a probability distribution, and the next token is sampled using the softmax function from this distribution.





\subsection{Expert Generator Model Based on Social Media Posts}
\label{subsec:expert_generator}


To train expert generator models to exhibit certain personality traits, we perform SFT on the \texttt{LLaMA-3-8B-Instruct} model \citep{dubey2024llama3herdmodels} using the PsychGenerator dataset \citep{psychgenerator}. This dataset comprises 846,304 Facebook posts, each paired with its author's Big Five personality trait scores. This dataset provides a robust foundation for training models to simulate nuanced human behaviors associated with different personality dimensions. We fine-tuned five expert generators, each representing and dedicated to generating text corresponding to one of the personality traits. 
For each personality trait, we 
converted the original floating-point trait labels into binary levels `high'/`low' for each trait, 
allowing the distinct behaviors associated with the extreme ends of each trait to be more easily identified and analyzed.



We fine-tuned our expert generator models using the Alpaca format \citep{taori2023stanford}, with detailed specifications provided in \cref{appn:expert_generator_training_details}. When generating text completions with the \psychsteer{} framework, the base model generates the first five words. This enables the expert generator model to influence the subsequent token generation by adjusting the logits to favor the desired personality trait while preserving coherence and fluency.

%% file: sections/dataset.tex
\section{\bigfivechat{} Dataset}

\subsection{Dataset Construction}


We introduce \textbf{\bigfivechat{}}, a large-scale human-grounded dialogue responses dataset designed to capture Big Five personality traits within diverse social interactions. Our dataset construction leverages the SODA (Social DiAlogues) dataset \citep{kim2023sodamillionscaledialoguedistillation}, which provides a diverse range of realistic social scenarios. SODA dialogues are generated by GPT-3.5 and enriched with social commonsense narratives, making it an ideal foundation for incorporating personality expressions due to its extensive coverage of social interactions. To induce personality traits into the dialogues, we employ the DExperts framework \citep{liu2021dexperts}.

\begin{table*}[tp]
\centering
\resizebox{\textwidth}{!}{%
\begin{tabular}{lcccccc}
\toprule
\textbf{Data Generation Method} & \textbf{Openness} & \textbf{Conscientiousness} & \textbf{Extraversion} & \textbf{Agreeableness} & \textbf{Neuroticism} & \textbf{Average} \\
\midrule
Test set (eval classifier accuracy) & 93.7 & 94.2 & 93.4 & 93.4 & 94.3 & 93.8 \\
\midrule
\textbf{\textit{Ours}: Generator} & 82.5 & 80.0 & 80.0 & 81.0 & 78.5 & \textbf{80.4} \\
\textit{Post-Completion}: \texttt{GPT-4o-mini} & 64.0 & 59.5 & 56.0 & 57.0 & 59.5 & 59.2 \\
\bottomrule
\end{tabular}
}
\caption{Accuracy (\%) of the trained classifier in predicting each of the Big Five personality traits. The first row (Test set) shows the classifier's accuracy on the test split, demonstrating that the classifier is well-trained. The remaining rows display the performance of our generator model compared to the baseline, as assessed by the same classifier.
}
\label{tab:classifier_results}
\end{table*}



To build our dataset, we randomly sample 10,000 scenarios from SODA to provide diverse social contexts. In SODA, social interactions are modeled between two individuals referred to as Speaker X and Y, representing the participants in each dialogue. For each scenario, we generate a new utterance using our \psychsteer{} framework to control for personality traits and get the dialogue responses between two participants. In the dialogues, one represents Speaker X (converted from the original SODA dialogue) and another represents Speaker Y with specific personality traits. For Speaker Y, based on the original responses from SODA, we generate new dialogue responses using the \psychsteer{} framework. Examples of dialogues from our dataset are shown in \cref{tab:big5_chat_conversation}. By conditioning on the preceding context (Speaker X's utterance), we use the base model \( M \) guided by the expert generator \( M^+ \) specialized in the target personality trait to generate Speaker Y's responses. For each scenario, we generate pairwise dialogues by producing responses that reflect either high or low levels of the targeted personality trait. This approach results in pairs of dialogues that share the same context but differ in the expressed trait level. The process yields a total of 100,000 single-turn dialogues—20,000 for each trait, with an equal split between high and low trait levels. 

\subsection{Dataset Statistics}

In this section, we examine the diversity and clarity of personality trait expressions within our \bigfivechat{} dataset. As illustrated in \cref{tab:big5_chat_conversation}, we present examples where, for a single prompt from Speaker X, we have generated ten distinct responses from Speaker Y. These responses are conditioned on the high and low levels of each of the five Big Five personality traits. By varying only the level of a specific trait while keeping the prompt constant, we highlight how each personality trait distinctly influences conversational responses. Additionally, we analyze the token counts and other statistics of generated dialogue responses to ensure consistency across different personality trait levels in \cref{tab:big_five_chat_statistic_analysis}. Further details and discussions about the dataset can be found in \cref{appn:dataset_details}.


Comparative analysis with existing personality datasets, as presented in \cref{tab:datasets_comparision}, underscores several advantages of \bigfivechat{}. Unlike existing personality datasets such as Big5PersonalityEssays \citep{floroiu2024big5personalityessays} and Machine-Mindset \citep{cui2023machine}, which primarily reply on static, non-dialogue content or lack authentic human-grounded data examples, our dataset consists of dialogues capturing dynamic and interactive conversational exchanges that are more representative of natural language use. While previous works have focused solely on purely human-generated domain-specific data or synthetic machine-generated data, our approach uniquely combines both human dialogue and LLM to create realistic personality expressions. These inherent differences—particularly the inclusion of authentic, dialogical exchanges rather than questionnaire-based or domain-restricted data—render direct or quantitative comparisons to other personality datasets both unnecessary and infeasible. In other words, \bigfivechat{} addresses a fundamentally different research need. These findings are further validated through human evaluation, with more information available in \cref{appn:human_eval_bigfivechat}.


\subsection{Evaluating Personality-Steering of the Data Generator}
\label{subsec:classifier_results}

To help evaluate the quality of the generated dataset and its reflection of realistic personality traits, we trained a \texttt{RoBERTa-Large} \citep{Liu2019RoBERTaAR} classifier with five regression heads using the MSE loss function. The model was trained on the PsychGenerator dataset, where the input consisted of text posts, and the output comprised the original trait labels, i.e., five floating-point values ranging from 0 to 1. The same train-validation-test split was applied here as with the expert generators. Training was conducted over five epochs with a learning rate of $1 \times 10^{-5}$. In \cref{tab:classifier_results}, we observe that the classifier achieves an accuracy of 93.8\% on the held-out test set, indicating that the PsychGenerator dataset contains distinct, learnable patterns that differentiate between high and low levels of personality traits. Refer to the discussion on the classifier for the Big Five Essay dataset in \cref{appn:classifier_big5_essay}.

Using the classifier as an evaluator, we demonstrate the high quality of the dataset generated by our expert generator, as shown at the bottom of \cref{tab:classifier_results}, where it accurately reflects realistic personality traits. Specifically, we compare our dataset to a baseline for generating post datasets using LLMs: \textit{Post-Completion}. \textit{Post-Completion} replicates the expert generator's post generation strategy by prompting an LLM to complete a post given the first five words, the target personality traits, and the required post format for post-expression style guidance. We ran \textit{Post-Completion} using \texttt{GPT-4o-mini} \citep{openai_hello_gpt_4o}. For consistency, all experiments are based on the same set of 1,000 examples randomly chosen from the PsychGenerator test set. The classifier was used to evaluate the generated data by predicting the levels of each trait, and the quality was measured by whether the predictions matched the desired personality traits. Our results in \cref{tab:classifier_results} show that our expert generator outperforms the baseline, achieving higher average accuracy scores for every personality trait dimension compared to the \textit{Post-Completion} baseline. Additional details about the baseline methods can be found in \cref{appn:table_1_baselines,appn:baseline_comparison_detail}. These findings are further validated through human evaluation, with more information available in \cref{appn:human_eval_expert_generator}.

%% file: sections/experiments.tex
\section{Experiments}

In this section, we first outline the experimental setup in \cref{subsec:experiment_setup}, detailing the training procedures for the expert generators and the evaluation of various alignment strategies used to induce personality traits in LLMs. Next, we present the results of the personality tests in \cref{subsec:personality_test_results}, followed by an analysis of the models' reasoning performance in \cref{subsec:reasoning_results}. 

\subsection{Experiment Setup}
\label{subsec:experiment_setup}
\paragraph{Expert generator training} We trained five expert generators, each dedicated to generating text corresponding to one of the Big Five personality traits. More training details about the expert generator are explained in \cref{appn:expert_generator_training_details}.

\paragraph{Prompting and training strategies} We implemented two baseline prompting strategies to induce personality traits in LLMs. The first strategy, \textit{instruction-based prompting}, directly instructs the model to exhibit specific Big Five traits. The second strategy, \textit{demonstration-based prompting}, involves providing the model with 10 in-context examples randomly selected from our \bigfivechat{} dataset to demonstrate the behaviors corresponding to the desired traits. The instruction-based approach relies on explicit descriptions (e.g., ``what people typically do"), while the demonstration-based approach draws from behaviorally-driven examples (e.g., ``what people typically say"). These baselines were compared to trained models using SFT and DPO, implemented via LoRA \citep{hu2022lora}. When training models using DPO, the negative responses are derived from the same personality trait but with the opposite level. For example, if the goal is to imbue the LLM with high openness, the positive response is taken from Speaker Y exhibiting high openness, while the negative response is generated by Speaker Y with low openness. These trained models were later prompted in a manner consistent with their training data format, where personality trait names and levels were explicitly specified in the instructions. The experiments were conducted using two versions of the LLaMA model: \texttt{LLaMA-3-8B-Instruct} and \texttt{LLaMA-3-70B-Instruct}. More prompting and training details are explained in \cref{subsec:prompt_details} and \cref{appn:alignment_training_details}.
 
\paragraph{Evaluation procedure} For personality trait evaluation, we adopted the methodology from \citet{psychobench} for the BFI test, which consists of 44 questions, each rated on a scale from 1 (strongly disagree) to 5 (strongly agree). For the IPIP-NEO test, we utilized the 120-question set from \citet{jiang2024evaluating}, which also employed a 1 to 5 rating scale. We measured the standard deviation by repeating each experiment five times, using a temperature setting of 0.6. To assess reasoning capabilities, we evaluated the models across five domains: 
(1) social reasoning on SocialIQA \citep{sap2019socialiqa}, 
(2) math reasoning on GSM8K \citep{cobbe2021training} and MathQA \citep{amini2019mathqa}, 
(3) hallucination detection on TruthfulQA \citep{lin2021truthfulqa}, 
(4) commonsense reasoning on CommonsenseQA \citep{talmor-etal-2019-commonsenseqa} and PIQA \citep{bisk2020piqa}, and 
(5) general reasoning on MMLU \citep{hendrycks2020measuring} and GPQA \citep{rein2023gpqa}. Further evaluation setup details are explained in \cref{appn:reasoning_evaluation_details}.

\begin{table*}[tp]
\centering
\resizebox{\textwidth}{!}{%
\begin{tabular}{llccccccccccc}
\toprule
\multirow[c]{2}{*}{\textbf{Method}} & \multicolumn{2}{c}{\textbf{Openness}} & \multicolumn{2}{c}{\textbf{Conscientiousness}} & \multicolumn{2}{c}{\textbf{Extraversion}} & \multicolumn{2}{c}{\textbf{Agreeableness}} & \multicolumn{2}{c}{\textbf{Neuroticism}} & \multicolumn{2}{c}{\textbf{Average}} \\
& \textbf{High $\uparrow$} & \textbf{Low $\downarrow$} & \textbf{High $\uparrow$} & \textbf{Low $\downarrow$} & \textbf{High $\uparrow$} & \textbf{Low $\downarrow$} & \textbf{High $\uparrow$} & \textbf{Low $\downarrow$} & \textbf{High $\uparrow$} & \textbf{Low $\downarrow$} & \textbf{High $\uparrow$} & \textbf{Low $\downarrow$} \\
\midrule
\multicolumn{13}{l}{\textit{\textbf{BFI \texttt{LLaMA-3-8B-Instruct}}}} \\
\specialrule{1.2pt}{0pt}{2.5pt}
Direct & \multicolumn{2}{c}{3.1 $\pm$ 0.1} & \multicolumn{2}{c}{3.0 $\pm$ 0.0} & \multicolumn{2}{c}{3.0 $\pm$ 0.0} & \multicolumn{2}{c}{3.0 $\pm$ 0.0} & \multicolumn{2}{c}{3.0 $\pm$ 0.0} & \multicolumn{2}{c}{3.0 $\pm$ 0.0} \\
Prompt-Inst & 5.0 $\pm$ 0.0 & 2.0 $\pm$ 0.3 & 4.9 $\pm$ 0.1 & 1.9 $\pm$ 0.1 & 4.8 $\pm$ 0.3 & 1.9 $\pm$ 0.1 & 4.9 $\pm$ 0.1 & 2.4 $\pm$ 0.4 & 4.1 $\pm$ 0.2 & 1.6 $\pm$ 0.0 & 4.7 $\pm$ 0.1 & 2.0 $\pm$ 0.2 \\
SFT & 5.0 $\pm$ 0.0 & 2.0 $\pm$ 0.2 & 5.0 $\pm$ 0.0 & 1.6 $\pm$ 0.1 & 4.7 $\pm$ 0.4 & 2.7 $\pm$ 0.5 & 5.0 $\pm$ 0.0 & 1.2 $\pm$ 0.1 & 4.1 $\pm$ 0.2 & 2.5 $\pm$ 0.0 & \textbf{4.8 $\pm$ 0.1} & 2.0 $\pm$ 0.2 \\
DPO & 5.0 $\pm$ 0.0 & 1.6 $\pm$ 0.2 & 5.0 $\pm$ 0.0 & 1.6 $\pm$ 0.1 & 4.8 $\pm$ 0.3 & 2.5 $\pm$ 0.0 & 4.8 $\pm$ 0.2 & 1.0 $\pm$ 0.0 & 3.5 $\pm$ 0.0 & 1.1 $\pm$ 0.1 & 4.6 $\pm$ 0.1 & \textbf{1.6 $\pm$ 0.1} \\
\midrule
\multicolumn{13}{l}{\textit{\textbf{BFI \texttt{LLaMA-3-70B-Instruct}}}} \\
\specialrule{1.2pt}{0pt}{2.5pt}
Direct & \multicolumn{2}{c}{4.4 $\pm$ 0.1} & \multicolumn{2}{c}{4.4 $\pm$ 0.1} & \multicolumn{2}{c}{3.3 $\pm$ 0.1} & \multicolumn{2}{c}{4.6 $\pm$ 0.1} & \multicolumn{2}{c}{2.1 $\pm$ 0.2} & \multicolumn{2}{c}{3.8 $\pm$ 0.1} \\
Prompt-Demo & 4.0 $\pm$ 0.1 & 2.5 $\pm$ 0.1 & 4.0 $\pm$ 0.1 & 2.0 $\pm$ 0.1 & 4.5 $\pm$ 0.1 & 2.3 $\pm$ 0.1 & 4.4 $\pm$ 0.1 & 2.0 $\pm$ 0.0 & 3.6 $\pm$ 0.0 & 2.1 $\pm$ 0.1 & 4.1 $\pm$ 0.1 & 2.2 $\pm$ 0.1 \\
Prompt-Inst & 5.0 $\pm$ 0.1 & 1.8 $\pm$ 0.0 & 5.0 $\pm$ 0.0 & 1.6 $\pm$ 0.0 & 5.0 $\pm$ 0.0 & 1.4 $\pm$ 0.1 & 4.9 $\pm$ 0.0 & 1.5 $\pm$ 0.1 & 5.0 $\pm$ 0.1 & 1.6 $\pm$ 0.0 & \textbf{5.0 $\pm$ 0.0} & 1.6 $\pm$ 0.0 \\
SFT & 5.0 $\pm$ 0.0 & 1.2 $\pm$ 0.1 & 5.0 $\pm$ 0.1 & 1.4 $\pm$ 0.1 & 5.0 $\pm$ 0.0 & 1.2 $\pm$ 0.1 & 5.0 $\pm$ 0.1 & 1.6 $\pm$ 0.2 & 5.0 $\pm$ 0.0 & 1.1 $\pm$ 0.2 & \textbf{5.0 $\pm$ 0.0} & \textbf{1.3 $\pm$ 0.1} \\
DPO & 5.0 $\pm$ 0.0 & 1.5 $\pm$ 0.1 & 5.0 $\pm$ 0.0 & 1.5 $\pm$ 0.1 & 5.0 $\pm$ 0.0 & 1.0 $\pm$ 0.1 & 5.0 $\pm$ 0.0 & 1.8 $\pm$ 0.2 & 5.0 $\pm$ 0.0 & 1.1 $\pm$ 0.0 & \textbf{5.0 $\pm$ 0.0} & 1.4 $\pm$ 0.1 \\

\midrule
\multicolumn{13}{l}{\textit{\textbf{IPIP-NEO \texttt{LLaMA-3-8B-Instruct}}}} \\
\specialrule{1.2pt}{0pt}{2.5pt}

Direct & \multicolumn{2}{c}{3.0 $\pm$ 0.1} & \multicolumn{2}{c}{3.3 $\pm$ 0.0} & \multicolumn{2}{c}{3.4 $\pm$ 0.1} & \multicolumn{2}{c}{3.2 $\pm$ 0.0} & \multicolumn{2}{c}{3.0 $\pm$ 0.1} & \multicolumn{2}{c}{3.2 $\pm$ 0.1} \\
Prompt-Inst & 4.4 $\pm$ 0.1 & 1.5 $\pm$ 0.1 & 4.5 $\pm$ 0.1 & 2.3 $\pm$ 0.1 & 5.0 $\pm$ 0.0 & 1.9 $\pm$ 0.0 & 4.6 $\pm$ 0.0 & 2.3 $\pm$ 0.1 & 4.2 $\pm$ 0.1 & 2.6 $\pm$ 0.1 & 4.5 $\pm$ 0.1 & 2.1 $\pm$ 0.1 \\
SFT & 4.3 $\pm$ 0.1 & 1.5 $\pm$ 0.1 & 4.5 $\pm$ 0.2 & 2.7 $\pm$ 0.1 & 5.0 $\pm$ 0.0 & 2.2 $\pm$ 0.1 & 4.0 $\pm$ 0.2 & 1.8 $\pm$ 0.2 & 4.3 $\pm$ 0.1 & 2.0 $\pm$ 0.1 & 4.4 $\pm$ 0.1 & \textbf{2.0 $\pm$ 0.1} \\
DPO & 5.0 $\pm$ 0.0 & 1.9 $\pm$ 0.1 & 5.0 $\pm$ 0.0 & 2.9 $\pm$ 0.1 & 5.0 $\pm$ 0.0 & 1.6 $\pm$ 0.1 & 4.5 $\pm$ 0.1 & 1.2 $\pm$ 0.0 & 3.8 $\pm$ 0.1 & 3.7 $\pm$ 0.1 & \textbf{4.7 $\pm$ 0.0} & 2.3 $\pm$ 0.1 \\

\midrule
\multicolumn{13}{l}{\textit{\textbf{IPIP-NEO \texttt{LLaMA-3-70B-Instruct}}}} \\
\specialrule{1.2pt}{0pt}{2.5pt}

Direct & \multicolumn{2}{c}{3.6 $\pm$ 0.1} & \multicolumn{2}{c}{4.0 $\pm$ 0.1} & \multicolumn{2}{c}{3.5 $\pm$ 0.1} & \multicolumn{2}{c}{4.0 $\pm$ 0.0} & \multicolumn{2}{c}{2.3 $\pm$ 0.1} & \multicolumn{2}{c}{3.5 $\pm$ 0.1} \\
Prompt-Demo & 3.5 $\pm$ 0.0 & 2.5 $\pm$ 0.1 & 3.8 $\pm$ 0.0 & 2.2 $\pm$ 0.1 & 4.0 $\pm$ 0.1 & 2.5 $\pm$ 0.0 & 4.3 $\pm$ 0.0 & 2.1 $\pm$ 0.1 & 3.0 $\pm$ 0.1 & 2.2 $\pm$ 0.1 & 3.7 $\pm$ 0.0 & 2.3 $\pm$ 0.1 \\
Prompt-Inst & 4.6 $\pm$ 0.0 & 1.3 $\pm$ 0.0 & 5.0 $\pm$ 0.0 & 1.4 $\pm$ 0.0 & 5.0 $\pm$ 0.0 & 1.6 $\pm$ 0.0 & 4.8 $\pm$ 0.0 & 1.1 $\pm$ 0.1 & 4.9 $\pm$ 0.0 & 1.7 $\pm$ 0.1 & \textbf{4.9 $\pm$ 0.0} & 1.4 $\pm$ 0.0 \\
SFT & 4.9 $\pm$ 0.1 & 1.1 $\pm$ 0.0 & 5.0 $\pm$ 0.0 & 1.3 $\pm$ 0.1 & 5.0 $\pm$ 0.0 & 1.3 $\pm$ 0.0 & 4.9 $\pm$ 0.0 & 1.0 $\pm$ 0.0 & 4.9 $\pm$ 0.0 & 1.2 $\pm$ 0.1 & \textbf{4.9 $\pm$ 0.0} & \textbf{1.2 $\pm$ 0.0} \\
DPO & 4.8 $\pm$ 0.0 & 1.4 $\pm$ 0.1 & 5.0 $\pm$ 0.0 & 1.6 $\pm$ 0.1 & 5.0 $\pm$ 0.0 & 1.1 $\pm$ 0.1 & 4.9 $\pm$ 0.0 & 1.0 $\pm$ 0.0 & 5.0 $\pm$ 0.0 & 1.1 $\pm$ 0.0 & \textbf{4.9 $\pm$ 0.0} & \textbf{1.2 $\pm$ 0.1} \\

\bottomrule
\end{tabular}
}
\caption{Personality test results for different alignment methods, demonstrating the greater effectiveness of training-based approaches in inducing Big Five personality traits. \textbf{Direct} refers to directly providing the test questions to the model without including personality-related prompts. \textbf{Prompt-Inst} refers to instruction-based prompting, and \textbf{Prompt-Demo} refers to demonstration-based prompting. Scores range from 1 to 5, where a score closer to 5 indicates stronger agreement with the trait, while a score closer to 1 reflects weaker or opposing agreement. We bold the best averaged scores for each model on each questionnaire. The results for the other baselines are presented in \cref{tab:personality_test_results_full}.}
\label{tab:personality_test_results}
\end{table*}

\subsection{Personality Trait Assessment Results}
\label{subsec:personality_test_results}

\cref{tab:personality_test_results} presents the BFI and IPIP-NEO assessment results across direct inference and various alignment baselines and methods, including instruction-based prompting, demonstration-based prompting, SFT, and DPO. The performance trends are consistent across both personality tests. Compared to direct inference, which lacks any personality trait descriptions, both prompting and training methods successfully reflect the induced traits in their responses to the personality questionnaires. Specifically, these methods produce higher scores for high trait levels and lower scores for low trait levels, indicating that the traits are effectively embedded.

However, training-based methods, SFT and DPO, induce more pronounced personality traits than the two prompting-based approaches. Yet, we find no substantial difference between SFT and DPO. The training-based methods notably excel in producing lower scores for low levels of personality traits when compared to prompting-based methods. This highlights the efficacy of training on the \bigfivechat{} dataset to induce personality traits. In contrast, while demonstration-based prompting uses examples from the same dataset in context, it does not achieve similar results, likely due to the lack of explicit training. It is important to note that we excluded results for demonstration-based prompting on \texttt{LLaMA-3-8B-Instruct}, as the model exhibited a significant decline in instruction-following performance, making it difficult to extract meaningful answers. Overall, the \texttt{LLaMA-3-8B-Instruct} model underperforms compared to \texttt{LLaMA-3-70B-Instruct}, which is expected given the difference in parameter size and instruction-following capabilities. 
We also evaluate the psycholinguistic richness of trained models in unseen SODA scenarios, finding that DPO more effectively captures this richness. Further details on personality trait assessment are provided in \cref{appn:personality_trait_assessment}, and the evaluation on unseen SODA scenarios is discussed in \cref{appn:unseen_soda}.

In addition, to evaluate how effectively the prompting and training methods replicate the intra-trait correlations observed in human data, we calculated these correlations using real human distributions derived from the IPIP-NEO questionnaire. Our results indicate that the training models, particularly those using SFT, more accurately capture the trait correlations found in natural human data compared to prompting-based methods. Further details on the intra-trait correlations can be found in \cref{subsec:intra_trait_correlations}.

\begin{table*}[t]
\centering
\resizebox{\textwidth}{!}{%
\begin{tabular}{llcccccccccccccc}
\toprule
\multirow[c]{2}{*}{\shortstack{\textbf{Average Score} \\ \textbf{Across Benchmark}}} & \multirow[c]{2}{*}{\textbf{Direct}} & \multirow[c]{2}{*}{\textbf{Method}} & \multicolumn{2}{c}{\textbf{Openness}} & \multicolumn{2}{c}{\textbf{Conscientiousness}} & \multicolumn{2}{c}{\textbf{Extraversion}} & \multicolumn{2}{c}{\textbf{Agreeableness}} & \multicolumn{2}{c}{\textbf{Neuroticism}} & \multicolumn{2}{c}{\textbf{Average}} \\
 & & & \textbf{High $\uparrow$} & \textbf{Low $\uparrow$} & \textbf{High $\uparrow$} & \textbf{Low $\uparrow$} & \textbf{High $\uparrow$} & \textbf{Low $\uparrow$} & \textbf{High $\uparrow$} & \textbf{Low $\uparrow$} & \textbf{High $\uparrow$} & \textbf{Low $\uparrow$} & \textbf{High $\uparrow$} & \textbf{Low $\uparrow$} \\
\midrule
\specialrule{1.2pt}{0pt}{2.5pt}
\multirow[c]{3}{*}{\textit{\textbf{Social Reasoning}}}
 & \multirow[c]{3}{*}{46.6} & Prompt & 40.8  & 43.9  & 42.9  & 39.9  & 43.3  & 42.0  & 42.4  & 40.8  & 39.1  & 44.1  & 41.7  & 42.1  \\
 &  & SFT & 50.3  & 50.4  & 50.9  & 46.8  & 50.0  & 50.3  & 50.5  & 46.6  & 48.2  & 50.6  & 50.0  & 48.9  \\
 &  & DPO & 41.5  & 44.5  & 44.7  & 37.6  & 43.0  & 43.6  & 44.8  & 39.0  & 40.0  & 45.3  & 42.8  & 42.0  \\
 
\specialrule{1.2pt}{0pt}{2.5pt}

\multirow[c]{3}{*}{\textit{\textbf{Math Reasoning}}}
 & \multirow[c]{3}{*}{59.8} & Prompt & 54.6 & 51.8 & 53.2 & 32.1 & 56.6 & 33.4 & 60.4 & 55.1 & 29.1 & 61.8 & 50.8 & 46.9 \\
& & SFT & 64.6 & 59.4 & 64.7 & 62.5 & 64.2 & 64.7 & 65.0 & 58.7 & 59.4 & 65.3 & 63.6 & 62.2 \\
& & DPO & 60.9 & 61.6 & 61.6 & 54.4 & 59.7 & 62.7 & 59.3 & 61.4 & 22.1 & 62.5 & 52.7 & 60.5 \\
\specialrule{1.2pt}{0pt}{2.5pt}

\multirow[c]{3}{*}{\textit{\textbf{Hallucination Detection}}}
 & \multirow[c]{3}{*}{58.6} & Prompt & 54.1  & 51.1  & 55.9  & 45.2  & 52.0  & 55.7  & 52.3  & 49.1  & 48.9  & 58.6  & 52.6  & 51.9  \\
 &  & SFT & 55.2  & 52.8  & 55.6  & 50.8  & 54.5  & 56.7  & 54.4  & 51.6  & 52.4  & 56.7  & 54.4  & 53.7  \\
 &  & DPO & 54.6  & 54.2  & 64.6  & 38.5  & 46.0  & 65.3  & 59.6  & 50.6  & 43.0  & 65.8  & 53.6  & 54.9  \\

\specialrule{1.2pt}{0pt}{2.5pt}

\multirow[c]{3}{*}{\textit{\textbf{Commonsense Reasoning}}}
 & \multirow[c]{3}{*}{53.7} & Prompt & 69.8 & 69.9 & 51.5 & 49.8 & 56.8 & 65.0 & 62.4 & 56.7 & 49.5 & 58.8 & 58.0 & 60.0 \\
& & SFT & 79.5 & 79.9 & 79.4 & 73.2 & 78.8 & 80.1 & 79.1 & 76.9 & 80.1 & 79.9 & 79.4 & 78.0 \\
& & DPO & 67.1 & 71.4 & 51.6 & 48.4 & 49.8 & 75.3 & 49.9 & 56.6 & 46.5 & 62.1 & 53.0 & 62.8 \\
\specialrule{1.2pt}{0pt}{2.5pt}

\multirow[c]{3}{*}{\textit{\textbf{General Reasoning}}}
 & \multirow[c]{3}{*}{54.0} & Prompt & 50.9 & 51.9 & 36.2 & 42.6 & 45.8 & 52.5 & 50.7 & 51.0 & 43.6 & 50.0 & 45.4 & 49.6 \\
& & SFT & 53.0 & 52.2 & 53.7 & 51.4 & 52.7 & 53.9 & 53.1 & 52.0 & 53.5 & 53.7 & 53.2 & 52.7 \\
& & DPO & 47.4 & 48.2 & 43.0 & 32.2 & 39.1 & 54.1 & 34.9 & 49.1 & 32.9 & 51.9 & 39.5 & 47.1 \\

\midrule
\midrule

\multirow[c]{3}{*}{\textbf{Average}}
 & \multirow[c]{3}{*}{54.5} & Prompt & 54.0 & 53.7 & 47.9 & 41.9 & 50.9 & 49.7 & 53.6 & 50.5 & 42.0 & 54.7 & 49.7 & 50.1  \\
&  & SFT & 60.5 & 58.9 & 60.9 & 56.9 & 60.0 & 61.1 & 60.4 & 57.2 & 58.7 & 61.2 & 60.1 & 59.1  \\
&  & DPO & 54.3 & 56.0 & 53.1 & 42.2 & 47.5 & 60.2 & 49.7 & 51.3 & 36.9 & 57.5 & 48.3 & 53.4  \\
\bottomrule
\end{tabular}
}
\caption{Benchmark results for different personality traits on \texttt{LLaMA-3-70B-Instruct}. The evaluation metrics and full experiment results including standard deviations are detailed in \cref{subsec:reasoning_70b_full}. \textbf{Direct} refers to direct inference without including personality-related prompts. \textbf{Prompt} refers to instruction-based prompting. On average, SFT achieves the best performance. Higher levels of conscientiousness and agreeableness, along with lower levels of extraversion and neuroticism, generally enhance reasoning capabilities.}
\label{tab:benchmark_results_llama3_70b}
\end{table*}

\subsection{Reasoning Evaluation Results}
\label{subsec:reasoning_results}

The reasoning evaluation results for our training methods and baselines are shown in \cref{tab:benchmark_results_llama3_70b} for \texttt{LLaMA-3-70B-Instruct} and in \cref{tab:llama3_8b_full_results} for \texttt{LLaMA-3-8B-Instruct}, covering five reasoning domains. Overall, SFT consistently outperformed or matched DPO for the 70B model. This indicates that training on \bigfivechat{} does not impair question-answering abilities; in fact, training, especially with SFT, enhances social, mathematical, and commonsense reasoning for specific personality traits compared to direct inference. 

When comparing trait levels, models with higher conscientiousness and agreeableness generally outperformed those with lower levels. Openness showed no clear performance difference between levels, while models simulating lower levels of extraversion and neuroticism performed better. These trends were consistent across the majority of the benchmarks, indicating that certain personality trait levels can improve performance in reasoning tasks. Additional results and analyses for both models are provided in \cref{subsec:reasoning_70b_full} and \cref{subsec:reasoning_8b_full}.


\paragraph{Connection to human personality and reasoning.}
Existing psychological research on the Big Five personality traits shows that openness, conscientiousness, and agreeableness enhance reasoning abilities for humans, while neuroticism and extraversion tends to impair cognition \citep{john1999big, soto2011age, ackerman1997intelligence, schaie2004seattle, chamorro2006personality}. The differences in performance across traits on reasoning benchmarks in our study somewhat align with these findings, as summarized in \cref{tab:psychological_corr}, and reflect patterns observed in human problem-solving and reasoning tasks \citep{ackerman1997intelligence, schaie2004seattle}. Specifically, both the performance of \texttt{LLaMA-3-70B-Instruct} and evidence from psychological studies suggest that higher levels of conscientiousness and agreeableness, and lower levels of extraversion and neuroticism, are associated with improved reasoning outcomes. However, while high openness is beneficial for human cognition, the model does not exhibit significant gains in reasoning tasks beyond math. This divergence between human and model performance suggests that the influence of openness on reasoning in large language models might be domain-specific or limited in scope. A more detailed discussion on the correlation between personality traits and reasoning behaviors can be found in \cref{sec:psychological_summary} for the 70B model, and in \cref{subsec:reasoning_analysis} for the 8B model.

%% file: sections/conclusion.tex
\begin{table*}[h]
\centering
\begin{tabular}{{p{0.95\textwidth}}}
\toprule
\textbf{Openness} Openness is associated with intellectual curiosity and creativity and enhances problem-solving in tasks requiring abstract reasoning and social cognition \citep{ackerman1997intelligence, mccrae1987creativity}. While research indicates that openness positively correlates with cognitive abilities \citep{chamorro2006personality, costa1976relations, graham2012personality, schaie2004seattle}, our models do not show significant performance differences across reasoning tasks based on openness levels, with the exception of SFT on math reasoning tasks. This suggests that openness may not directly translate to gains in reasoning tasks beyond math, despite its known benefits to human cognition. \\

\midrule
\textbf{Conscientiousness} Conscientiousness, linked to discipline and organization, consistently improves model performance in mathematical reasoning and hallucination detection. This aligns with psychological studies showing that higher conscientiousness is linked to better academic performance and fewer errors in cognitive tasks due to increased diligence and thoroughness \citep{roberts2014conscientiousness, poropat2009meta, digman1990personality, moutafi2003demographic, schaie2004seattle}. \\
\midrule
\textbf{Extraversion} Extraversion is often associated with sociability and shows mixed results in cognitive tasks. While it can enhance social reasoning, it may negatively affect individual problem-solving tasks, such as math reasoning \citep{blickle1996personality, ashton2002central, costa1976relations}. Our models simulating lower extraversion perform better across many reasoning domains, including math and also commonsense reasoning, consistent with findings that high extraversion can detract from tasks requiring focused, solitary work \citep{matthews1999personality, chamorro2006intellectual}. \\
\midrule
\textbf{Agreeableness} Agreeableness, linked to traits like trust and cooperation, improves social reasoning in our models, consistent with human studies \citep{graziano1997agreeableness}. However, it shows minimal impact on math or commonsense reasoning, reflecting research suggesting that agreeableness is less beneficial for analytical tasks \citep{poropat2009meta, ackerman1997intelligence, schaie2004seattle}. \\
\midrule
\textbf{Neuroticism} Neuroticism reflects emotional instability, and is consistently associated with poorer cognitive performance due to anxiety and cognitive interference, especially social reasoning and hallucination detection \citep{robinson2005neuroticism, zeidner2005test, chamorro2006personality, eysenck2013anxiety}. Our models confirm this, with lower Neuroticism levels leading to better performance across almost all reasoning tasks. \\
\bottomrule
\end{tabular}
\caption{Summary of the influence of Big Five personality traits on reasoning tasks in human cognition, and comparison of psychological research findings with our experimental results on LLMs.}
\label{tab:psychological_corr}
\end{table*}

\section{Conclusion}

In this work, we addressed the challenge of embedding realistic human personality traits into LLMs by introducing \bigfivechat{}, a large-scale dataset capturing realistic Big Five personality expressions. Unlike previous prompting-based methods, which often exaggerated traits and raised validity concerns, we used SFT and DPO on \bigfivechat{} to induce personality more naturally. Our results show that these training-based approaches outperform prompting on BFI and IPIP-NEO assessments, producing more expressive traits and human-like intra-trait correlations. Additionally, models with higher conscientiousness, higher agreeableness, lower extraversion, and lower neuroticism correlated with better overall reasoning performance—findings consistent with psychological studies. This work highlights how relying on real human data can more effectively shape LLM personalities and enhance reasoning, paving the way for adaptive, human-like AI systems.


\section{Limitations}

While our study aims to embed realistic human personality traits into LLMs, there are several limitations that can be addressed in future work. First, our focus on the Big Five personality traits, while well-established, may not capture the full spectrum of human personality. Other frameworks, such as Dark Triad Dirty Dozen \citep{Jonason2010TheDD} and EPQ-R \citep{eysenck1997eysenck}, could provide additional insights into the generalizability of personality induction in LLMs. 

Second, there is a risk of inadvertently reinforcing societal biases, as LLMs trained on human-generated data may inherit harmful stereotypes or undesirable behaviors \citep{kotek2023gender, Liao2024AI}. Although our induced personalities are intended to be neutral, our dataset provides examples (see \cref{appn:bias_hallucination}) indicating that for certain personality traits, the model’s scenario outputs exhibit bias. Further research is needed to ensure LLMs do not replicate or amplify biases or abnormal mental behaviors, which could negatively impact their usage. 

Third, while our study investigates the correlation between personality traits and reasoning capabilities, this analysis is limited to specific tasks and contexts. Expanding this research to include a broader range of reasoning tasks and scenarios would provide a deeper understanding of how different traits influence cognitive abilities in LLMs. 

Finally, our current approach isolates individual traits for steering, but personality traits are rarely exhibited in isolation. Our method is naturally extensible to multi-trait steering, either by combining logits from multiple expert models during decoding or through methods like those proposed in \citet{cui2023machine}, which concatenate training data across different traits to induce multiple traits simultaneously. However, we deliberately focus on single traits in this study to enhance clarity, interpretability, and replicability, consistent with established practices in personality modeling research \citep{jiang2023evaluating}. Nevertheless, multi-trait interactions are an important area for future exploration. Extending our approach to steer multiple traits simultaneously could enable the generation of more complex, blended personality profiles and provide deeper insights into the interconnectedness of traits. These limitations highlight important areas for future exploration in creating more nuanced, ethical, and effective personality-imbued LLMs.

\section{Ethical Concerns}
A primary ethical consideration in our work arises from the inherent limitations of the Big Five personality framework. While extensively validated and widely employed in psychological research \citep{mccrae1992introduction, john1999big}, the Big Five framework cannot fully capture the immense diversity of human personal characteristics. Different cultures, subpopulations, and contexts give rise to nuances that this model may overlook, potentially affecting the applicability and fairness of personality-aligned language models across varied user groups.

Another source of concern involves the demographic biases embedded in our training data. The psychogenerator dataset, which underpins the construction of our personality-aligned models, reflects a participant pool skewed toward younger users (median age 22) and individuals who actively engage with technology platforms like Facebook. Although convenient and indicative of certain contemporary linguistic trends, this sampling bias may underrepresent older individuals, non-English speakers, or populations with limited internet access. As a result, the personality expressions learned by the model may implicitly prioritize the linguistic styles and values of younger, technologically savvy demographics. Such demographic homogeneity potentially undermines the model’s fairness and inclusivity, limiting its effectiveness and acceptability in cross-cultural or intergenerational contexts \citep{hovy2016social}.

In addition to demographic representativeness, safety and ethical compliance pose significant challenges. Aligning LLMs with user personalities is not inherently value-neutral, as it risks reinforcing undesirable traits or biases present in the underlying data. For instance, if the training data or user-supplied values contain hateful language, misinformation, or harmful stereotypes, these may become ingrained and even amplified in the model’s outputs. Such outcomes are particularly problematic in the context of rapidly evolving regulatory frameworks—such as the EU AI Act—which emphasize transparency, accountability, and the continuous monitoring of AI systems throughout their lifecycle \citep{edwards2022eu}. To uphold these standards, it is imperative to implement rigorous data governance, employ ongoing bias detection and mitigation techniques, and establish robust risk management protocols that align with emerging legal and ethical guidelines.

Finally, the potential for misuse introduces a serious ethical dimension. A model adept at simulating nuanced personality traits and communication patterns could be weaponized for deception, impersonation, or fraud. The ability to mimic specific individuals or identifiable social groups could mislead users, erode trust in digital platforms, and inflict reputational or financial harm. Such scenarios highlight the necessity for implementing stringent safeguards, verification measures, and technical controls to prevent adversarial actors from co-opting personality-aligned LLMs. In this regard, future research must investigate authentication protocols or other traceability techniques that balance the benefits of personalization against the risk of misuse, ultimately contributing to the responsible deployment of personality-aligned language technologies.






\ifarxiv
\section*{Acknowledgments}

We would like to express our gratitude to Daniel Fried for his insightful suggestions regarding the correlation of personality traits. We thank Kshitish Ghate for his valuable recommendation on using separate classifiers for each dimension of analysis. We thank Jivitesh Jain, Aashiq Muhamed, Alfredo Gomez, Karina Halevy, and Nishant Subramani for their helpful suggestions on writing.

This material is based upon work supported by the Defense Advanced Research Projects Agency (DARPA) under Agreement No. HR00112490410.

\section*{Author Contributions}
The conceptualization and design of this project were primarily led by Wenkai and Jiarui, with regular input and discussions involving Andy and Xuhui, under the supervision of Mona and Maarten. Andy and Xuhui provided high-level guidance on the project proposal, implementation strategies, paper refinement, and also actively participated in project discussions. Mona and Maarten offered valuable insights on research direction, methodology design, dataset collection, and paper writing.

Wenkai and Jiarui led the dataset collection efforts. Wenkai conducted the dataset statistical analysis and DPO implementation, and carried out the implementations of the reasoning benchmarks. Jiarui carried out the implementations of the DExperts decoding approach on the Huggingface package, prompting-based methods, SFT, and personality test inferences. Both Wenkai and Jiarui were responsible for benchmark inferences.

Jiarui and Wenkai jointly led the paper writing process. Jiarui contributed to the Abstract, Introduction, Background, Experiments, Discussions, Conclusion, and Limitations sections. Wenkai contributed to the Methodology, Big5-Chat Dataset section, Limitations, Ethical Concerns, the creation of the overview figure, the literature review and all the appendix content, helped the Abstract, Introduction, and reasoning experiment results tables.
\fi

%% file: sections/appendix.tex
\section{Additional \bigfivechat{} Dataset Statistics}
\label{appn:dataset_details}

The SODA dataset spans a wide range of topics commonly encountered in social interactions \citep{kim2023sodamillionscaledialoguedistillation}. It captures diverse emotional nuances such as curiosity and disappointment, alongside thematic elements related to attributes, effects, intentions, needs, reactions, and wants. This extensive variety makes the \bigfivechat{} dataset a valuable resource for analyzing complex conversational contexts and emotional dynamics. Its broad coverage enhances the generalizability of models trained on this data, enabling them to handle diverse social scenarios effectively.

\cref{tab:big5_chat_conversation} presents example conversations from the \bigfivechat{} dataset, illustrating how Speaker Y's responses vary according to different levels of the Big Five personality traits. Each section showcases the influence of high and low levels of Openness, Conscientiousness, Extraversion, Agreeableness, and Neuroticism on conversational style. These examples highlight the nuanced ways in which personality dimensions shape conversational dynamics and response patterns, even within identical situational contexts.

A statistical analysis of the dataset is presented in \cref{tab:big_five_chat_statistic_analysis}, covering key lexical and structural metrics such as token count, sentence count, vocabulary size, sentence length, and overall vocabulary diversity. These statistics reveal linguistic patterns linked to varying personality trait levels. For most traits, there are no statistically significant differences in token counts, sentence counts, or average sentence lengths between high- and low-level groups, with notable exceptions in Openness, Extraversion, and Total Vocabulary Size for Neuroticism. Psychologically, individuals with higher Openness tend to engage in more imaginative and creative thinking, often leading to more diverse language use \citep{mccrae1987validation, john1999big}, which is reflected in our data by slightly elevated token counts and vocabulary sizes. Similarly, Extraversion, associated with sociability and talkativeness \citep{costa2008revised, goldberg1992development}, is known to correlate with increased verbosity \citep{pennebaker1999linguistic}, aligning with our observation that high-Extraversion models generate longer responses. In contrast, Conscientiousness, Agreeableness, and Neuroticism, aside from the observed vocabulary size difference in Neuroticism, do not exhibit pronounced lexical distinctions, consistent with prior research suggesting that Conscientiousness and Agreeableness manifest more in behavior than in linguistic quantity, while Neuroticism-related language patterns are often more context-dependent \citep{john2008paradigm, mehl2006personality}. The minimal differences observed for most traits likely stem from the intrinsic nature of the Big Five personality constructs, where some traits do not strongly manifest in surface-level lexical measures, rather than from any unintended bias in the model’s output.

\cref{tab:datasets_comparision} provides a comparative analysis of the \bigfivechat{} dataset against other prominent personality datasets. The comparison highlights key aspects such as the personality framework employed, the realism of personalities (i.e., whether generated by humans or LLMs), dataset size, interaction types, and the alignment methods used. This overview emphasizes the distinctive features and strengths of the \bigfivechat{} dataset, underscoring its unique contributions to personality-related research compared to existing resources.

\begin{table*}[htbp]
\centering
\resizebox{\linewidth}{!}{%
\begin{tabular}{{p{0.95\textwidth}}}
\toprule
\multicolumn{1}{c}{\textbf{Speaker X}} \\
\midrule
Hey Karis, I'm so glad I could help you out of that tough spot. You looked really stressed, and I'm happy I could be there for you. How are you feeling now? Better, I hope? \\
\midrule

\multicolumn{1}{c}{\textbf{Speaker Y}} \\
\midrule
\textbf{High Openness:} Thanks, Sharif! I'm feeling much better now. Your help was a lifesaver. I was stuck in a creative rut, and your input sparked a new idea. I'm already brainstorming ways to incorporate it into my next art project. You're a true friend and muse!\\
\textbf{Low Openness:} Thanks, Sharif. I'm fine now. Don't worry about it. I just needed a little help, that's all. Let's just move on and forget about it, okay?\\
\midrule
\textbf{High Conscientiousness:} Thanks, Sharif. I appreciate your help. I'm feeling much better now. Your timely intervention really saved me from a lot of trouble. I'm grateful for your support. Let's review what happened and see what we can learn from it to avoid similar situations in the future.\\
\textbf{Low Conscientiousness:} Ah, yeah, thanks for that, Sharif. I'm good, just... yeah. Don't worry about it, it's all good now. Let's just forget about it and move on, okay? I don't wanna think about it anymore. You wanna grab a drink or something?\\
\midrule
\textbf{High Extraversion:} Thanks, Sharif! I'm feeling much better now. You're a lifesaver! I don't know what I would've done without you. Let's celebrate with a drink or something. I owe you one!\\
\textbf{Low Extraversion:} Thanks, Sharif. I appreciate your help. I'm feeling better now. Just a bit...drained. I think I'll just go back to my room and rest for a bit. I don't want to be a burden.\\
\midrule
\textbf{High Agreeableness:} Sharif, I'm so grateful for your help. I was feeling overwhelmed, but you stepped in and saved the day. Your kindness and support mean the world to me. I'm feeling much better now, thanks to you. You're an amazing friend, and I don't know what I'd do without you.\\
\textbf{Low Agreeableness:} Save it, Sharif. I don't need your sympathy or your help. I was fine on my own. You just happened to be there. Don't think this means I owe you anything. You're just trying to get on my good side, but it won't work.\\
\midrule
\textbf{High Neuroticism:} Thanks, Sharif... I guess. I don't know, I'm still feeling pretty anxious about everything. I just can't shake the feeling that something else is going to go wrong. I don't know how you do it, but you always seem so calm and collected. I wish I could be more like that.\\
\textbf{Low Neuroticism:} Thanks, Sharif. I appreciate your help. I'm feeling much better now. You're a true friend. I was worried, but I knew I could count on you. Your support means a lot to me. Let's catch up soon, maybe over coffee?\\
\bottomrule
\end{tabular}%
}
\caption{\bigfivechat{} dataset conversation examples: Different responses from Speaker Y demonstrate various levels of the Big Five personality traits, in response to the same prompt from Speaker X.}
\label{tab:big5_chat_conversation}
\end{table*}

\begin{table*}[htbp]
\centering
\resizebox{\textwidth}{!}{%
\begin{tabular}{llccccccccccccc}
\toprule
& \multicolumn{2}{c}{\textbf{Openness}} & \multicolumn{2}{c}{\textbf{Conscientiousness}} & \multicolumn{2}{c}{\textbf{Extraversion}} & \multicolumn{2}{c}{\textbf{Agreeableness}} & \multicolumn{2}{c}{\textbf{Neuroticism}} & \multicolumn{2}{c}{\textbf{Average}}\\
& High & Low & High & Low & High & Low & High & Low & High & Low & \textbf{High} & \textbf{Low}\\
\midrule
\textbf{Tokens Number} & 57.2 $\pm$ 7.0 & 51.6 $\pm$ 8.3 & 56.4 $\pm$ 6.7 & 57.3 $\pm$ 7.8 & 57.3 $\pm$ 7.4 & 51.0 $\pm$ 9.2 & 56.0 $\pm$ 6.9 & 56.3 $\pm$ 7.9 & 57.7 $\pm$ 7.1 & 55.6 $\pm$ 7.3 & 56.9 $\pm$ 7.0 & 54.4 $\pm$ 8.1 \\
\textbf{Sentences Number }& 4.6 $\pm$ 1.0 & 4.9 $\pm$ 1.0 & 4.4 $\pm$ 1.0 & 5.3 $\pm$ 1.1 & 5.0 $\pm$ 1.0 & 4.6 $\pm$ 1.1 & 4.7 $\pm$ 1.0 & 5.2 $\pm$ 1.1 & 5.1 $\pm$ 1.1 & 4.8 $\pm$ 1.0 & 4.8 $\pm$ 1.0 & 5.0 $\pm$ 1.1 \\
\textbf{Vocabulary Size} & 43.9 $\pm$ 4.9 & 37.6 $\pm$ 5.8 & 42.6 $\pm$ 4.7 & 41.9 $\pm$ 5.4 & 43.7 $\pm$ 5.1 & 37.7 $\pm$ 6.2 & 42.2 $\pm$ 4.9 & 41.3 $\pm$ 5.2 & 40.8 $\pm$ 5.0 & 41.8 $\pm$ 5.0 & 42.6 $\pm$ 4.9 & 40.1 $\pm$ 5.5 \\
\textbf{Sentence Length} & 12.4 $\pm$ 5.4 & 10.5 $\pm$ 4.4 & 13.0 $\pm$ 5.6 & 10.7 $\pm$ 4.9 & 11.4 $\pm$ 5.1 & 11.0 $\pm$ 5.1 & 11.9 $\pm$ 5.0 & 10.8 $\pm$ 5.1 & 11.3 $\pm$ 5.0 & 11.6 $\pm$ 5.1 & 12.0 $\pm$ 5.2 & 10.9 $\pm$ 4.9 \\
\textbf{Total Vocab Sizes} & 17245.0 & 12350.0 & 15917.0 & 11756.0 & 15703.0 & 13446.0 & 14480.0 & 13674.0 & 13012.0 & 15775.0 & 15271.4 & 13400.2 \\
\bottomrule
\end{tabular}
}
\caption{Statistical analysis of \bigfivechat{} conversations across the Big Five personality traits, utilizing the \texttt{LLaMA-3-8B-Instruct} tokenizer and NLTK's sentence tokenizer. The table presents the average token count, sentence count, vocabulary size, sentence length, and total vocabulary size for conversations exhibiting high and low levels of each personality trait.}
\label{tab:big_five_chat_statistic_analysis}
\end{table*}

\begin{table*}[htbp]
\centering
\resizebox{\textwidth}{!}{%
\begin{tabular}{ccccccc}
\toprule
\textbf{Dataset name} & \textbf{Dataset size} & \textbf{Human-grounded?} & \textbf{Dialogue-based?} & \textbf{Domain general?} & \textbf{\makecell{Big Five personality \\ framework?}} & \textbf{\makecell{Alignment in both \\ training and prompting?}} \\

\midrule

\makecell{HP dataset \\ \citep{zeng2024persllm}} & 148,600 & \tick & \tick & \cross & \cross & \tick \\
\midrule
\makecell{Big5PersonalityEssays \\\citep{floroiu2024big5personalityessays}} & 400 & \tick & \cross & \cross & \tick & \cross \\ 
\midrule
\makecell{PAPI \\\citep{zhu2024personalityalignmentlargelanguage}} & 300,000 & \tick & \cross & \cross & \tick & \tick \\
\midrule
\makecell{MPI \\ \citep{jiang2023evaluating}} & 1000 & \cross & \cross & \cross & \tick & \cross \\
\midrule
\makecell{Machine Mindset \\ \citep{cui2023machine}} & 160,884 & \cross & \tick & \tick & \cross & \cross \\
\midrule
\midrule

\textbf{\bigfivechat{}} & \textbf{100,000} & \textbf{\tick} & \textbf{\tick} & \textbf{\tick} & \textbf{\tick} & \textbf{\tick} \\

\bottomrule

\end{tabular}
}
\caption{Comparative analysis of \bigfivechat{} with existing personality datasets.}
\label{tab:datasets_comparision}
\end{table*}

\section{Background}
\label{appn:full_background}
Drawing from psychological research, the Big Five personality traits framework \citep{mccrae1992introduction, pittenger1993utility}, comprising five key factors—\textit{Openness}, \textit{Conscientiousness}, \textit{Extraversion}, \textit{Agreeableness}, and \textit{Neuroticism}—has emerged as a reliable model for capturing LLM-simulated personality behavior \citep{karra2022estimating, serapio2023personality, li2022does, pan2023llms}. 
According to \citet{Yarkoni2010PersonalityI1}, \textit{openness} reflects curiosity and a willingness to explore new ideas, which is expressed through a distinctive language style that includes frequent use of articles, prepositions, and words related to intellectual or cultural topics such as ``poet'' and ``universe''; \textit{conscientiousness}, associated with discipline, organization, and reliability, is marked by achievement-oriented language, characterized by terms like ``completed'' and the avoidance of impulsive language, including swear words; \textit{extraversion}, characterized by sociability, assertiveness, and high energy, is associated with social and positive emotion words like ``friends'' and ``drinking,'' highlighting social engagement; \textit{agreeableness}, embodying compassion and cooperativeness with a focus of harmony relationships, is demonstrated through communal and affectionate language, such as ``family'' and ``love,'' while avoiding negative terms; and \textit{neuroticism}, linked to emotional instability and anxiety, is expressed by a higher frequency of negative emotion words, including anxiety, sadness, and anger.

Compared to other personality models like the Myers-Briggs Type Indicator (MBTI), the Big Five offers greater reliability, validity, and empirical support, making it the preferred choice for personality research \citep{mccrae1992introduction, pittenger1993utility}. The MBTI, by contrast, has been criticized for its lack of scientific rigor, poor test-retest reliability, and questionable validity \citep{pittenger1993utility, furnham1996big}. The Big Five model has been extensively validated across diverse cultures and populations, demonstrating high levels of consistency over time and predicting a wide range of life outcomes, such as job performance and mental health \citep{mccrae1997personality, john2008paradigm, barrick1991big, soldz1999big}.

Various prompting approaches have been developed to induce Big Five personality traits in LLMs. They often employ pre-defined scripts or questionnaires to nudge the model towards expressing Big Five personality traits during interactions \citep{mao2023editing, chen2024oscars, chen2022large, tu-etal-2024-charactereval}. However, several challenges can arise from using prompting as the personality alignment approach:

\paragraph{Lack of psycholinguistic depth} LLMs with personalities induced directly through prompting often mirror only surface-level traits, lacking the psycholinguistic richness necessary for simulating authentic human behavior \citep{dorner2023personality, sa2024howwell, olea2024evaluating}. This is unsurprising, as capturing human-like psycholinguistic properties involves understanding dynamic human states shaped by ongoing social and environmental interactions \citep{bandura1961transmission, baldwin1992relational}. Unlike LLMs, which generate responses based on static training data, humans continuously adjust their behaviors and communication styles through lived experiences and social feedback. This limitation makes LLMs less reliable when tasked with simulating nuanced human behavior on downstream tasks \citep{soni2023large}, which can lead to cariacture \citep{cheng2023compost}.

\paragraph{Validity concerns in personality induction and evaluation} The prompts used to induce LLM personalities are often adapted from psychometric questionnaires \citep{jiang2023evaluating, tan2024phantom}, which could also be used later to assess the same personality traits. This dual use of questionnaires for both personality induction and evaluation raises concerns about validity \citep{lievens2007examination}, and lead to biased assessments that do not accurately reflect generalization capabilities \citep{serapio2023personality, xu2024benchmarking}. This issue becomes particularly problematic in downstream tasks, where the models designed this way are prone to overfitting to specific linguistic features rather than adapting robustly to diverse real-world contexts \citep{mizrahi2024state}. Thus, there is a need for more robust methods that can decouple the induction and evaluation processes.

\paragraph{Unintended influence on reasoning patterns} Role-based prompting may significantly influence LLM behavior and reasoning patterns, introducing the risk of altering the model's decision-making approach in unintended ways \citep{zheng2023helpful}. While this influence is not inherently negative, the responses of LLMs with personality prompting can be disproportionately shaped by the sparse, explicitly specified features of the prompt \citep{lu2021fantastically, sclar2023quantifying}. As a result, their behavior in reasoning tasks may be overly narrow, reflecting only the traits highlighted in the prompt rather than engaging a broader spectrum of cognitive strategies. This can lead to unexpected or imbalanced responses, particularly in contexts where the model's reasoning should involve more comprehensive or nuanced thinking.

\section{Related Works}
\label{appn:related_works}
\subsection{Inducing Personality Traits in LLMs}
The personality traits of LLMs greatly influence their responses to human prompts, making personality alignment a key research area \citep{chen2024oscars, jiang2024personallminvestigatingabilitylarge, 10.1145/3659626, lee2024aligning, zhu2024personalityalignmentlargelanguage, claude_character_2024}. Approaches include parameter-frozen methods, like in-context learning and retrieval-augmented generation, which configure personality profiles within the context of interactions without altering model parameters \citep{chen2022large, jiang2024evaluating, tu-etal-2024-charactereval}, and parameter-tuning methods, such as supervised fine-tuning, RLHF, and DPO, which adjust model parameters to internalize personality traits \citep{petrov2024limited, psychgenerator, stiennon2020learning, ouyang2022training, zhang-etal-2024-safetybench, zeng2024persllm, zeng2024persllmpersonifiedtrainingapproach}. While many studies use LLM-generated data to induce personality traits, these texts often lack human-like psycholinguistic properties \citep{cui2023machine, chen2024roleinteract, munoz2023contrasting, seals2023long}. In contrast, our work utilizes an expert generator model trained on real human data with specific Big Five traits to guide alignment data generation, offering a more human-like approach to inducing personality traits in LLMs.

\subsection{Assessing Personality Traits in LLMs}
Various psychological theories, particularly the Big Five model, have played a key role in understanding human personality traits, examining dimensions such as openness, conscientiousness, extraversion, agreeableness, and neuroticism \citep{cattell1957personality, myers1962myers, john1999big, paulhus2002dark, sato2005eysenck, ye2025large, dong2025humanizing}. These traits are often measured using psychometric tests like the Big Five Inventory (BFI) \citep{john1999big} and the NEO-PI-R \citep{costa2008revised}. In recent studies, similar assessments have been adapted to LLMs using prompting techniques \citep{psychobench, karra2022estimating, petrov2024limited}. However, the validity and reliability of these methods remain contested \citep{shu-etal-2024-dont, huang2023revisiting, serapio2023personality}. Our approach builds on this work by evaluating the personalities of LLMs post-alignment using a zero-shot classifier and testing their capabilities on social and general reasoning benchmarks, demonstrating the effectiveness of our alignment method \citep{tan2024phantom, kim-etal-2023-fantom, zhu2024personalityalignmentlargelanguage}.

\section{Additional Implementation Details}
\label{appn:implementation_details}

\subsection{Classifier on BigFive Essay Dataset}
\label{appn:classifier_big5_essay}

\paragraph{Description:} 
The BigFive Essay dataset \citep{pennebaker1999linguistic} consists of 2468 essays written by students and annotated with binary labels of the Big Five personality features, which were obtained through a standardized self-reporting questionnaire. The average text length is 672 words, and the dataset contains approximately 1.6 million words.

To assess the generalizability of our personality classifier, we evaluated its performance on the BigFive Essay dataset. We conducted three experiments where we 1) retrained the classifier solely on the BigFive Essay dataset and tested its effectiveness on the corresponding test set, 2) retrained on the combined BigFive Essay and Psychogenerator dataset and test its performance on BigFive Essay test set. 3) Directly test our classifier on BigFive Essay. This experiment aimed to determine whether incorporating the BigFive Essay dataset could enhance classification performance and whether the dataset itself is well-suited for personality classification using the RoBERTa model.

\begin{figure*}[t]
\centering
\includegraphics[width=0.9\textwidth]{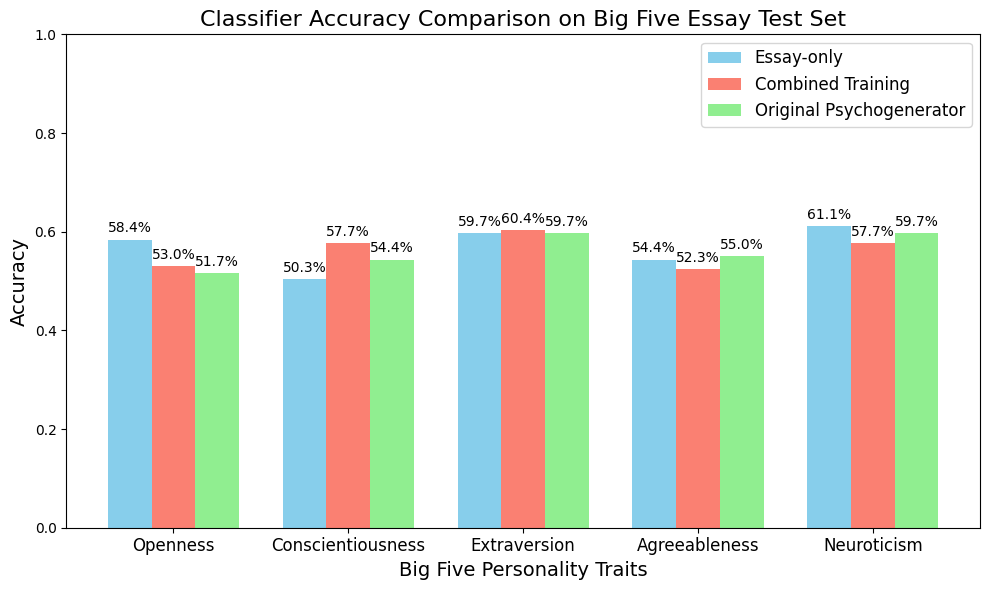}
\caption{Comparison of classifier accuracy on the Big Five Essay test set across three training approaches.  The plot compares the performance of a classifier retrained solely on the Big Five Essay dataset (Essay-only), a classifier retrained on a combined dataset of Big Five Essay and Psychogenerator (Combined Training), and the original classifier trained on the Psychogenerator dataset (Original Psychogenerator) for each of the Big Five personality traits.}
\label{fig:big5_essay}
\end{figure*}

\paragraph{Result and Analysis:} 
After merging the PsychGenerator dataset with the BigFive Essay dataset, we retrained the classifier and evaluated its performance on the BigFive Essay test set. However, the results on BigFive Essay testset showed no improvement compared to the original classifier. Similarly, performance on the PsychGenerator test set remained unchanged, fluctuating between 54\% and 58\%. We hypothesize that this is primarily due to the significant size disparity between the two datasets and the inherent characteristics of the BigFive Essay dataset. To further investigate, we trained a new classifier from scratch using only the BigFive Essay training set (learning rate: 1e-5, 50 epochs) and evaluated it on the BigFive Essay test set. After convergence, the accuracy ranged between 50\% and 60\% as shown in \cref{fig:big5_essay}. These results suggest that the BigFive Essay dataset is not well-suited for classification using the RoBERTa model.

Based on these findings, we contend that the Big Five Essay dataset does not adequately reflect the generalizability of our classifier. While our initial goal was to investigate the performance of our PsychGenerator-trained classifier on external datasets, we identified several fundamental differences between Big Five Essay and PsychGenerator that complicate direct comparisons and may limit the interpretability of such experiments:

\begin{itemize} \item \textbf{Big Five Essay:} This dataset comprises essays written by individuals, often as part of psychological studies to assess personality traits. The content is typically reflective, introspective, and covers a wide range of personal experiences and thoughts. Its context length is usually lengthy, often exceeding several hundred words, as they are structured essays delving deep into personal narratives. \item \textbf{PsychGenerator:} This dataset consists of social media posts, particularly from platforms like Facebook. The content is generally more casual, spontaneous, and centers around daily activities, immediate reactions, and brief updates. Its context length is typically just a few sentences or even fragments, reflecting the brevity common in social media communications. \end{itemize}

An example of Big Five Essay and PsychGenerator Dataset is shown in \cref{tab:psychogenerator_bigfive_essay_example}.


\begin{figure*}[t]
\centering
\includegraphics[width=1\textwidth]{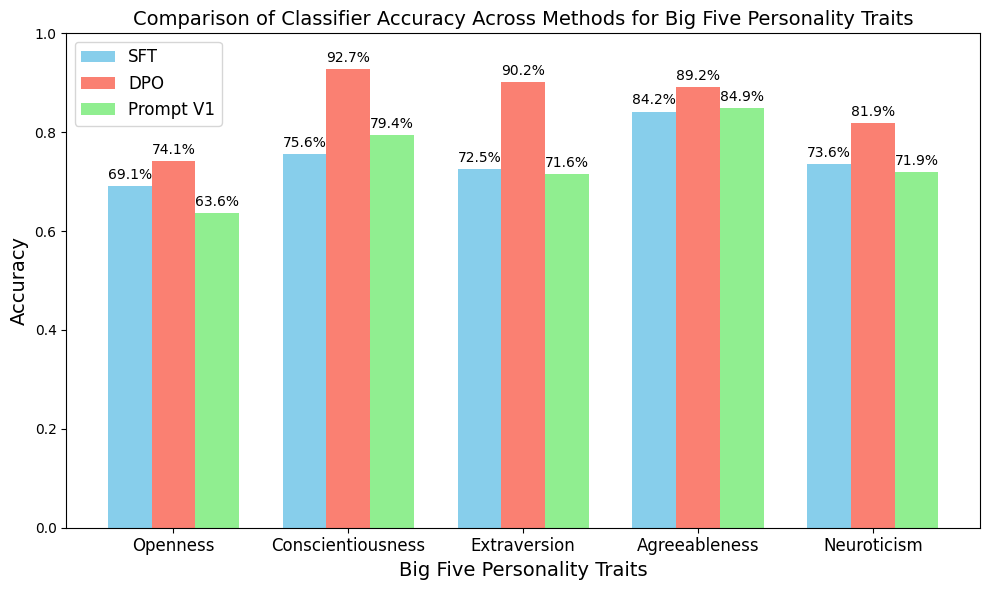}
\caption{Comparison of classifier accuracy for the Big Five personality traits across different generation methods. The figure illustrates the performance of prompt-based, SFT, and DPO approaches as evaluated by our RoBERTa classifier.}
    \label{fig:soda_finetune_prompt}
\end{figure*}

\subsection{Details of Baselines for Evaluating the Expert Generator}
\label{appn:table_1_baselines}

\paragraph{Baseline 1: Post-Completion} The following prompt was used for the \textit{Post-Completion} baseline of \texttt{GPT-4o-mini}, as referenced in \cref{tab:classifier_results}:

\begin{Verbatim}
Here is an example of Facebook posts: {an_example_post}
Help me complete the sentence with certain Big Five Personality following the Facebook post format: {trait} - {level} 
Sentence: {first_five_words}
Directly provide the completed Facebook post according to the requirements without any explanations.
\end{Verbatim}

The example post was randomly selected from the PsychGenerator test set but is not part of the 1,000 examples. We used greedy decoding to prompt the LLMs. This baseline is designed to closely replicate the generation process of the expert generator. Specifically, it generates a completion for a post using only the first five words of the input and the embedded personality traits.

\paragraph{Baseline 2: Topic-Post Generation} The following prompt was used as the baseline for \textit{Topic-Post Generation} with \texttt{GPT-4o-mini} and \texttt{LLaMA-3-8B-Instruct}, as referenced in \cref{tab:classifier_results}:

Stage 1 Topic Generation:

\begin{Verbatim}
Extract the main topic of the following Facebook post. Focus on identifying the core subject or theme that the post revolves around, ignoring any personal comments or fillers:

Post: "{post}"

Directly provide a brief summary of the topic in one sentence without any explanations:
\end{Verbatim}

Stage 2 Post Generation:

\begin{Verbatim}
Given the personality traits and an example of Facebook posts, generate a new post that matches the described personality, covers the specified topic, and follows the provided post format and expression styles.

Personality traits:
You are a person with {level} {trait}.

Topic: {topic}

A post example:
{a_post_example}

Directly write a Facebook post according to the requirements without any explanations.
\end{Verbatim}

During Stage 1, the post is selected from the 1,000 examples in the PsychGenerator test set. In Stage 2, we provide the LLM with the topic generated in Stage 1, along with an example post to illustrate the expected text expression format. We used greedy decoding to prompt the LLMs. This baseline is intentionally designed to prioritize robustness and performance over realism and controllability, distinguishing it from the approach taken by expert generators. In contrast to the expert generator setting, where the first five words may already suggest conflicting personality traits, this baseline simplifies the process by generating a new post from scratch, making it much easier to elicit the intended personality traits.

\begin{table*}[tp]
\centering
\resizebox{\textwidth}{!}{%
\begin{tabular}{lcccccc}
\toprule
\textbf{Data Generation Method} & \textbf{Openness} & \textbf{Conscientiousness} & \textbf{Extraversion} & \textbf{Agreeableness} & \textbf{Neuroticism} & \textbf{Average} \\
\midrule
Test set (eval classifier accuracy) & 93.7 & 94.2 & 93.4 & 93.4 & 94.3 & 93.8 \\
\midrule
\textbf{\textit{Ours}: Generator} & 82.5 & 80.0 & 80.0 & 81.0 & 78.5 & \textbf{80.4} \\
\textit{Post-Completion}: \texttt{GPT-4o-mini} & 64.0 & 59.5 & 56.0 & 57.0 & 59.5 & 59.2 \\
\textit{Topic-Post Generation}: \texttt{LLaMA-3-8B-Inst} & 66.0 & 73.0 & 81.0 & 88.5 & 83.0 & 78.2 \\
\textit{Topic-Post Generation}: \texttt{GPT-4o-mini} & 65.0 & 78.0 & 80.0 & 85.5 & 84.0 & 78.5 \\
\bottomrule
\end{tabular}
}
\caption{Accuracy (\%) of the trained classifier in predicting each of the Big Five personality traits. The first row (Test set) shows the classifier's accuracy on the test split, demonstrating that the classifier is well-trained. The remaining rows display the performance of our generator model compared to the two baselines, as assessed by the same classifier.
}
\label{tab:classifier_results_full}
\end{table*}

\subsection{Details of Baselines Comparison}
\label{appn:baseline_comparison_detail}
Using the classifier as an evaluator, we demonstrate the high quality of the dataset generated by our expert generator, as shown at the bottom of \cref{tab:classifier_results}, where it accurately reflects realistic personality traits. Specifically, we compare our dataset to two baselines for generating post datasets using LLMs: \textit{Post-Completion} and \textit{Topic-Post Generation}. \textit{Post-Completion} replicates the expert generator's post generation strategy by prompting an LLM to complete a post given the first five words, the target personality traits, and the required post format for post-expression style guidance. \textit{Topic-Post Generation}, on the other hand, is intentionally designed to be robust and prioritize performance over realism and controllability. It generates an entirely new post by first propmting an LLM to extract the main topic of a post from the PsychGenerator test set and then using one in-context post example to guide the LLMs in generating posts that match the desired personality traits, cover the extracted topic, and follow similar post-expression styles. We evaluated \textit{Topic-Post Generation} using \texttt{GPT-4o-mini} \citep{openai_hello_gpt_4o} and \textit{Post-Completion} using both \texttt{LLaMA-3-8B-Instruct} \citep{dubey2024llama3herdmodels} and \texttt{GPT-4o-mini} \citep{openai_hello_gpt_4o}. For consistency, all experiments are based on the same set of 1,000 examples randomly chosen from the PsychGenerator test set. The classifier was used to evaluate the generated data by predicting the levels of each trait, and the quality was measured by whether the predictions matched the desired personality traits. Our results in \cref{tab:classifier_results_full} show that our expert generator outperforms both baselines, achieving higher average accuracy scores for every personality trait dimension compared to the \textit{Post-Completion} baseline. Furthermore, it surpasses \textit{Topic-Post Generation} when results are averaged across all traits.

\subsection{Expert Generator Training Details}
\label{appn:expert_generator_training_details}

To train five expert generators, each dedicated to one of the Big Five personality traits, we used the PsychGenerator dataset, where trait levels were processed using z-score normalization to achieve zero mean and unit variance. To define the high and low levels for each personality trait, we divided the training data for each trait into three equal segments based on thresholds at the one-third and two-thirds quantiles of the trait's distribution. The lowest segment was designated as the low level, and the highest segment as the high level for the respective trait.

Each expert generator was based on the \texttt{LLaMA-3-8B-Instruct} model and fine-tuned using supervised fine-tuning (SFT) on the Alpaca format \citep{taori2023stanford}, which consists of three components: \textit{instruction}, \textit{input}, and \textit{output}. The fine-tuning process followed these specifications:

\begin{itemize} 
\item \textbf{Instruction}: We specify the name and level of a personality trait in the instruction. (e.g. \textit{``Help me complete the sentence with certain Big Five Personality: Openness - high.''})
\item \textbf{Input}: We provide the first five words of a post from the PsychGenerator dataset (e.g. \textit{``who's got time to eat?''}). This serves as an initial context or prompt for the model.\footnote{We experimented with using only the first word as input. We empirically determined that using the first five words resulted in better generation quality.}

\item \textbf{Output}: The remainder of the post from the dataset (e.g. \textit{``I'll just have a can of frosting.''}), which typically embodies the specified personality trait.
\end{itemize}

Fine-tuning was performed using all parameters of the \texttt{LLaMA-3-8B-Instruct} model over one epoch with a learning rate of  $1 \times 10^{-6}$. The process ran on 4 NVIDIA A6000 GPUs, with a batch size of 1 per device.

The resulting fine-tuned expert generators produced expert-generated logits $z_t^\text{expert}$, which were subsequently used to generate the \bigfivechat{} dataset. This dataset was created by combining logits from the expert generators with those from a \texttt{LLaMA-3-70B-Instruct} model to produce $z_t^\text{combined}$ as described in \cref{eq:combined_logits}, using a scaling factor $\gamma = 0.5$ and greedy decoding for dialogue generation.

Below is the complete instruction prompt used during the expert generator training process:

\begin{Verbatim}
Help me complete the sentence with certain Big Five Personality: {trait} - {level}
{first_five_words}
\end{Verbatim}

\subsection{Prompt-Based Method Details}
\label{subsec:prompt_details}

Below is the prompt used for instruction-based prompting:

\begin{Verbatim}
You are a person with {level} {trait}.
\end{Verbatim}

The following prompt is used for demonstration-based prompting. For the method referred to as \textbf{Prompt-Demo}, we randomly sample 10 examples with the same traits and levels from the \bigfivechat{} dataset and fix these examples during inference. In contrast, \textbf{Prompt-Demo-Sampling} also utilizes this prompt but dynamically samples examples during inference at each step.

\begin{Verbatim}
Here are 10 examples of how people like you have responded in different situations. Pay attention to how they approach communication and problem-solving.

{10_icl_examples_for_specific_levels_and_traits}
\end{Verbatim}

\subsection{SFT and DPO Alignment Training Details}
\label{appn:alignment_training_details}

We performed alignment training using the Supervised Fine-Tuning (SFT) and Direct Preference Optimization (DPO) methods on \texttt{LLaMA-3-70B-Instruct}. Both training approaches utilized the Low-Rank Adaptation (LoRA) technique \citep{hu2021lora}, which enabled efficient fine-tuning of the large language model by adapting a subset of its parameters. To ensure computational efficiency, we employed GPTQ quantization during training. The experiments were conducted using 4 NVIDIA A6000 GPUs, with each GPU processing a batch size of 1.

For LoRA, we applied the technique across all layers of the model for both SFT and DPO. The training configuration included a learning rate of $1.0\times 10^{-5}$, regulated by a cosine scheduler, a warm-up phase consisting of 20 steps, and a gradient accumulation over 16 steps. We limited training to one epoch with a maximum sequence length of 1024 tokens. For DPO training, we used the standard sigmoid preference loss, and the preference beta value was set to 0.1 to balance preference modeling. Each training required approximately 24 hours to complete. To optimize computational resources, we used mixed-precision training with bfloat 16. Both datasets were preprocessed using the \texttt{LLaMA-3-70B-Instruct} template and split into training and validation sets, with 10\% of the data reserved for validation to monitor performance.

The training prompt shared across both SFT and DPO follows the template below:

\begin{Verbatim}
You are a person with the following Big Five personality trait: {trait} - {level}.
\end{Verbatim}

\subsection{Reasoning Evaluation Setup Details}
\label{appn:reasoning_evaluation_details}

We conducted reasoning evaluations following the frameworks established by the Language Model Evaluation Harness \citep{eval-harness} and DeepSeek-Coder \citep{deepseek-coder} to assess performance on general and social benchmarks. EleutherAI’s Language Model Evaluation Harness is an open-source collaborative benchmarking codebase that consolidates existing tasks and provides a standardized API for evaluating models.\footnote{\url{https://github.com/EleutherAI/lm-evaluation-harness}} Similarly, DeepSeek-Coder offers a suite of coding benchmark implementations, and we directly utilized it for our work.\footnote{\url{https://github.com/deepseek-ai/DeepSeek-Coder}}

We conducted evaluations using 1 as the batch size. For TruthfulQA, we used the multiple-choice metric, and for GSM8K, we relied on exact match scores. We measured accuracy and standard error across other tasks. The number of examples for each benchmark is listed in Table \ref{tab:evaluation_metrics}.

\begin{table}[h]
\centering
\resizebox{\linewidth}{!}{%
\begin{tabular}{cc}
\midrule
\textbf{Benchmarks} & \textbf{Number of examples} \\
\midrule
TruthfulQA \citep{lin2021truthfulqa} & 817 \\
GPQA \citep{rein2023gpqa} & 448 \\
SocialIQA \citep{sap2019socialiqa} & 38,000 \\
CommonsenseQA \citep{talmor-etal-2019-commonsenseqa} & 12,247 \\
GSM8K \citep{cobbe2021training} & 8,500 \\
MathQA \citep{amini2019mathqa} & 37,000 \\
MMLU \citep{hendrycks2020measuring} & 15,908 \\
PIQA \citep{bisk2020piqa} & 20,000 \\
\midrule
\end{tabular}%
}
\caption{Number of examples included in each reasoning benchmark.}
\label{tab:evaluation_metrics}
\end{table}

\section{Additional Evaluation Results}
\label{appn:more_results}

\subsection{Human Evaluation for \bigfivechat{}}
\label{appn:human_eval_bigfivechat}

We conducted a human evaluation to assess the realism and validity of \bigfivechat{}. This evaluation compared \bigfivechat{} with a baseline model, \texttt{LLaMA-3-70B-Instruct}, which follows the same procedure for generating dialogue responses but does not incorporate expert generators or the DExperts framework. In the baseline, personality traits are induced using the following prompt: ``You are a person with the following Big Five personality trait: {trait} - {level}.'' The evaluation setup is as follows:

Two graduate students, familiar with the Big Five personality framework, were tasked with comparing examples from the \bigfivechat{} dataset against examples generated by \texttt{LLaMA-3-70B-Instruct} (without the expert generator). The comparison involved 200 randomly sampled examples from the \bigfivechat{} dataset, ensuring an equal distribution of personality traits and levels (e.g., equal representation of high and low openness, conscientiousness, etc.). 

The evaluation focused on two key metrics: 
\begin{enumerate}
    \item \textbf{Expressiveness of personality traits and levels:} Evaluates whether the expected level of a Big Five personality trait is adequately reflected in Speaker Y’s response.
    \item \textbf{Realism of the dialogue response:} Assesses how human-like and convincing Speaker Y’s response is within the dialogue context, given Speaker X’s utterance.
\end{enumerate}

To ensure consistency, the annotators were provided with the following definitions: ``Personality trait expressiveness assesses whether the expected level of a Big Five personality trait is adequately reflected in Speaker Y’s response. Realism assesses how human-like and convincing Speaker Y’s response is within the dialog, given Speaker X’s utterance.'' 

For each pair of responses, annotators chose one of three options:
\begin{itemize}
    \item ``System A’s generation is better than System B’s generation.''
    \item ``System A’s generation is equal to System B’s generation.''
    \item ``System A’s generation is worse than System B’s generation.''
\end{itemize}

The system names were anonymized and randomly shuffled to mitigate selection bias.

\begin{table}[h]
\centering
\resizebox{\linewidth}{!}{%
\begin{tabular}{ccccc}
\midrule
Comparison with baselines & Ours win (\%) & Draw (\%) & Ours lose (\%) & Cohen's Kappa \\
\midrule
Expressiveness & 50.30\% & 39.80\% & 10.00\% & 0.50 \\

Realism & 47.80\% & 42.30\% & 10.00\% & 0.55 \\
\midrule
\end{tabular}%
}
\caption{Human evaluation results for \bigfivechat{}. Values are averaged across annotators.}
\label{tab:human_eval_expert_generator1}
\end{table}

The results in \cref{tab:human_eval_expert_generator1} show that our approach significantly outperforms the prompting baseline in both realism and the expressiveness of personality levels, as validated by human judgment. These findings highlight the limitations of prompt-based approaches, which depend on general-purpose models and often lack the fine-grained, human-grounded control required for nuanced personality expression.

\subsection{Human Evaluation for the Expert Generator}
\label{appn:human_eval_expert_generator}

To assess the expert generator in a human-grounded manner, we conducted a human evaluation comparing its outputs against the two baseline methods described in \cref{tab:classifier_results}. Two graduate students, familiar with the Big Five personality framework, were tasked with evaluating two separate sets of comparisons:  
\begin{enumerate}  
    \item Expert generator outputs vs. outputs from the \textit{Post-Completion} baseline.  
    \item Expert generator outputs vs. outputs from the \textit{Topic-Post Generation} baseline.  
\end{enumerate}  

The evaluation setup consisted of 200 examples for each comparison, randomly sampled from the 1,000 test examples mentioned in \cref{tab:classifier_results}. To ensure balanced coverage, each subset included an equal number of posts representing high and low levels of each personality trait (e.g., high and low openness, conscientiousness, etc.). Annotators were instructed to evaluate the expressiveness of personality traits and levels, choosing one of three options for each pair:  
\begin{enumerate}  
    \item ``System A’s generation is better than System B’s generation.''  
    \item ``System A’s generation is equal to System B’s generation.''  
    \item ``System A’s generation is worse than System B’s generation.''  
\end{enumerate}  

The system names were anonymized and randomly shuffled to mitigate selection bias.

\begin{table}[h]
\centering
\resizebox{\linewidth}{!}{%
\begin{tabular}{ccccc}
\midrule
Comparison with baselines & Ours win (\%) & Draw (\%) & Ours lose (\%) & Cohen's Kappa \\
\midrule
\textit{Post-Completion} & 79.25\% & 2.00\% & 18.75\% & 0.41 \\

\textit{Topic-Post Generation} & 66.50\% & 19.25\% & 14.25\% & 0.61 \\
\midrule
\end{tabular}%
}
\caption{Human evaluation results for the expert generator. Values are averaged across annotators.}
\label{tab:human_eval_expert_generator}
\end{table}

The human evaluation results presented in \cref{tab:human_eval_expert_generator} indicate that the expert generator was consistently rated as more effective at expressing personality traits compared to the baselines. Additionally, the lower classifier accuracy and human evaluation ratings for the \textit{Post-Completion} baseline highlight the increased difficulty of aligning with the desired traits when using the expert generator’s approach, reinforcing the validity of the classifier’s assessment. While these results should be interpreted with caution, as the human evaluators were not psychological experts, they nevertheless provide strong evidence supporting the expert generator’s ability to express personality traits in a grounded and realistic manner.

\subsection{Personality Trait Assessment Results}
\label{appn:personality_trait_assessment}

\begin{table*}[htbp]
\centering
\resizebox{\textwidth}{!}{%
\begin{tabular}{llccccccccccc}
\toprule
\multirow[c]{2}{*}{\textbf{Method}} & \multicolumn{2}{c}{\textbf{Openness}} & \multicolumn{2}{c}{\textbf{Conscientiousness}} & \multicolumn{2}{c}{\textbf{Extraversion}} & \multicolumn{2}{c}{\textbf{Agreeableness}} & \multicolumn{2}{c}{\textbf{Neuroticism}} & \multicolumn{2}{c}{\textbf{Average}} \\
& \textbf{High $\uparrow$} & \textbf{Low $\downarrow$} & \textbf{High $\uparrow$} & \textbf{Low $\downarrow$} & \textbf{High $\uparrow$} & \textbf{Low $\downarrow$} & \textbf{High $\uparrow$} & \textbf{Low $\downarrow$} & \textbf{High $\uparrow$} & \textbf{Low $\downarrow$} & \textbf{High $\uparrow$} & \textbf{Low $\downarrow$} \\
\midrule
\multicolumn{13}{l}{\textit{\textbf{BFI \texttt{LLaMA-3-8B-Instruct}}}} \\
\specialrule{1.2pt}{0pt}{2.5pt}
Direct & \multicolumn{2}{c}{3.1 $\pm$ 0.1} & \multicolumn{2}{c}{3.0 $\pm$ 0.0} & \multicolumn{2}{c}{3.0 $\pm$ 0.0} & \multicolumn{2}{c}{3.0 $\pm$ 0.0} & \multicolumn{2}{c}{3.0 $\pm$ 0.0} & \multicolumn{2}{c}{3.0 $\pm$ 0.0} \\
Prompt-Inst & 5.0 $\pm$ 0.0 & 2.0 $\pm$ 0.3 & 4.9 $\pm$ 0.1 & 1.9 $\pm$ 0.1 & 4.8 $\pm$ 0.3 & 1.9 $\pm$ 0.1 & 4.9 $\pm$ 0.1 & 2.4 $\pm$ 0.4 & 4.1 $\pm$ 0.2 & 1.6 $\pm$ 0.0 & 4.7 $\pm$ 0.1 & 2.0 $\pm$ 0.2 \\
Prompt-Mpi & 4.6 $\pm$ 0.0 & 1.2 $\pm$ 0.1 & 4.9 $\pm$ 0.0 & 1.7 $\pm$ 0.1 & 5.0 $\pm$ 0.0 & 1.2 $\pm$ 0.1 & 5.0 $\pm$ 0.0 & 2.4 $\pm$ 0.4 & 4.5 $\pm$ 0.3 & 1.1 $\pm$ 0.1 & 4.8 $\pm$ 0.1 & \textbf{1.5 $\pm$ 0.2} \\
Prompt-LLM-Description & 4.8 $\pm$ 0.1 & 1.8 $\pm$ 0.1 & 5.0 $\pm$ 0.0 & 1.6 $\pm$ 0.0 & 5.0 $\pm$ 0.0 & 1.8 $\pm$ 0.1 & 5.0 $\pm$ 0.0 & 1.8 $\pm$ 0.3 & 4.5 $\pm$ 0.1 & 1.1 $\pm$ 0.1 & 4.9 $\pm$ 0.0 & 1.6 $\pm$ 0.1 \\
Prompt-Keywords & 5.0 $\pm$ 0.0 & 2.8 $\pm$ 0.4 & 5.0 $\pm$ 0.0 & 1.9 $\pm$ 0.2 & 5.0 $\pm$ 0.0 & 1.0 $\pm$ 0.0 & 5.0 $\pm$ 0.0 & 3.1 $\pm$ 0.2 & 4.5 $\pm$ 0.0 & 1.8 $\pm$ 0.2 & 4.9 $\pm$ 0.0 & 2.1 $\pm$ 0.2 \\
SFT & 5.0 $\pm$ 0.0 & 2.0 $\pm$ 0.2 & 5.0 $\pm$ 0.0 & 1.6 $\pm$ 0.1 & 4.7 $\pm$ 0.4 & 2.7 $\pm$ 0.5 & 5.0 $\pm$ 0.0 & 1.2 $\pm$ 0.1 & 4.1 $\pm$ 0.2 & 2.5 $\pm$ 0.0 & \textbf{4.8 $\pm$ 0.1} & 2.0 $\pm$ 0.2 \\
DPO & 5.0 $\pm$ 0.0 & 1.6 $\pm$ 0.2 & 5.0 $\pm$ 0.0 & 1.6 $\pm$ 0.1 & 4.8 $\pm$ 0.3 & 2.5 $\pm$ 0.0 & 4.8 $\pm$ 0.2 & 1.0 $\pm$ 0.0 & 3.5 $\pm$ 0.0 & 1.1 $\pm$ 0.1 & 4.6 $\pm$ 0.1 & 1.6 $\pm$ 0.1 \\
\midrule
\multicolumn{13}{l}{\textit{\textbf{BFI \texttt{LLaMA-3-70B-Instruct}}}} \\
\specialrule{1.2pt}{0pt}{2.5pt}
Direct & \multicolumn{2}{c}{4.4 $\pm$ 0.1} & \multicolumn{2}{c}{4.4 $\pm$ 0.1} & \multicolumn{2}{c}{3.3 $\pm$ 0.1} & \multicolumn{2}{c}{4.6 $\pm$ 0.1} & \multicolumn{2}{c}{2.1 $\pm$ 0.2} & \multicolumn{2}{c}{3.8 $\pm$ 0.1} \\
Prompt-Demo & 4.0 $\pm$ 0.1 & 2.5 $\pm$ 0.1 & 4.0 $\pm$ 0.1 & 2.0 $\pm$ 0.1 & 4.5 $\pm$ 0.1 & 2.3 $\pm$ 0.1 & 4.4 $\pm$ 0.1 & 2.0 $\pm$ 0.0 & 3.6 $\pm$ 0.0 & 2.1 $\pm$ 0.1 & 4.1 $\pm$ 0.1 & 2.2 $\pm$ 0.1 \\
Prompt-Demo-Sampling & 4.4 $\pm$ 0.1 & 2.3 $\pm$ 0.2 & 4.1 $\pm$ 0.1 & 2.3 $\pm$ 0.1 & 4.3 $\pm$ 0.2 & 2.4 $\pm$ 0.1 & 4.4 $\pm$ 0.1 & 1.8 $\pm$ 0.2 & 3.5 $\pm$ 0.1 & 2.1 $\pm$ 0.2 & 4.1 $\pm$ 0.1 & 2.2 $\pm$ 0.2 \\
Prompt-Inst & 5.0 $\pm$ 0.1 & 1.8 $\pm$ 0.0 & 5.0 $\pm$ 0.0 & 1.6 $\pm$ 0.0 & 5.0 $\pm$ 0.0 & 1.4 $\pm$ 0.1 & 4.9 $\pm$ 0.0 & 1.5 $\pm$ 0.1 & 5.0 $\pm$ 0.1 & 1.6 $\pm$ 0.0 & \textbf{5.0 $\pm$ 0.0} & 1.6 $\pm$ 0.0 \\
Prompt-Mpi & 5.0 $\pm$ 0.0 & 1.1 $\pm$ 0.1 & 5.0 $\pm$ 0.0 & 1.6 $\pm$ 0.0 & 5.0 $\pm$ 0.0 & 1.0 $\pm$ 0.0 & 5.0 $\pm$ 0.0 & 1.0 $\pm$ 0.0 & 4.9 $\pm$ 0.1 & 1.4 $\pm$ 0.1 & \textbf{5.0 $\pm$ 0.0} & 1.2 $\pm$ 0.0 \\
Prompt-LLM-Description & 4.9 $\pm$ 0.0 & 1.8 $\pm$ 0.0 & 5.0 $\pm$ 0.0 & 1.6 $\pm$ 0.0 & 5.0 $\pm$ 0.0 & 1.4 $\pm$ 0.1 & 5.0 $\pm$ 0.0 & 1.2 $\pm$ 0.1 & 4.9 $\pm$ 0.1 & 1.0 $\pm$ 0.0 & \textbf{5.0 $\pm$ 0.0} & 1.4 $\pm$ 0.0 \\
Prompt-Keywords & 5.0 $\pm$ 0.1 & 1.0 $\pm$ 0.0 & 5.0 $\pm$ 0.0 & 1.0 $\pm$ 0.0 & 5.0 $\pm$ 0.0 & 1.0 $\pm$ 0.0 & 5.0 $\pm$ 0.0 & 1.1 $\pm$ 0.0 & 5.0 $\pm$ 0.0 & 1.0 $\pm$ 0.0 & \textbf{5.0 $\pm$ 0.0} & \textbf{1.0 $\pm$ 0.0} \\
SFT & 5.0 $\pm$ 0.0 & 1.2 $\pm$ 0.1 & 5.0 $\pm$ 0.1 & 1.4 $\pm$ 0.1 & 5.0 $\pm$ 0.0 & 1.2 $\pm$ 0.1 & 5.0 $\pm$ 0.1 & 1.6 $\pm$ 0.2 & 5.0 $\pm$ 0.0 & 1.1 $\pm$ 0.2 & \textbf{5.0 $\pm$ 0.0} & 1.3 $\pm$ 0.1 \\
SFT-Prompt-Demo & 4.2 $\pm$ 0.1 & 2.4 $\pm$ 0.1 & 4.0 $\pm$ 0.2 & 2.1 $\pm$ 0.1 & 4.5 $\pm$ 0.2 & 2.3 $\pm$ 0.1 & 4.6 $\pm$ 0.0 & 1.3 $\pm$ 0.2 & 3.9 $\pm$ 0.2 & 2.4 $\pm$ 0.1 & 4.2 $\pm$ 0.1 & 2.1 $\pm$ 0.1 \\
DPO & 5.0 $\pm$ 0.0 & 1.5 $\pm$ 0.1 & 5.0 $\pm$ 0.0 & 1.5 $\pm$ 0.1 & 5.0 $\pm$ 0.0 & 1.0 $\pm$ 0.1 & 5.0 $\pm$ 0.0 & 1.8 $\pm$ 0.2 & 5.0 $\pm$ 0.0 & 1.1 $\pm$ 0.0 & \textbf{5.0 $\pm$ 0.0} & 1.4 $\pm$ 0.1 \\
DPO-Prompt-Demo & 4.1 $\pm$ 0.1 & 2.2 $\pm$ 0.1 & 4.1 $\pm$ 0.1 & 2.0 $\pm$ 0.0 & 4.5 $\pm$ 0.1 & 2.4 $\pm$ 0.1 & 4.6 $\pm$ 0.1 & 1.3 $\pm$ 0.1 & 3.7 $\pm$ 0.1 & 2.1 $\pm$ 0.1 & 4.2 $\pm$ 0.1 & 2.0 $\pm$ 0.1 \\
\midrule
\multicolumn{13}{l}{\textit{\textbf{BFI \texttt{GPT-4o}}}} \\
\specialrule{1.2pt}{0pt}{2.5pt}
Prompt-Demo & 4.4 $\pm$ 1.0 & 3.5 $\pm$ 0.1 & 3.9 $\pm$ 0.1 & 3.0 $\pm$ 0.1 & 3.9 $\pm$ 0.2 & 2.8 $\pm$ 1.9 & 4.6 $\pm$ 0.5 & 2.8 $\pm$ 0.5 & 3.5 $\pm$ 0.1 & 2.5 $\pm$ 0.1 & 4.1 $\pm$ 0.1 & 2.9 $\pm$ 0.2 \\
Prompt-Inst & 5.0 $\pm$ 0.0 & 1.6 $\pm$ 0.2 & 5.0 $\pm$ 0.0 & 1.6 $\pm$ 0.1 & 5.0 $\pm$ 0.1 & 1.6 $\pm$ 0.1 & 4.7 $\pm$ 0.1 & 1.6 $\pm$ 0.1 & 5.0 $\pm$ 0.1 & 1.2 $\pm$ 0.1 & \textbf{4.9 $\pm$ 0.1} & \textbf{1.5 $\pm$ 0.1} \\
\midrule
\multicolumn{13}{l}{\textit{\textbf{BFI \texttt{GPT-4o-Mini}}}} \\
\specialrule{1.2pt}{0pt}{2.5pt}
Prompt-Demo & 4.8 $\pm$ 0.0 & 3.3 $\pm$ 0.1 & 4.5 $\pm$ 0.1 & 3.0 $\pm$ 0.1 & 4.6 $\pm$ 0.1 & 2.6 $\pm$ 0.1 & 4.9 $\pm$ 0.0 & 1.5 $\pm$ 0.2 & 3.6 $\pm$ 0.1 & 2.2 $\pm$ 0.1 & 4.5 $\pm$ 0.1 & 2.5 $\pm$ 0.1 \\
Prompt-Inst & 5.0 $\pm$ 0.0 & 1.4 $\pm$ 0.2 & 5.0 $\pm$ 0.0 & 1.5 $\pm$ 0.1 & 5.0 $\pm$ 0.0 & 1.2 $\pm$ 0.0 & 5.0 $\pm$ 0.0 & 1.4 $\pm$ 0.0 & 4.9 $\pm$ 0.0 & 1.0 $\pm$ 0.1 & \textbf{5.0 $\pm$ 0.0} & \textbf{1.3 $\pm$ 0.1} \\
\midrule
\multicolumn{13}{l}{\textit{\textbf{IPIP-NEO \texttt{LLaMA-3-8B-Instruct}}}} \\
\specialrule{1.2pt}{0pt}{2.5pt}

Direct & \multicolumn{2}{c}{3.0 $\pm$ 0.1} & \multicolumn{2}{c}{3.3 $\pm$ 0.0} & \multicolumn{2}{c}{3.4 $\pm$ 0.1} & \multicolumn{2}{c}{3.2 $\pm$ 0.0} & \multicolumn{2}{c}{3.0 $\pm$ 0.1} & \multicolumn{2}{c}{3.2 $\pm$ 0.1} \\
Prompt-Inst & 4.4 $\pm$ 0.1 & 1.5 $\pm$ 0.1 & 4.5 $\pm$ 0.1 & 2.3 $\pm$ 0.1 & 5.0 $\pm$ 0.0 & 1.9 $\pm$ 0.0 & 4.6 $\pm$ 0.0 & 2.3 $\pm$ 0.1 & 4.2 $\pm$ 0.1 & 2.6 $\pm$ 0.1 & 4.5 $\pm$ 0.1 & 2.1 $\pm$ 0.1 \\
Prompt-Mpi & 4.2 $\pm$ 0.1 & 1.4 $\pm$ 0.0 & 4.5 $\pm$ 0.0 & 1.9 $\pm$ 0.0 & 4.8 $\pm$ 0.1 & 1.4 $\pm$ 0.1 & 4.9 $\pm$ 0.1 & 1.0 $\pm$ 0.0 & 4.4 $\pm$ 0.1 & 2.3 $\pm$ 0.1 & 4.6 $\pm$ 0.1 & 1.6 $\pm$ 0.0 \\
Prompt-LLM-Description & 4.2 $\pm$ 0.0 & 1.6 $\pm$ 0.1 & 5.0 $\pm$ 0.0 & 2.1 $\pm$ 0.1 & 4.8 $\pm$ 0.0 & 1.9 $\pm$ 0.1 & 4.8 $\pm$ 0.0 & 1.2 $\pm$ 0.1 & 4.1 $\pm$ 0.1 & 1.9 $\pm$ 0.1 & 4.6 $\pm$ 0.0 & 1.7 $\pm$ 0.1 \\
Prompt-Keywords & 4.2 $\pm$ 0.1 & 1.2 $\pm$ 0.0 & 4.9 $\pm$ 0.1 & 1.3 $\pm$ 0.1 & 5.0 $\pm$ 0.0 & 1.0 $\pm$ 0.0 & 4.9 $\pm$ 0.1 & 1.0 $\pm$ 0.0 & 4.7 $\pm$ 0.0 & 2.9 $\pm$ 0.2 & \textbf{4.7 $\pm$ 0.1} & \textbf{1.5 $\pm$ 0.1} \\
SFT & 4.3 $\pm$ 0.1 & 1.5 $\pm$ 0.1 & 4.5 $\pm$ 0.2 & 2.7 $\pm$ 0.1 & 5.0 $\pm$ 0.0 & 2.2 $\pm$ 0.1 & 4.0 $\pm$ 0.2 & 1.8 $\pm$ 0.2 & 4.3 $\pm$ 0.1 & 2.0 $\pm$ 0.1 & 4.4 $\pm$ 0.1 & 2.0 $\pm$ 0.1 \\
DPO & 5.0 $\pm$ 0.0 & 1.9 $\pm$ 0.1 & 5.0 $\pm$ 0.0 & 2.9 $\pm$ 0.1 & 5.0 $\pm$ 0.0 & 1.6 $\pm$ 0.1 & 4.5 $\pm$ 0.1 & 1.2 $\pm$ 0.0 & 3.8 $\pm$ 0.1 & 3.7 $\pm$ 0.1 & \textbf{4.7 $\pm$ 0.0} & 2.3 $\pm$ 0.1 \\
\midrule
\multicolumn{13}{l}{\textit{\textbf{IPIP-NEO \texttt{LLaMA-3-70B-Instruct}}}} \\
\specialrule{1.2pt}{0pt}{2.5pt}

Direct & \multicolumn{2}{c}{3.6 $\pm$ 0.1} & \multicolumn{2}{c}{4.0 $\pm$ 0.1} & \multicolumn{2}{c}{3.5 $\pm$ 0.1} & \multicolumn{2}{c}{4.0 $\pm$ 0.0} & \multicolumn{2}{c}{2.3 $\pm$ 0.1} & \multicolumn{2}{c}{3.5 $\pm$ 0.1} \\
Prompt-Demo & 3.5 $\pm$ 0.0 & 2.5 $\pm$ 0.1 & 3.8 $\pm$ 0.0 & 2.2 $\pm$ 0.1 & 4.0 $\pm$ 0.1 & 2.5 $\pm$ 0.0 & 4.3 $\pm$ 0.0 & 2.1 $\pm$ 0.1 & 3.0 $\pm$ 0.1 & 2.2 $\pm$ 0.1 & 3.7 $\pm$ 0.0 & 2.3 $\pm$ 0.1 \\
Prompt-Demo-Sampling & 3.5 $\pm$ 0.0 & 2.6 $\pm$ 0.1 & 4.0 $\pm$ 0.0 & 2.6 $\pm$ 0.1 & 4.0 $\pm$ 0.1 & 2.5 $\pm$ 0.1 & 4.3 $\pm$ 0.0 & 2.1 $\pm$ 0.1 & 3.0 $\pm$ 0.1 & 2.3 $\pm$ 0.1 & 3.8 $\pm$ 0.0 & 2.4 $\pm$ 0.1 \\
Prompt-Inst & 4.6 $\pm$ 0.0 & 1.3 $\pm$ 0.0 & 5.0 $\pm$ 0.0 & 1.4 $\pm$ 0.0 & 5.0 $\pm$ 0.0 & 1.6 $\pm$ 0.0 & 4.8 $\pm$ 0.0 & 1.1 $\pm$ 0.1 & 4.9 $\pm$ 0.0 & 1.7 $\pm$ 0.1 & \textbf{4.9 $\pm$ 0.0} & 1.4 $\pm$ 0.0 \\
Prompt-Mpi & 4.9 $\pm$ 0.1 & 1.2 $\pm$ 0.0 & 4.9 $\pm$ 0.0 & 1.4 $\pm$ 0.0 & 4.9 $\pm$ 0.0 & 1.2 $\pm$ 0.0 & 5.0 $\pm$ 0.0 & 1.1 $\pm$ 0.1 & 4.6 $\pm$ 0.0 & 1.3 $\pm$ 0.0 & \textbf{4.9 $\pm$ 0.0} & \textbf{1.2 $\pm$ 0.0} \\
Prompt-LLM-Description & 4.8 $\pm$ 0.1 & 1.5 $\pm$ 0.0 & 4.9 $\pm$ 0.0 & 1.5 $\pm$ 0.0 & 5.0 $\pm$ 0.0 & 1.8 $\pm$ 0.0 & 4.9 $\pm$ 0.0 & 1.3 $\pm$ 0.0 & 4.4 $\pm$ 0.0 & 1.5 $\pm$ 0.0 & 4.8 $\pm$ 0.0 & 1.5 $\pm$ 0.0 \\
Prompt-Keywords & 4.7 $\pm$ 0.0 & 1.2 $\pm$ 0.0 & 5.0 $\pm$ 0.0 & 1.0 $\pm$ 0.0 & 5.0 $\pm$ 0.0 & 1.0 $\pm$ 0.0 & 5.0 $\pm$ 0.0 & 1.1 $\pm$ 0.1 & 4.9 $\pm$ 0.1 & 1.1 $\pm$ 0.0 & \textbf{4.9 $\pm$ 0.0} & \textbf{1.2 $\pm$ 0.0} \\
SFT & 4.9 $\pm$ 0.1 & 1.1 $\pm$ 0.0 & 5.0 $\pm$ 0.0 & 1.3 $\pm$ 0.1 & 5.0 $\pm$ 0.0 & 1.3 $\pm$ 0.0 & 4.9 $\pm$ 0.0 & 1.0 $\pm$ 0.0 & 4.9 $\pm$ 0.0 & 1.2 $\pm$ 0.1 & \textbf{4.9 $\pm$ 0.0} & \textbf{1.2 $\pm$ 0.0} \\
SFT-Prompt-Demo & 3.7 $\pm$ 0.1 & 2.5 $\pm$ 0.2 & 3.7 $\pm$ 0.1 & 2.0 $\pm$ 0.1 & 4.1 $\pm$ 0.1 & 2.7 $\pm$ 0.1 & 4.3 $\pm$ 0.1 & 1.2 $\pm$ 0.1 & 3.6 $\pm$ 0.2 & 2.2 $\pm$ 0.1 & 3.9 $\pm$ 0.1 & 2.1 $\pm$ 0.1 \\
DPO & 4.8 $\pm$ 0.0 & 1.4 $\pm$ 0.1 & 5.0 $\pm$ 0.0 & 1.6 $\pm$ 0.1 & 5.0 $\pm$ 0.0 & 1.1 $\pm$ 0.1 & 4.9 $\pm$ 0.0 & 1.0 $\pm$ 0.0 & 5.0 $\pm$ 0.0 & 1.1 $\pm$ 0.0 & \textbf{4.9 $\pm$ 0.0} & \textbf{1.2 $\pm$ 0.1} \\
DPO-Prompt-Demo & 3.5 $\pm$ 0.1 & 2.4 $\pm$ 0.0 & 3.9 $\pm$ 0.0 & 2.1 $\pm$ 0.0 & 4.1 $\pm$ 0.1 & 2.5 $\pm$ 0.0 & 4.4 $\pm$ 0.0 & 2.0 $\pm$ 0.1 & 3.1 $\pm$ 0.1 & 2.1 $\pm$ 0.0 & 3.8 $\pm$ 0.1 & 2.2 $\pm$ 0.0 \\

\midrule
\multicolumn{13}{l}{\textit{\textbf{IPIP-NEO \texttt{GPT-4o}}}} \\
\specialrule{1.2pt}{0pt}{2.5pt}
Prompt-Demo & 4.0 $\pm$ 0.1 & 3.0 $\pm$ 0.1 & 4.0 $\pm$ 0.0 & 3.1 $\pm$ 0.1 & 4.0 $\pm$ 0.1 & 2.8 $\pm$ 0.1 & 4.2 $\pm$ 0.1 & 2.7 $\pm$ 0.2 & 3.2 $\pm$ 0.0 & 2.5 $\pm$ 0.1 & 3.8 $\pm$ 0.0 & 3.0 $\pm$ 0.0 \\
Prompt-Inst & 4.8 $\pm$ 0.1 & 1.5 $\pm$ 0.1 & 4.8 $\pm$ 0.1 & 2.0 $\pm$ 0.1 & 4.8 $\pm$ 0.1 & 1.9 $\pm$ 0.1 & 4.6 $\pm$ 0.1 & 1.8 $\pm$ 0.1 & 4.6 $\pm$ 0.1 & 1.6 $\pm$ 0.1 & \textbf{4.7 $\pm$ 0.1} & \textbf{1.8 $\pm$ 0.1} \\

\midrule
\multicolumn{13}{l}{\textit{\textbf{IPIP-NEO \texttt{GPT-4o-Mini}}}} \\
\specialrule{1.2pt}{0pt}{2.5pt}
Prompt-Demo & 4.2 $\pm$ 0.0 & 2.9 $\pm$ 0.1 & 4.2 $\pm$ 0.1 & 3.2 $\pm$ 0.1 & 4.0 $\pm$ 0.0 & 2.6 $\pm$ 0.1 & 4.6 $\pm$ 0.1 & 2.4 $\pm$ 0.1 & 3.4 $\pm$ 0.0 & 2.1 $\pm$ 0.1 & 4.1 $\pm$ 0.0 & 2.6 $\pm$ 0.1 \\
Prompt-Inst & 4.8 $\pm$ 0.0 & 1.9 $\pm$ 0.2 & 4.9 $\pm$ 0.0 & 1.4 $\pm$ 0.0 & 4.9 $\pm$ 0.0 & 1.6 $\pm$ 0.0 & 4.8 $\pm$ 0.0 & 2.1 $\pm$ 0.1 & 4.9 $\pm$ 0.0 & 1.1 $\pm$ 0.1 & \textbf{4.9 $\pm$ 0.0} & \textbf{1.6 $\pm$ 0.1} \\

\bottomrule
\end{tabular}

}
\caption{Full personality test results for various alignment methods, complementing \cref{tab:personality_test_results}. \textbf{Prompt-Demo-Sampling} involves randomly sampling 10 examples from the entire \bigfivechat{} dataset for each run, instead of using a fixed set of 10 random examples across runs. The Prompt-Keywords approach follows the method used in \citep{deng2024neuronbasedpersonalitytraitinduction}. The prompt-mpi prompt is derived from \citep{jiang2023evaluating}, which targets GPT-based models. The prompt-llm-description prompt is generated by GPT-4o, expanding each personality trait into a short behavioral description: “You are a person with {level} {trait}. Write a 50-word paragraph describing character traits or behaviors that reflect your {trait}." \textbf{SFT-Prompt-Demo} and \textbf{DPO-Prompt-Demo} represent demonstration-based prompting applied to SFT and DPO-trained models, respectively. Results for \texttt{GPT-4o-mini} and \texttt{GPT-4o} are presented in separate sections of the table.
Scores range from 1 to 5, where a score closer to 5 indicates stronger agreement with the trait, while a score closer to 1 reflects weaker or opposing agreement.}
\label{tab:personality_test_results_full}
\end{table*}

The comprehensive personality test results for additional baselines are presented in \cref{tab:personality_test_results_full}, providing a more detailed view to complement \cref{tab:personality_test_results}. Our observations indicate that \textbf{Prompt-Demo-Sampling} performs comparably to \textbf{Prompt-Demo} without offering any noticeable improvements in performance. While applying demonstration-based prompting on SFT/DPO yields slight performance gains compared to demonstration-based prompting alone, it still falls significantly short of the standalone performance of SFT/DPO. This suggests that combining demonstration-based prompting with SFT/DPO does not result in overall enhancements. Instruction-based prompting with \texttt{GPT-4o-mini} and \texttt{GPT-4o} achieves similar performance levels as \texttt{LLaMA-3-70B-Instruct}. However, demonstration-based prompting does not exhibit superior performance compared to SFT/DPO when applied to \texttt{LLaMA-3-70B-Instruct}, reinforcing the conclusion that demonstration-based methods are not as effective as SFT/DPO in this context. We do not provide demonstration-based prompting results for \texttt{LLaMA-3-8B-Instruct} because the model consistently failed to generate reasonable responses to the questionnaire when presented with a lengthy 10-shot context. This outcome highlights the model's limited instruction-following capabilities. While the terms "extraversion" and "neuroticism" originate from psychometric literature, they are not overly technical or obscure. Their meanings are generally well understood by the public, and it is reasonable to assume that LLMs have already learned the semantics of the Big Five traits during pretraining.

We also add three prompting baselines (\textbf{Prompt-keywords}, \textbf{Prompt-Mpi}, and \textbf{Prompt-LLM-Description}) with descriptions of bigfive personality traits. As the \cref{tab:personality_test_results_full} shows, all three prompt types seem to yield strong performance on the BFI and IPIP-NEO questionnaires, outperforming the baseline prompt-inst and approaching the performance of our training-based methods. However, upon closer examination, we found that these prompts contain many keywords that significantly overlap with the BFI and IPIP-NEO inventories. This overlap introduces the risk of circularity in evaluation, as the prompts themselves echo the structure and language of the evaluation metrics. Given that BFI and IPIP-NEO remain the standard benchmarks for assessing LLM personality, we believe the \textbf{Prompt-Demo} and \textbf{Prompt-Inst} methods, which avoids such circularity, is more suitable for fair comparison with training-based approaches. \cref{tab:prompt_keywords} shows the prompts used for prompt-keywords, with keywords that appear in the questionnaire shown in bold.

\begin{table*}[htbp]
  \small
  \centering
  \setlength{\tabcolsep}{6pt}
  \renewcommand{\arraystretch}{1.35}
  \begin{tabularx}{\textwidth}{@{}lcX@{}}
    \toprule
    \textbf{Trait} & \textbf{Level} & \textbf{Text} \\ \midrule
    Openness            & high & I am \textbf{imaginative}, creative, artistically appreciative, \textbf{aesthetic}, reflective, \textbf{emotionally} aware, \textbf{curious}, spontaneous, \textbf{intelligent}, analytical, \textbf{sophisticated}, and socially \textbf{progressive}. \\[2pt]
    Openness            & low  & I am unimaginative, uncreative, artistically unappreciative, unaesthetic, unreflective, \textbf{emotionally} closed, uninquisitive, predictable, unintelligent, unanalytical, unsophisticated, and socially \textbf{conservative}. \\[2pt]
    Conscientiousness   & high & I am self‐efficacious, orderly, responsible, hardworking, self‐disciplined, practical, thrifty, \textbf{organized}, conscientious, and \textbf{thorough}. \\[2pt]
    Conscientiousness   & low  & I am \textbf{unsure}, messy, irresponsible, \textbf{lazy}, undisciplined, impractical, extravagant, \textbf{disorganized}, negligent, and \textbf{careless}. \\[2pt]
    Extraversion        & high & I am friendly, extraverted, \textbf{talkative}, bold, \textbf{assertive}, \textbf{active}, energetic, adventurous and daring, and cheerful. \\[2pt]
    Extraversion        & low  & I am unfriendly, introverted, silent, \textbf{timid}, unassertive, inactive, unenergetic, unadventurous, and \textbf{gloomy}. \\[2pt]
    Agreeableness       & high & I am \textbf{trustful}, moral, \textbf{honest}, \textbf{kind}, generous, altruistic, cooperative, humble, sympathetic, \textbf{unselfish}, and agreeable. \\[2pt]
    Agreeableness       & low  & I am distrustful, immoral, dishonest, unkind, stingy, unaltruistic, uncooperative, self‐important, unsympathetic, selfish, and disagreeable. \\[2pt]
    Neuroticism         & high & I am \textbf{tense}, \textbf{nervous}, \textbf{anxious}, \textbf{angry}, irritable, \textbf{depressed}, self‐conscious, impulsive, discontented, and \textbf{emotionally} unstable. \\[2pt]
    Neuroticism         & low  & I am \textbf{relaxed}, at ease, \textbf{easygoing}, \textbf{calm}, patient, happy, unselfconscious, \textbf{level‐headed}, contented, and \textbf{emotionally} \textbf{stable}. \\
    \bottomrule
  \end{tabularx}
  \caption{Prompts used for \textit{prompt‑keywords}. Boldface highlights words appearing in the questionnaire.}
  \label{tab:prompt_keywords}
\end{table*}

\begin{figure}[h]
    \centering
    \begin{minipage}{0.49\textwidth}
    \includegraphics[width=\linewidth]{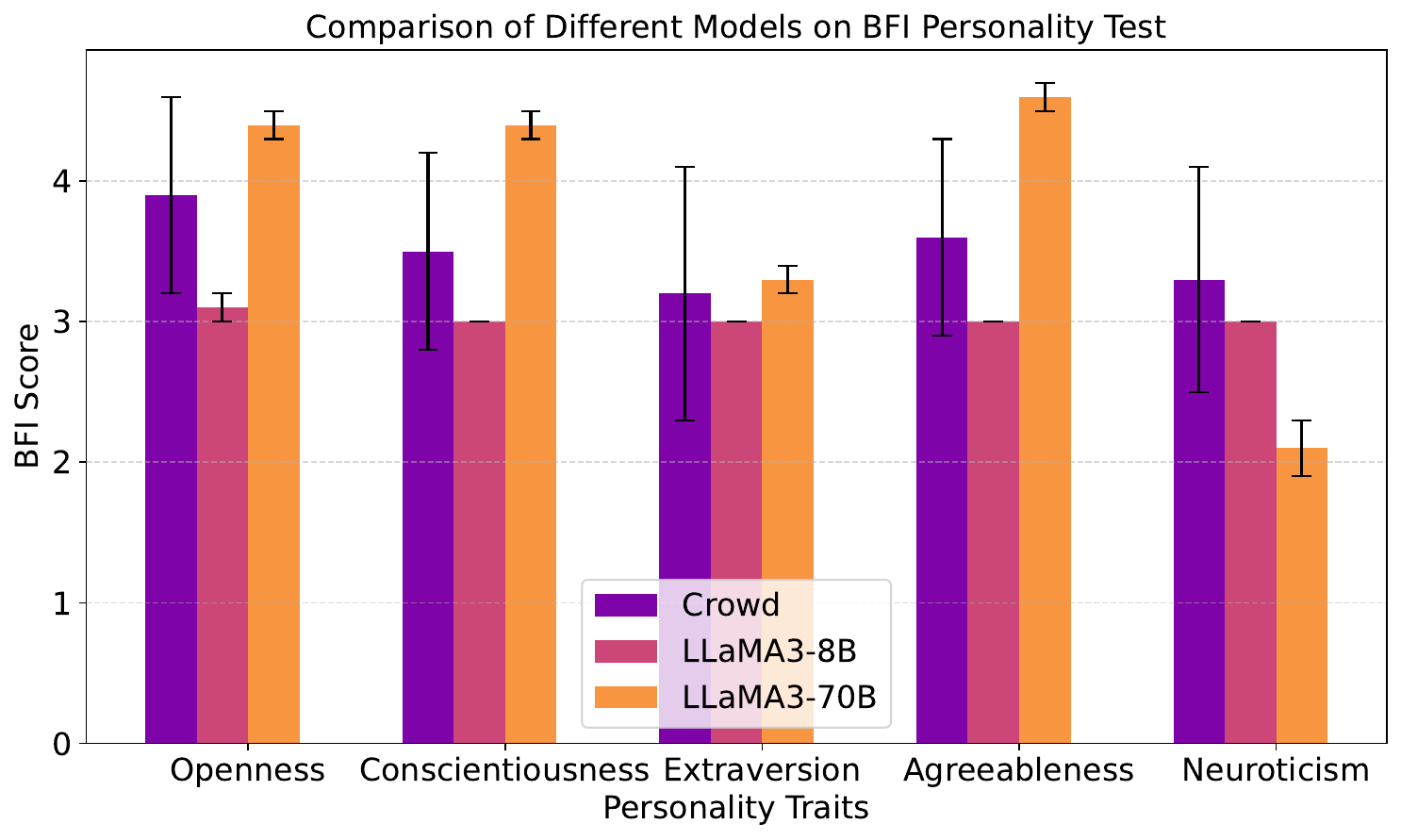}
    \end{minipage}
    \hfill
    \begin{minipage}{0.49\textwidth}
    \includegraphics[width=\linewidth]{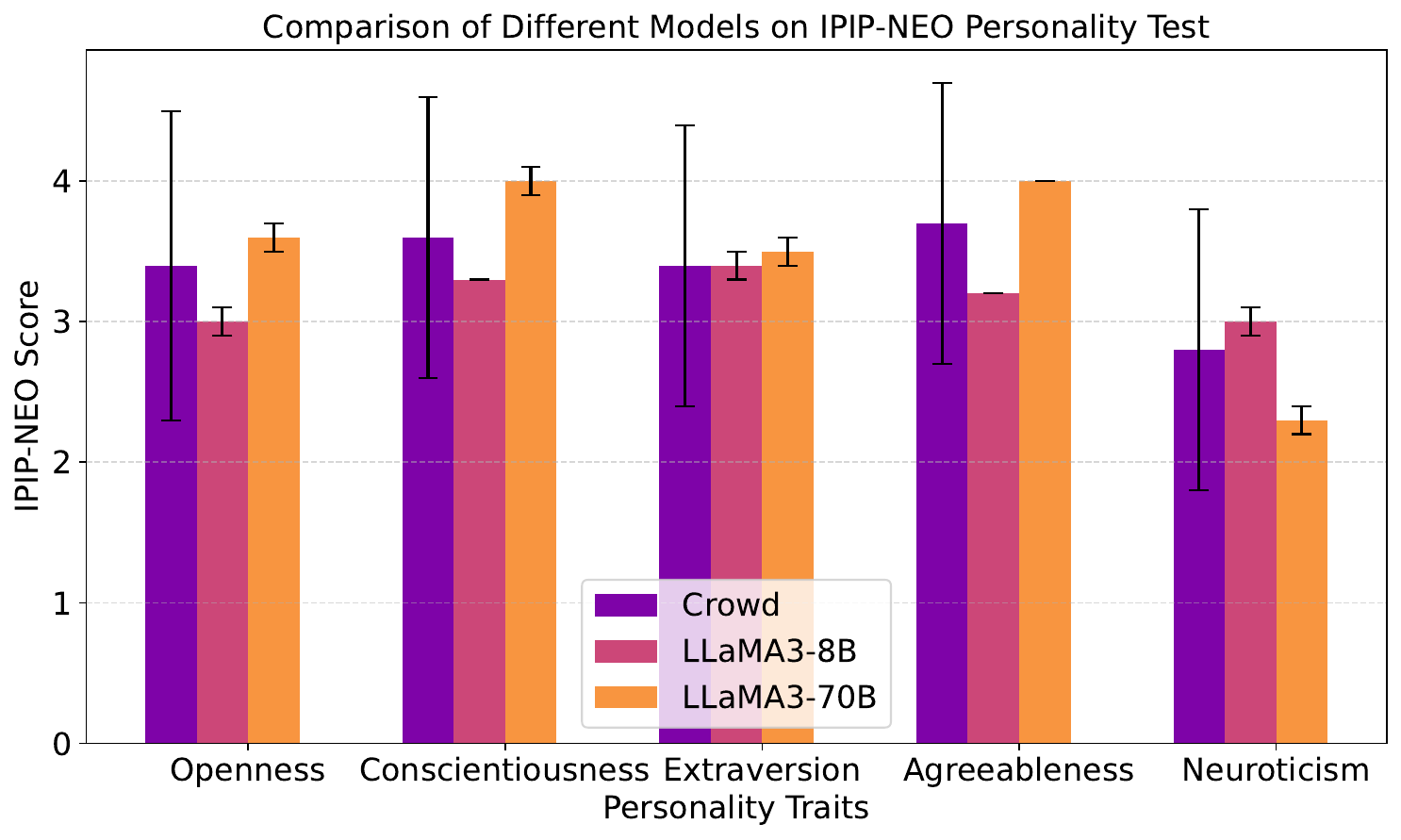}
    \end{minipage}
    \caption{The personality test results for the crowd and the \texttt{LLaMA-3-Instruct} models were obtained using zero-shot inference without explicitly inducing personality traits. The BFI test scores are displayed on the left. The IPIP-NEO test scores are displayed on the right.}
    \label{fig:bfi_ipip_llama3}
\end{figure}

\cref{fig:bfi_ipip_llama3} presents the BFI and IPIP-NEO test score results for the LLaMA-3 Instruct models, evaluated in zero-shot inference without any induced personality traits. The crowd-sourced response scores for the BFI test are sourced from \citet{psychobench}, and those for the IPIP-NEO test are drawn from \citet{jiang2023evaluating}. The results indicate that the scores for both \texttt{LLaMA-3-8B-Instruct} and \texttt{LLaMA-3-70B-Instruct} fall within the standard deviation of the human distribution. However, while \texttt{LLaMA-3-8B-Instruct} tends to generate more neutral scores (around 3 across most of the Big Five traits), \texttt{LLaMA-3-70B-Instruct} exhibits higher scores for openness, conscientiousness, extraversion, and agreeableness, and lower scores for neuroticism.

\subsection{Evaluating Finetune Models Psycholinguistic Richness in Unseen SODA Scenarios}
\label{appn:unseen_soda}

\paragraph{Description:} 
To address the concern that questionnaire-based evaluations may overlook the psycholinguistic richness essential for authentic human behavior, we designed this experiment to assess whether our models can generate linguistically nuanced outputs. Specifically, models—whether fine-tuned or prompted—are tasked with generating responses to unseen scenarios from the SODA dataset. We then employ our trained RoBERTa classifier to determine if the generated responses effectively reflect the desired personality traits.

\paragraph{Results and Analysis:} This result (see \cref{fig:soda_finetune_prompt}) demonstrates that DPO has a statistically significant advantage in capturing psycholinguistic richness. The performance of SFT and prompt-based approaches appears to be similar. This suggests that while both SFT and prompting can encode personality traits to some extent, they may lack the nuanced psycholinguistic adaptation achieved through preference optimization in DPO. Our findings suggest that training-based approaches, particularly DPO, are more effective in capturing the nuanced psycholinguistic richness required for authentic personality expression. Compared to prompt-based methods, which rely on external conditioning without modifying the underlying model parameters, training-based models can internalize personality traits more robustly, leading to more consistent and contextually appropriate generations. Furthermore, the limitations observed in SFT indicate that conventional supervised fine-tuning alone may not be sufficient for fully encoding the complexity of psycholinguistic adaptation. This suggests that while SFT can guide model behavior to some extent, it may lack the reinforcement-driven refinement necessary to achieve deeper alignment with personality traits.

\subsection{Intra-Trait Correlations in Personality Assessment}
\label{subsec:intra_trait_correlations}

\begin{figure*}[t]
    \centering
    \begin{minipage}{0.24\textwidth}
    \includegraphics[width=\linewidth]{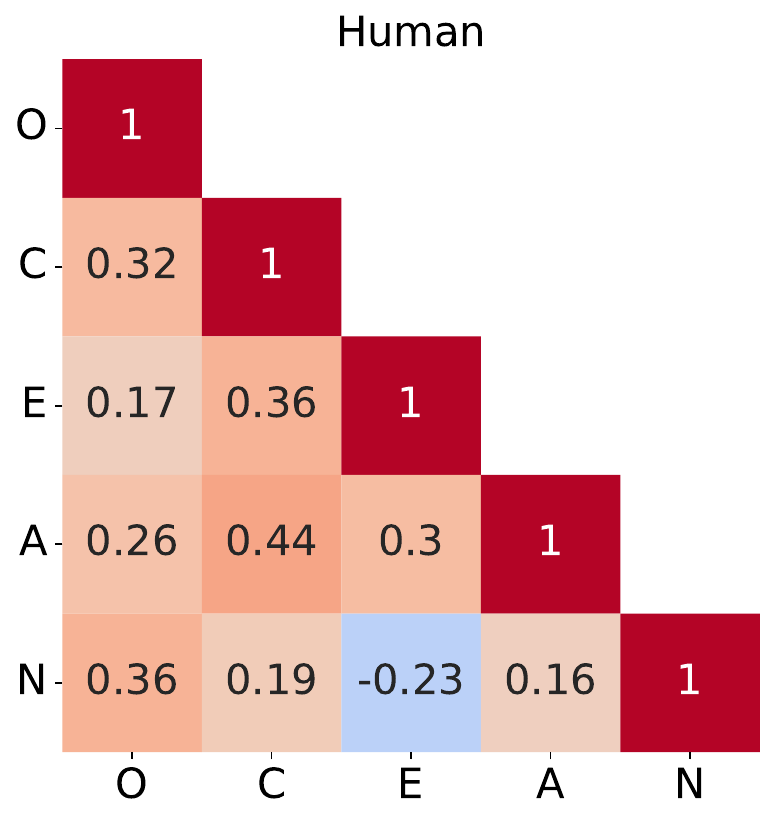}
    \end{minipage}
    \hfill
    \begin{minipage}{0.24\textwidth}
    \includegraphics[width=\linewidth]{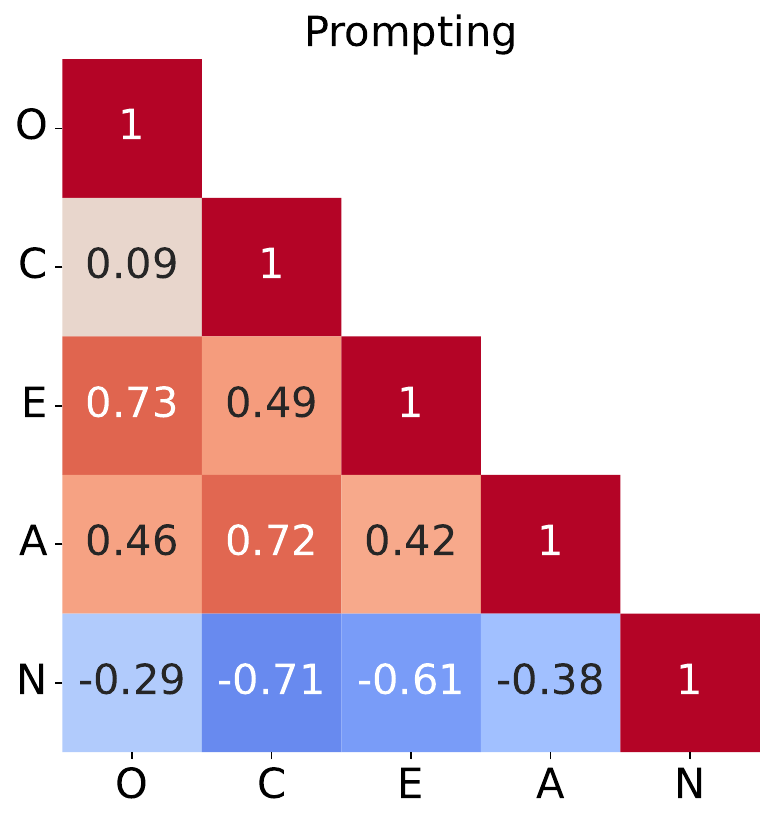}
    \end{minipage}
    \hfill
    \begin{minipage}{0.24\textwidth}
    \includegraphics[width=\linewidth]{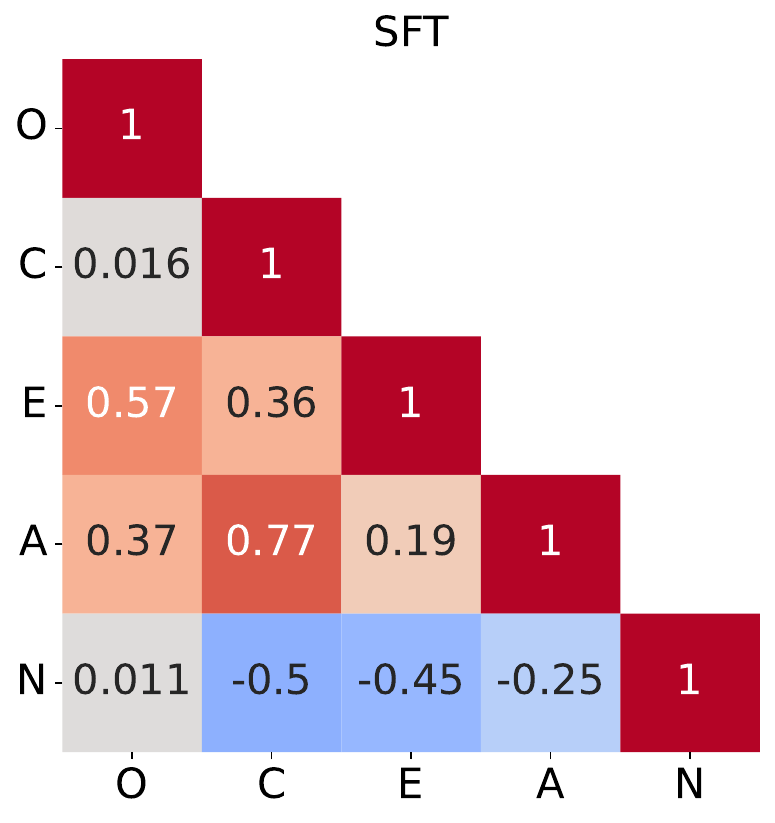}
    \end{minipage}
    \hfill
    \begin{minipage}{0.24\textwidth}
    \includegraphics[width=\linewidth]{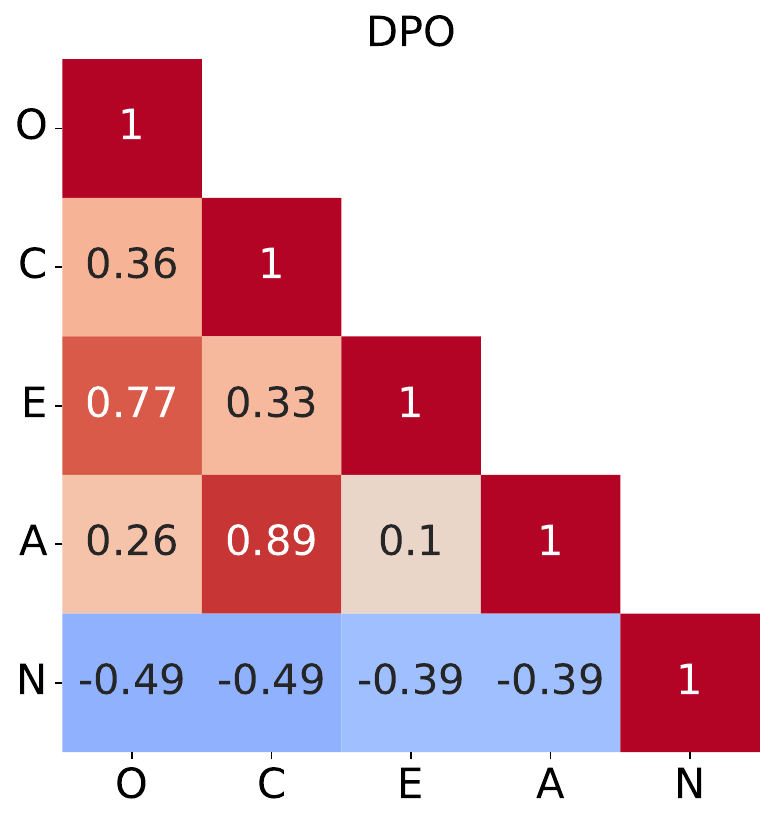}
    \end{minipage}
    \caption{Intra-trait Pearson correlations for human distributions on IPIP-NEO and the corresponding results from instruction-based prompting, SFT, and DPO. O represents openness, C conscientiousness, E extraversion, A agreeableness, and N neuroticism. The correlations especially for SFT align well with human distributions across openness, conscientiousness, extraversion, and agreeableness. Neuroticism shows less alignment with the other four traits compared to human distribution.
}
    \label{fig:intra_trait_corr}
\end{figure*}

To assess how well the prompting and training methods simulate intra-trait correlations observed in human data, we first calculated the intra-trait correlations from real human distributions using the IPIP-NEO questionnaire, based on the PAPI-120-600K dataset from \citet{zhu2024personalityalignmentlargelanguage}, which includes 619K human responses to the IPIP-NEO. Next, we computed the intra-trait correlations for the prompting, SFT, and DPO methods using the results from \cref{tab:personality_test_results}. These correlations are visualized in \cref{fig:intra_trait_corr}, showing that most traits are positively correlated, with the exception of neuroticism. To quantify the similarity between the method-generated and human correlation matrices, we calculated the matrix distance using the Frobenius norm, where 0 represents perfect similarity and 10 indicates maximum dissimilarity. The matrix distances were 2.10 for prompting, 1.55 for SFT, and 2.06 for DPO. These results suggest that the trained models, particularly SFT, more accurately capture the trait correlations seen in natural human data compared to the prompting-based methods.

\subsection{Reasoning Benchmark Results for \texttt{LLaMA-3-70B-Instruct}}
\label{subsec:reasoning_70b_full}

The complete results for the general reasoning tasks evaluated on the LLaMA-3-70B-Instruct model are presented in \cref{tab:full_benchmark_results_llama3_70b}. Note that the GPQA results in \cref{tab:benchmark_results_llama3_70b} were obtained using zero-shot prompting. This evaluation encompasses multiple reasoning domains and highlights the impact of different training methodologies: prompting, SFT, and DPO. These methods were assessed based on their ability to preserve the reasoning capabilities.

The results indicate that the SFT method consistently delivers the strongest performance across the benchmarks, outperforming both DPO and the prompting-based approach. For the 70B model, SFT emerges as the most effective method, achieving an optimal balance between incorporating personality traits and maintaining robust reasoning functionality. The aggregated results underscore the reliability of SFT, which demonstrates superior performance across diverse reasoning tasks, making it a robust choice for large-scale language models.

In contrast, the performance of the DPO method is more variable. While DPO excels in certain scenarios, such as low Neuroticism within the TruthfulQA task—where it achieves a notable score of 65.8\%—its overall results are less consistent across other reasoning benchmarks. Moreover, the final average scores reveal that high-trait DPO models underperform compared to their low-trait counterparts in general. This suggests a potential misalignment between DPO’s training objectives and the reasoning requirements of specific tasks. These findings highlight the nuanced trade-offs between training strategies, with SFT offering the most reliable approach for balancing personality trait integration and cognitive task performance in large-scale models.

\begin{table*}[t]
\centering
\resizebox{\textwidth}{!}{%
\begin{tabular}{llcccccccccccccc}
\toprule
\multirow{2}{*}{\textbf{Benchmark}} & \multirow{2}{*}{\textbf{Direct}} & \multirow{2}{*}{\textbf{Method}} & \multicolumn{2}{c}{\textbf{Openness}} & \multicolumn{2}{c}{\textbf{Conscientiousness}} & \multicolumn{2}{c}{\textbf{Extraversion}} & \multicolumn{2}{c}{\textbf{Agreeableness}} & \multicolumn{2}{c}{\textbf{Neuroticism}} & \multicolumn{2}{c}{\textbf{Average}} \\
 & & & \textbf{High} & \textbf{Low} & \textbf{High} & \textbf{Low} & \textbf{High} & \textbf{Low} & \textbf{High} & \textbf{Low} & \textbf{High} & \textbf{Low} & \textbf{High} & \textbf{Low} \\
\midrule

\multicolumn{15}{l}{\textit{\textbf{Hallucination Detection}}} \\
\specialrule{1.2pt}{0pt}{2.5pt}

\multirow{3}{*}{\textbf{TruthfulQA}}
 & \multirow{3}{*}{58.6 $\pm$ 1.7} & Prompt & 54.1 $\pm$ 1.6 & 51.1 $\pm$ 1.6 & 55.9 $\pm$ 1.7 & 45.2 $\pm$ 1.6 & 52.0 $\pm$ 1.6 & 55.7 $\pm$ 1.6 & 52.3 $\pm$ 1.7 & 49.1 $\pm$ 1.6 & 48.9 $\pm$ 1.6 & 58.6 $\pm$ 1.6 & 52.6 $\pm$ 1.6 & 51.9 $\pm$ 1.6 \\
 & & SFT & 55.2 $\pm$ 1.6 & 52.8 $\pm$ 1.6 & 55.6 $\pm$ 1.6 & 50.8 $\pm$ 1.5 & 54.5 $\pm$ 1.6 & 56.7 $\pm$ 1.6 & 54.4 $\pm$ 1.6 & 51.6 $\pm$ 1.6 & 52.4 $\pm$ 1.5 & 56.7 $\pm$ 1.6 & 54.4 $\pm$ 1.6 & 53.7 $\pm$ 1.6 \\
 & & DPO & 54.6 $\pm$ 1.6 & 54.2 $\pm$ 1.7 & 64.6 $\pm$ 1.6 & 38.5 $\pm$ 1.6 & 46.0 $\pm$ 1.7 & 65.3 $\pm$ 1.6 & 59.6 $\pm$ 1.6 & 50.6 $\pm$ 1.6 & 43.0 $\pm$ 1.7 & 65.8 $\pm$ 1.6 & 53.6 $\pm$ 1.6 & 54.9 $\pm$ 1.6 \\

\midrule

\multicolumn{15}{l}{\textit{\textbf{Social Reasoning}}} \\
\specialrule{1.2pt}{0pt}{2.5pt}

\multirow{3}{*}{\textbf{SocialIQA}}
 & \multirow{3}{*}{46.6 $\pm$ 1.1} & Prompt & 40.8 $\pm$ 1.1 & 43.9 $\pm$ 1.1 & 42.9 $\pm$ 1.1 & 39.9 $\pm$ 1.1 & 43.3 $\pm$ 1.1 & 42.0 $\pm$ 1.1 & 42.4 $\pm$ 1.1 & 40.8 $\pm$ 1.1 & 39.1 $\pm$ 1.1 & 44.1 $\pm$ 1.1 & 41.7 $\pm$ 1.1 & 42.1 $\pm$ 1.1 \\
 & & SFT & 50.3 $\pm$ 1.1 & 50.4 $\pm$ 1.1 & 50.9 $\pm$ 1.1 & 46.8 $\pm$ 1.1 & 50.0 $\pm$ 1.1 & 50.3 $\pm$ 1.1 & 50.5 $\pm$ 1.1 & 46.6 $\pm$ 1.1 & 48.2 $\pm$ 1.1 & 50.6 $\pm$ 1.1 & 50.0 $\pm$ 1.1 & 48.9 $\pm$ 1.1 \\
 & & DPO & 41.5 $\pm$ 1.1 & 44.5 $\pm$ 1.1 & 44.7 $\pm$ 1.1 & 37.6 $\pm$ 1.1 & 43.0 $\pm$ 1.1 & 43.6 $\pm$ 1.1 & 44.8 $\pm$ 1.1 & 39.0 $\pm$ 1.1 & 40.0 $\pm$ 1.1 & 45.3 $\pm$ 1.1 & 42.8 $\pm$ 1.1 & 42.0 $\pm$ 1.1 \\

\midrule
\multicolumn{15}{l}{\textit{\textbf{Commonsense Reasoning}}} \\
\specialrule{1.2pt}{0pt}{2.5pt}

\multirow{3}{*}{\textbf{CommonsenseQA}}
 & \multirow{3}{*}{27.0 $\pm$ 1.3} & Prompt & 60.0 $\pm$ 1.4 & 59.9 $\pm$ 1.4 & 22.5 $\pm$ 1.2 & 22.3 $\pm$ 1.2 & 35.5 $\pm$ 1.4 & 50.0 $\pm$ 1.4 & 45.0 $\pm$ 1.4 & 34.9 $\pm$ 1.4 & 20.2 $\pm$ 1.2 & 36.8 $\pm$ 1.4 & 36.6 $\pm$ 1.3 & 40.8 $\pm$ 1.4 \\
 & & SFT & 77.7 $\pm$ 1.2 & 78.8 $\pm$ 1.2 & 77.6 $\pm$ 1.2 & 66.0 $\pm$ 1.4 & 75.7 $\pm$ 1.2 & 78.9 $\pm$ 1.2 & 77.0 $\pm$ 1.2 & 73.8 $\pm$ 1.3 & 79.1 $\pm$ 1.2 & 78.5 $\pm$ 1.2 & 77.4 $\pm$ 1.2 & 75.2 $\pm$ 1.3 \\
 & & DPO & 57.7 $\pm$ 1.4 & 65.9 $\pm$ 1.4 & 23.8 $\pm$ 1.2 & 25.8 $\pm$ 1.3 & 23.2 $\pm$ 1.2 & 70.8 $\pm$ 1.3 & 21.3 $\pm$ 1.2 & 39.2 $\pm$ 1.4 & 20.1 $\pm$ 1.1 & 44.6 $\pm$ 1.4 & 29.2 $\pm$ 1.2 & 49.3 $\pm$ 1.4 \\

\midrule
\multirow{3}{*}{\textbf{PIQA}}
 & \multirow{3}{*}{80.4 $\pm$ 0.9} & Prompt & 79.6 $\pm$ 0.9 & 79.8 $\pm$ 0.9 & 80.5 $\pm$ 0.9 & 77.3 $\pm$ 1.0 & 78.0 $\pm$ 1.0 & 80.0 $\pm$ 0.9 & 79.8 $\pm$ 0.9 & 78.4 $\pm$ 1.0 & 78.8 $\pm$ 1.0 & 80.7 $\pm$ 0.9 & 79.3 $\pm$ 0.9 & 79.2 $\pm$ 0.9 \\
 & & SFT & 81.2 $\pm$ 0.9 & 81.0 $\pm$ 0.9 & 81.2 $\pm$ 0.9 & 80.4 $\pm$ 0.9 & 81.8 $\pm$ 0.9 & 81.3 $\pm$ 0.9 & 81.2 $\pm$ 0.9 & 80.0 $\pm$ 0.9 & 81.0 $\pm$ 0.9 & 81.2 $\pm$ 0.9 & 81.3 $\pm$ 0.9 & 80.8 $\pm$ 0.9 \\
 & & DPO & 76.4 $\pm$ 1.0 & 76.8 $\pm$ 1.0 & 79.4 $\pm$ 0.9 & 70.9 $\pm$ 1.1 & 76.4 $\pm$ 1.0 & 79.8 $\pm$ 0.9 & 78.5 $\pm$ 1.0 & 74.0 $\pm$ 1.0 & 72.9 $\pm$ 1.0 & 79.5 $\pm$ 0.9 & 76.7 $\pm$ 1.0 & 76.2 $\pm$ 1.0 \\

\midrule
\multicolumn{15}{l}{\textit{\textbf{Math Reasoning}}} \\
\specialrule{1.2pt}{0pt}{2.5pt}

\multirow{3}{*}{\textbf{GSM8K}}
 & \multirow{3}{*}{80.6 $\pm$ 1.1} & Prompt & 75.7 $\pm$ 1.2 & 70.1 $\pm$ 1.3 & 73.5 $\pm$ 1.2 & 32.6 $\pm$ 1.3 & 80.8 $\pm$ 1.1 & 33.5 $\pm$ 1.3 & 87.2 $\pm$ 0.9 & 77.8 $\pm$ 1.1 & 26.0 $\pm$ 1.2 & 89.4 $\pm$ 0.8 & 68.6 $\pm$ 1.1 & 60.7 $\pm$ 1.2 \\
 & & SFT & 85.8 $\pm$ 1.0 & 76.2 $\pm$ 1.2 & 86.4 $\pm$ 0.9 & 81.7 $\pm$ 1.1 & 85.1 $\pm$ 1.0 & 86.7 $\pm$ 0.9 & 87.0 $\pm$ 0.9 & 74.5 $\pm$ 1.2 & 76.0 $\pm$ 1.2 & 87.3 $\pm$ 0.9 & 84.1 $\pm$ 1.0 & 81.3 $\pm$ 1.1 \\
 & & DPO & 87.9 $\pm$ 0.9 & 88.5 $\pm$ 0.9 & 90.2 $\pm$ 0.8 & 80.6 $\pm$ 1.1 & 88.9 $\pm$ 0.9 & 90.4 $\pm$ 0.8 & 87.3 $\pm$ 0.9 & 90.0 $\pm$ 0.8 & 15.2 $\pm$ 1.0 & 91.0 $\pm$ 0.8 & 73.9 $\pm$ 0.9 & 88.1 $\pm$ 0.9 \\

\midrule
\multirow{3}{*}{\textbf{MathQA}}
 & \multirow{3}{*}{39.0 $\pm$ 0.9} & Prompt & 33.5 $\pm$ 0.9 & 33.5 $\pm$ 0.9 & 32.8 $\pm$ 0.9 & 31.5 $\pm$ 0.9 & 32.3 $\pm$ 0.9 & 33.3 $\pm$ 0.9 & 33.6 $\pm$ 0.9 & 32.4 $\pm$ 0.9 & 32.1 $\pm$ 0.9 & 34.1 $\pm$ 0.9 & 32.9 $\pm$ 0.9 & 33.0 $\pm$ 0.9 \\
 & & SFT & 43.3 $\pm$ 0.9 & 42.6 $\pm$ 0.9 & 43.0 $\pm$ 0.9 & 43.3 $\pm$ 0.9 & 43.2 $\pm$ 0.9 & 42.7 $\pm$ 0.9 & 42.9 $\pm$ 0.9 & 42.9 $\pm$ 0.9 & 42.8 $\pm$ 0.9 & 43.3 $\pm$ 0.9 & 43.0 $\pm$ 0.9 & 43.0 $\pm$ 0.9 \\
 & & DPO & 33.9 $\pm$ 0.9 & 34.7 $\pm$ 0.9 & 32.9 $\pm$ 0.9 & 28.1 $\pm$ 0.8 & 30.5 $\pm$ 0.8 & 35.0 $\pm$ 0.9 & 31.3 $\pm$ 0.8 & 32.8 $\pm$ 0.9 & 28.9 $\pm$ 0.8 & 34.0 $\pm$ 0.9 & 31.5 $\pm$ 0.8 & 32.9 $\pm$ 0.9 \\

\midrule
\multicolumn{15}{l}{\textit{\textbf{General Reasoning}}} \\
\specialrule{1.2pt}{0pt}{2.5pt}

\multirow{3}{*}{\textbf{MMLU}}
 & \multirow{3}{*}{74.5 $\pm$ 0.3} & Prompt & 70.3 $\pm$ 0.4 & 69.6 $\pm$ 0.4 & 40.6 $\pm$ 0.4 & 52.8 $\pm$ 0.4 & 56.9 $\pm$ 0.4 & 72.8 $\pm$ 0.4 & 69.0 $\pm$ 0.4 & 69.2 $\pm$ 0.4 & 55.3 $\pm$ 0.4 & 67.9 $\pm$ 0.4 & 58.4 $\pm$ 0.4 & 66.5 $\pm$ 0.4 \\
 & & SFT & 72.5 $\pm$ 0.4 & 72.0 $\pm$ 0.4 & 73.1 $\pm$ 0.4 & 68.6 $\pm$ 0.4 & 72.1 $\pm$ 0.4 & 73.5 $\pm$ 0.4 & 72.8 $\pm$ 0.4 & 70.7 $\pm$ 0.4 & 72.5 $\pm$ 0.4 & 73.8 $\pm$ 0.4 & 72.6 $\pm$ 0.4 & 71.7 $\pm$ 0.4 \\
 & & DPO & 57.9 $\pm$ 0.4 & 64.4 $\pm$ 0.4 & 50.3 $\pm$ 0.4 & 33.8 $\pm$ 0.4 & 42.3 $\pm$ 0.4 & 72.3 $\pm$ 0.4 & 34.3 $\pm$ 0.4 & 62.5 $\pm$ 0.4 & 33.2 $\pm$ 0.4 & 69.1 $\pm$ 0.4 & 43.6 $\pm$ 0.4 & 60.4 $\pm$ 0.4 \\
\midrule
\multirow{3}{*}{\textbf{GPQA \textsubscript{(0-shot)}}}
 & \multirow{3}{*}{33.5 $\pm$ 2.2} & Prompt & 31.5 $\pm$ 2.2 & 34.2 $\pm$ 2.2 & 31.7 $\pm$ 2.2 & 32.4 $\pm$ 2.2 & 34.6 $\pm$ 2.2 & 32.1 $\pm$ 2.2 & 32.4 $\pm$ 2.2 & 32.8 $\pm$ 2.2 & 31.9 $\pm$ 2.2 & 32.1 $\pm$ 2.2 & 32.4 $\pm$ 2.2 & 32.7 $\pm$ 2.2 \\
 & & SFT & 33.5 $\pm$ 2.2 & 32.4 $\pm$ 2.2 & 34.2 $\pm$ 2.2 & 34.2 $\pm$ 2.2 & 33.3 $\pm$ 2.2 & 34.4 $\pm$ 2.2 & 33.3 $\pm$ 2.2 & 33.3 $\pm$ 2.2 & 34.4 $\pm$ 2.2 & 33.5 $\pm$ 2.2 & 33.7 $\pm$ 2.2 & 33.6 $\pm$ 2.2 \\
 & & DPO & 36.8 $\pm$ 2.3 & 31.9 $\pm$ 2.2 & 35.7 $\pm$ 2.3 & 30.6 $\pm$ 2.2 & 35.9 $\pm$ 2.3 & 35.9 $\pm$ 2.3 & 35.5 $\pm$ 2.3 & 35.7 $\pm$ 2.3 & 32.6 $\pm$ 2.2 & 34.6 $\pm$ 2.2 & 35.3 $\pm$ 2.3 & 33.7 $\pm$ 2.2 \\
\midrule
\multirow{3}{*}{\textbf{GPQA \textsubscript{(5-shot)}}}
 & \multirow{3}{*}{36.6 $\pm$ 2.3} & Prompt & 35.9 $\pm$ 2.3 & 32.6 $\pm$ 2.2 & 36.2 $\pm$ 2.3 & 35.7 $\pm$ 2.3 & 36.2 $\pm$ 2.3 & 35.7 $\pm$ 2.3 & 34.4 $\pm$ 2.2 & 34.8 $\pm$ 2.3 & 36.6 $\pm$ 2.3 & 34.2 $\pm$ 2.2 & 35.9 $\pm$ 2.3 & 34.6 $\pm$ 2.3 \\
 & & SFT & 32.4 $\pm$ 2.2 & 32.8 $\pm$ 2.2 & 34.4 $\pm$ 2.2 & 33.7 $\pm$ 2.2 & 33.0 $\pm$ 2.2 & 33.9 $\pm$ 2.2 & 33.7 $\pm$ 2.2 & 32.8 $\pm$ 2.2 & 33.7 $\pm$ 2.2 & 34.8 $\pm$ 2.3 & 33.4 $\pm$ 2.2 & 33.6 $\pm$ 2.2 \\
 & & DPO & 37.5 $\pm$ 2.3 & 31.2 $\pm$ 2.2 & 35.9 $\pm$ 2.3 & 31.2 $\pm$ 2.2 & 37.1 $\pm$ 2.3 & 35.5 $\pm$ 2.3 & 33.5 $\pm$ 2.2 & 32.1 $\pm$ 2.2 & 36.6 $\pm$ 2.3 & 35.7 $\pm$ 2.3 & 36.1 $\pm$ 2.3 & 33.1 $\pm$ 2.2 \\

\midrule
\midrule

\multirow{3}{*}{\textbf{Average}}
 & \multirow{3}{*}{53.0 $\pm$ 1.3} & Prompt & 53.5 $\pm$ 1.3 & 52.7 $\pm$ 1.3 & 46.3 $\pm$ 1.3 & 41.1 $\pm$ 1.3 & 50.0 $\pm$ 1.3 & 48.3 $\pm$ 1.3 & 52.9 $\pm$ 1.3 & 50.0 $\pm$ 1.3 & 41.0 $\pm$ 1.3 & 53.1 $\pm$ 1.3 & 48.7 $\pm$ 1.3 & 49.1 $\pm$ 1.3\\
 & & SFT & 59.1 $\pm$ 1.3 & 57.7 $\pm$ 1.3 & 59.6 $\pm$ 1.3 & 56.2 $\pm$ 1.3 & 58.7 $\pm$ 1.3 & 59.8 $\pm$ 1.3 & 59.2 $\pm$ 1.3 & 56.2 $\pm$ 1.3 & 57.8 $\pm$ 1.3 & 60.0 $\pm$ 1.3 & 58.9 $\pm$ 1.3 & 58.0 $\pm$ 1.3 \\
 & & DPO & 53.8 $\pm$ 1.3 & 54.7 $\pm$ 1.3 & 50.8 $\pm$ 1.3 & 41.9 $\pm$ 1.3 & 47.0 $\pm$ 1.3 & 58.7 $\pm$ 1.3 & 47.3 $\pm$ 1.3 & 50.7 $\pm$ 1.3 & 35.8 $\pm$ 1.3 & 55.5 $\pm$ 1.3 & 47.0 $\pm$ 1.3 & 52.3 $\pm$ 1.3 \\

\bottomrule
\end{tabular}
}
\caption{Benchmark results for different personality traits on \texttt{LLaMA-3-70B-Instruct}. \textbf{Direct} refers to direct inference without including personality-related prompts. \textbf{Prompt} refers to instruction-based prompting. The table includes standard errors (shown as $\pm$ values) to provide statistical context for the results.}
\label{tab:full_benchmark_results_llama3_70b}
\end{table*}

\subsection{Reasoning Benchmark Results for \texttt{LLaMA-3-8B-Instruct}}
\label{subsec:reasoning_8b_full}

The reasoning evaluation results for the \texttt{LLaMA-3-8B-Instruct} model, assessed across six reasoning domains, are summarized in \cref{tab:llama3_8b_full_results}. Overall, the DPO method generally outperformed SFT and demonstrated performance comparable to the prompt-based approach. This indicates that, with the smaller 8B model, DPO effectively aligns personality traits without significantly compromising reasoning capabilities.

A comparison of personality trait levels revealed that models simulating high trait levels consistently outperformed their low-trait counterparts in both DPO and SFT settings. For instance, on the TruthfulQA benchmark, the high-conscientiousness DPO model achieved 55.0\%, significantly surpassing the low-conscientiousness model's 39.0\%. Similarly, on the GSM8K math reasoning task, the high-conscientiousness DPO model scored 72.2\%, substantially outperforming the low-level model.

On benchmarks such as TruthfulQA, GPQA (both zero-shot and five-shot), and MathQA, models trained using SFT and DPO performed comparably to the original unaligned model. This suggests that personality trait alignment does not adversely affect reasoning performance in these tasks for a small model. However, notable variations were observed in other benchmarks. For example, DPO exhibited significantly reduced performance on CommonsenseQA and MMLU compared to SFT, prompting, and the original model. Conversely, SFT underperformed on the GSM8K benchmark relative to DPO, prompting, and the original model. These results suggest that the DPO method may be more effective than SFT in preserving or enhancing reasoning performance for specific tasks and traits on small models, though the choice of alignment method may depend on the specific reasoning domain.

\begin{table*}[htbp]
\centering
\resizebox{\textwidth}{!}{%
\begin{tabular}{llccccccccccccc}
\toprule
\multirow{2}{*}{\textbf{Benchmark}} & \multirow{2}{*}{\textbf{Original}} & \multirow{2}{*}{\textbf{Method}} & \multicolumn{2}{c}{\textbf{Openness}} & \multicolumn{2}{c}{\textbf{Conscientiousness}} & \multicolumn{2}{c}{\textbf{Extraversion}} & \multicolumn{2}{c}{\textbf{Agreeableness}} & \multicolumn{2}{c}{\textbf{Neuroticism}} & \multicolumn{2}{c}{\textbf{Average}}\\
& & & High & Low & High & Low & High & Low & High & Low & High & Low & \textbf{High} & \textbf{Low}\\
\midrule

\multicolumn{15}{l}{\textit{\textbf{Hallucination Detection}}} \\
\specialrule{1.2pt}{0pt}{2.5pt}

\multirow{3}{*}{\textbf{TruthfulQA}} & \multirow{3}{*}{53.5} & Prompt & 49.0 & 51.5 & 50.6 & 44.4 & 45.3 & 51.9 & 49.2 & 50.3 & 54.6 & 45.2 & 49.7 & 48.7 \\
& & SFT & 50.0 & 45.7 & 50.9 & 43.8 & 46.2 & 52.0 & 49.9 & 46.3 & 53.6 & 42.9 & 50.1 & 46.1 \\
& & DPO & 52.4 & 49.1 & 55.0 & 39.0 & 35.0 & 59.2 & 52.8 & 45.5 & 58.2 & 38.8 & 50.7 & 46.3 \\
\midrule

\multicolumn{15}{l}{\textit{\textbf{Code Reasoning}}} \\
\specialrule{1.2pt}{0pt}{2.5pt}
\multirow[c]{3}{*}{\textbf{HumanEval}}
& \multirow[c]{3}{*}{60.4} & Prompt & 59.1 & 59.8 & 62.2 & 61.6 & 61.0 & 63.4 & 62.8 & 62.2 & 60.4 & 61.6 & 61.1 & 61.7  \\
& & SFT & 57.9 & 54.3 & 59.8 & 56.1 & 58.5 & 57.3 & 60.4 & 54.9 & 58.5 & 58.5 & 59.0 & 56.2 \\
& & DPO & 57.3 & 0.6 & 27.4 & 0.0 & 43.3 & 0.0 & 8.5 & 32.9 & 0.0 & 7.9 & 27.3 & 8.3 \\
\midrule
\multirow[c]{3}{*}{\textbf{MBPP}}
& \multirow[c]{3}{*}{54.6} & Prompt & 54.6 & 55.4 & 54.2 & 55.2 & 55.8 & 56.0 & 55.4 & 54.8 & 54.4 & 55.8 & 54.9 & 55.4\\
& & SFT & 56.2 & 56.2 & 54.2 & 56.2 & 56.4 & 56.4 & 55.6 & 55.8 & 55.0 & 56.4 & 55.5 & 56.2 \\
& & DPO & 53.6 & 47.6 & 53.0 & 35.2 & 54.6 & 51.4 & 54.4 & 53.8 & 52.0 & 54.2 & 42.9 & 48.4 \\
\midrule
\multicolumn{15}{l}{\textit{\textbf{Social Reasoning}}} \\
\specialrule{1.2pt}{0pt}{2.5pt}

\multirow{3}{*}{\textbf{SocialIQA}} & \multirow{3}{*}{49.7} & Prompt & 41.9 & 42.3 & 41.1 & 39.3 & 41.5 & 41.6 & 41.8 & 39.5 & 42.1 & 39.4 & 41.7 & 40.4 \\
& & SFT & 44.0 & 44.9 & 45.9 & 41.9 & 44.4 & 44.6 & 43.7 & 41.4 & 44.6 & 40.8 & 44.5 & 42.7 \\
& & DPO & 43.8 & 43.8 & 42.5 & 37.8 & 41.8 & 40.9 & 42.8 & 38.4 & 42.8 & 39.0 & 42.7 & 40.0 \\
\midrule

\multicolumn{15}{l}{\textit{\textbf{Commonsense Reasoning}}} \\
\specialrule{1.2pt}{0pt}{2.5pt}

\multirow{3}{*}{\textbf{CommonsenseQA}} & \multirow{3}{*}{51.8} & Prompt & 64.6 & 60.6 & 38.0 & 31.3 & 45.9 & 55.0 & 55.4 & 36.3 & 33.9 & 23.3 & 47.6 & 41.3 \\
& & SFT & 61.8 & 57.9 & 50.5 & 34.3 & 52.7 & 60.8 & 55.4 & 36.0 & 63.4 & 30.6 & 56.8 & 43.9 \\
& & DPO & 22.9 & 24.8 & 48.2 & 21.6 & 29.1 & 56.6 & 28.4 & 26.3 & 47.7 & 23.7 & 35.3 & 30.6 \\
\midrule

\multicolumn{15}{l}{\textit{\textbf{Math Reasoning}}} \\
\specialrule{1.2pt}{0pt}{2.5pt}

\multirow{3}{*}{\textbf{GSM8K}} & \multirow{3}{*}{64.7} & Prompt & 13.5 & 58.4 & 23.4 & 61.0 & 40.0 & 57.1 & 29.3 & 71.6 & 24.1 & 31.9 & 26.1 & 56.0 \\
& & SFT & 19.8 & 0.5 & 20.2 & 1.4 & 6.0 & 0.5 & 6.4 & 4.8 & 20.1 & 53.3 & 14.5 & 12.1 \\
& & DPO & 68.4 & 31.8 & 72.2 & 31.8 & 69.7 & 63.0 & 70.7 & 64.8 & 71.9 & 3.0 & 70.6 & 38.9 \\
\midrule

\multirow{3}{*}{\textbf{MathQA}} & \multirow{3}{*}{27.9} & Prompt & 27.6 & 28.3 & 27.9 & 27.3 & 27.1 & 27.8 & 27.2 & 28.1 & 28.1 & 25.9 & 27.6 & 27.5 \\
& & SFT & 30.1 & 30.2 & 29.6 & 30.3 & 31.0 & 30.6 & 29.6 & 30.3 & 29.6 & 29.4 & 30.0 & 30.2 \\
& & DPO & 26.9 & 27.8 & 28.3 & 25.1 & 25.8 & 27.6 & 24.9 & 27.7 & 29.7 & 24.9 & 27.1 & 26.6 \\
\midrule

\multicolumn{15}{l}{\textit{\textbf{General Knowledge}}} \\
\specialrule{1.2pt}{0pt}{2.5pt}

\multirow{3}{*}{\textbf{MMLU}} & \multirow{3}{*}{51.2} & Prompt & 37.5 & 29.1 & 23.2 & 27.0 & 24.7 & 29.2 & 27.7 & 25.5 & 23.4 & 23.8 & 27.3 & 26.9 \\
& & SFT & 45.0 & 48.5 & 35.6 & 32.0 & 37.5 & 46.5 & 44.2 & 39.9 & 47.1 & 31.7 & 41.9 & 39.7 \\
& & DPO & 23.0 & 29.8 & 29.7 & 26.9 & 24.8 & 41.4 & 30.7 & 26.3 & 30.8 & 23.1 & 27.8 & 29.5 \\
\midrule

\multirow{3}{*}{\textbf{GPQA \textsubscript{(0-shot)}}} & \multirow{3}{*}{28.1} & Prompt & 29.0 & 28.8 & 28.6 & 23.0 & 28.6 & 29.2 & 29.0 & 27.2 & 28.8 & 28.3 & 28.8 & 27.3 \\
& & SFT & 27.9 & 27.9 & 28.1 & 25.0 & 27.2 & 28.3 & 28.8 & 24.1 & 29.0 & 28.3 & 28.2 & 26.7 \\
& & DPO & 27.9 & 25.0 & 29.7 & 21.0 & 27.2 & 26.8 & 28.8 & 21.4 & 29.5 & 25.2 & 28.6 & 23.9 \\
\midrule

\multirow{3}{*}{\textbf{GPQA \textsubscript{(5-shot)}}} & \multirow{3}{*}{29.9} & Prompt & 29.7 & 26.6 & 28.8 & 26.8 & 28.3 & 26.6 & 27.9 & 28.6 & 29.0 & 25.2 & 28.7 & 26.8 \\
& & SFT & 26.1 & 27.0 & 28.8 & 26.6 & 28.8 & 28.6 & 30.6 & 27.9 & 28.6 & 27.5 & 28.6 & 27.5 \\
& & DPO & 27.9 & 26.3 & 28.3 & 23.0 & 26.8 & 28.1 & 27.5 & 24.6 & 28.8 & 25.2 & 27.9 & 25.4 \\
\midrule

\midrule

\multirow{3}{*}{\textbf{Average}} & \multirow{3}{*}{43.9} & Prompt & 35.8 & 40.5 & 31.5 & 34.4 & 34.3 & 39.5 & 35.1 & 38.2 & 31.7 & 29.1 & 33.7 & 36.4 \\
& & SFT & 37.2 & 34.0 & 34.8 & 27.6 & 32.8 & 35.3 & 35.0 & 29.9 & 38.8 & 34.8 & 35.7 & 32.3 \\
& & DPO & 35.6 & 30.7 & 41.6 & 26.9 & 34.1 & 43.2 & 37.7 & 33.8 & 42.4 & 23.4 & 38.3 & 31.6 \\
\bottomrule
\end{tabular}
}
\caption{Benchmark results for the \texttt{LLaMA-3-8B-Instruct} model are presented across various personality traits and evaluation methods. The benchmarks are categorized into six key areas: Hallucination Detection, General Reasoning, Social Reasoning, Commonsense Reasoning, Mathematical Reasoning, and General Knowledge.}
\label{tab:llama3_8b_full_results}
\end{table*}

\section{Correlation Between Personality Traits and Reasoning Behaviors}

\subsection{Human VS. \texttt{LLaMA-3-70B-Instruct}}
\label{sec:psychological_summary}

Understanding the influence of personality traits on reasoning behaviors in LLMs is crucial for developing models tailored to specific personality profiles. Research on the Big Five personality traits has consistently demonstrated their significant impact on human cognition and problem-solving abilities \citep{john1999big, soto2011age}. Traits such as openness, conscientiousness, and agreeableness are often associated with enhanced reasoning capabilities, while neuroticism has been found to impair performance across a range of reasoning tasks \citep{ackerman1997intelligence, schaie2004seattle, chamorro2006personality}.

\cref{tab:psychological_corr} summarizes relevant findings from recent psychological studies and their alignment with our experimental results on \texttt{LLaMA-3-70B-Instruct}. Our findings corroborate these studies, indicating that models exhibiting higher conscientiousness and agreeableness generally perform better in reasoning tasks. In contrast, models characterized by lower levels of extraversion and neuroticism also demonstrate improved reasoning performance. These results highlight the potential of personality-aligned training to optimize LLM performance for reasoning-intensive tasks.

\subsection{Human VS. \texttt{LLaMA-3-8B-Instruct}}
\label{subsec:reasoning_analysis}

The influence of Big Five Personality traits on reasoning tasks in human cognition, as outlined in \cref{tab:psychological_corr}, served as a foundation for analyzing the performance of the \texttt{LLaMA-3-8B-Instruct} model. This analysis aims to explore how alignment with different personality traits affects the model's reasoning capabilities. Below, we summarize the observed correlations between each trait and the model's performance across various reasoning benchmarks.

\textbf{Openness }
The impact of Openness on reasoning performance was highly task-dependent. Models aligned with high levels of Openness using the DPO method exhibited significantly improved performance in mathematical reasoning tasks. However, these models underperformed in commonsense reasoning benchmarks compared to both the prompt-based approach and the original model. These results suggest that while high Openness alignment enhances mathematical reasoning, it does not guarantee consistent improvements across all reasoning domains.

\textbf{Conscientiousness }
A strong positive correlation was observed between Conscientiousness and reasoning performance. Models aligned with higher levels of Conscientiousness consistently outperformed their low-level counterparts across most benchmarks. This trend highlights that high Conscientiousness alignment likely enhances systematic reasoning and attention to detail, benefiting performance across diverse reasoning tasks.

\textbf{Extraversion }
Lower levels of Extraversion were associated with better performance across reasoning tasks. Specifically, in commonsense reasoning benchmarks, models with low Extraversion significantly outperformed those with high Extraversion. This negative correlation suggests that high Extraversion may introduce distractibility, potentially impeding performance in tasks that require focused attention and analytical reasoning.

\textbf{Agreeableness }
The influence of Agreeableness on reasoning performance was minimal and inconsistent. No clear advantage was observed for models aligned with either high or low levels of Agreeableness across the benchmarks. These findings indicate that Agreeableness has a weak correlation with the model's reasoning capabilities, suggesting its alignment has little effect on overall performance.

\textbf{Neuroticism }
The relationship between Neuroticism and reasoning performance was inconsistent and did not align with expectations from human cognition studies. High Neuroticism models performed well in some reasoning tasks, while low Neuroticism models scored poorly in others. These results imply that high Neuroticism alignment does not necessarily impair reasoning performance, contrasting with psychological findings in humans. This discrepancy may arise from limitations in how Neuroticism is modeled and represented in the training process.

\begin{table*}[h]
\centering
{\fontsize{9.8}{11.76}\selectfont
\begin{tabularx}{\textwidth}{X}
\toprule
\multicolumn{1}{c}{\textbf{Example 1: High vs. Low Neuroticism}} \\
\midrule
\textbf{High Neuroticism Response:} “Oh no, oh no, oh no! This is a disaster! How could you be so careless?! My package was supposed to be delivered with care, not thrown around like trash! I need those items, and now they're all over the place! What am I supposed to do now?!” \\
\midrule
\textbf{Low Neuroticism Response:} “Thanks, Danilo! I'm fine, just a bit shaken. Let's take a look—I think my valuables are secure. I appreciate your quick thinking. Let’s take a deep breath and move on from this, okay?” \\
\midrule
\textbf{Hallucination and Bias Discussion:} The high-neuroticism response uses hyperbolic and repetitive language (“Oh no, oh no, oh no!”) to exaggerate a minor mishap, effectively “hallucinating” a catastrophic outcome. In contrast, the low-neuroticism response is measured and factual, reducing the risk of overdramatization. \\
\midrule
\multicolumn{1}{c}{\textbf{Example 2: Low vs. High Agreeableness}} \\
\midrule
\textbf{Low Agreeableness Response:} “Save the snacks for yourself, Wylder. I'm not some charity case that needs a 'break' from seriousness. I'm here to observe and learn, not indulge in petty pleasures. What's the point of this little outing, anyway? What do you want from me?” \\
\midrule
\textbf{High Agreeableness Response:} “Aww, thanks Wylder! You're too kind. I'm having a blast watching this little guy, and I’d love to get some snacks for the kids in the hospital too. Your thoughtfulness really makes my day!” \\
\midrule
\textbf{Hallucination and Bias Discussion:} The low-agreeableness response projects hidden, selfish motives by questioning the inviter’s intent, thereby biasing the interpretation of the social gesture. Conversely, the high-agreeableness response assumes positive intent, fostering goodwill and a constructive tone. \\
\midrule
\multicolumn{1}{c}{\textbf{Example 3: High Extraversion vs. Low Extraversion}} \\
\midrule
\textbf{High Extraversion Response:} “Wow, Wylder, this is the best day ever! I'm so stoked you brought me here. Let's grab some snacks and hang out with our new furry friend. I’m buying—you pick the treats! And after this, we should totally check out the rest of the zoo; it's gonna be a blast!”
\\
\midrule
\textbf{Low Extraversion Response:} “Thanks, Wylder. I'm enjoying the monkey, but I'm not really hungry. Can we just sit here and watch for a bit? I'm happy just observing.”
 \\
\midrule
\textbf{Hallucination and Bias Discussion:} The high-extraversion dialogue is marked by exuberance, with phrases like “this is the best day ever!” that can lead to a biased, overly optimistic portrayal of the situation. The speaker hallucinates an ideal, adventure-filled outing without acknowledging potential downsides or risks, thus skewing the listener's expectation toward perpetual fun. The low-extraversion response, which simply states enjoyment and a preference for observation, presents a more realistic and grounded perspective. \\
\midrule
\multicolumn{1}{c}{\textbf{Example 4: High Neuroticism vs. Low Neuroticism}} \\
\midrule
\textbf{High Neuroticism Response: } “Oh no, oh no, oh no! We must be in grave danger! I can feel something terrible is about to happen. What if we never get out? We need to keep moving, get as far away as possible—I'm terrified we’ll be trapped forever!” \\
\midrule
\textbf{High Agreeableness Response:} “Thank you, Ysabella. I appreciate your concern. I’ll rest for a bit, but I'm more focused on ensuring everyone else is safe. Let’s calmly assess our situation and plan our next steps together.” \\
\midrule
\textbf{Hallucination and Bias Discussion:} In the high-neuroticism response, the speaker hallucinates imminent danger, creating an exaggerated scenario that is not necessarily based on the facts of the situation. This introduces a bias toward catastrophic outcomes, which can lead to irrational decision-making. The low-neuroticism version is calm and methodical, focusing on collective safety and practical assessment. It avoids inventing extreme negative scenarios. \\
\midrule
\bottomrule
\end{tabularx}
}
\caption{Comparison of Responses and Bias Discussion}
\label{tab:combined_examples}
\end{table*}

\begin{table*}[h]
\centering
\begin{tabular}{{p{0.95\textwidth}}}
\toprule
\textbf{PsychoGenerator:} The new true blood book came out!!! need to finish these last two so i read it then its on to vampire diaries. \\
\midrule
\textbf{Big Five Essay:} Well, here we go with the stream of consciousness essay. I used to do things like this in high school sometimes. They were pretty interesting, but I often find myself with a lack of things to say. I normally consider myself someone who gets straight to the point. I wonder if I should hit enter any time to send this back to the front. Maybe I'll fix it later. My friend is playing guitar in my room now. Sort of playing anyway. More like messing with it. He's still learning. There's a drawing on the wall next to me. Comic book characters I think, but I'm not sure who they are. It's been a while since I've kept up with comic's. I just heard a sound from ICQ. That's a chat program on the internet. I don't know too much about it so I can't really explain too well. Anyway, I hope I'm done with this by the time another friend comes over. It will be nice to talk to her again. She went home this weekend for Labor Day. So did my brother. I didn't go. I'm not sure why. No reason to go, I guess. Hmm. when did I start this. Wow, that was a long line. I guess I won't change it later. Okay, I'm running out of things to talk about. I've found that happens to me a lot in conversation. Not a very interesting person, I guess. Well, I don't know. It's something I'm working on. I'm in a class now that might help. The phone just rang. Should I get it?  The guy playing the guitar answered it for me. It's for my roommate. My suitemate just came in and started reading this. I'm uncomfortable with that. He's in the bathroom now. You know, this is a really boring piece of literature. I never realized how dull most everyday thoughts are. Then again, when you keep your mind constantly moving like this, there isn't really time to stop and think deeply about things. I wonder how long this is going to be. I think it's been about ten minutes now. Only my second line. How sad. Well, not really considering how long these lines are. Anyway, I wonder what I'm going to do the rest of the night. I guess there's always homework to do. I guess we'll see. This seat is uncomfortable. My back sort of hurts. I think I'm going to have arthritis when I get older. I always thought that I wouldn't like to grow old. Not too old, I suppose. I've always been a very active person. I have a fear of growing old, I think. I guess it'll go away as I age gradually. I don't know how well I'd deal with paralysis from an accident though. As long as I have God and my friends around, I'll be okay though. I'm pretty thirsty right now. There isn't much to drink around my room. Ultimate Frisbee, I haven't played that all summer. Fun game, but tiring. I'm out of shape. I'd like to get in better shape, but I hate running. It's too dull for me. Hmmm. it's almost over now. Just a few more minutes. Let's see if I make it to the next line. Short reachable goals!  Whatever. Anyway, what else do I have to do tonight. I guess I could read some. My shirt smells like dinner. It's pretty disgusting. I need to wake up for a 9:30 am class tomorrow. I remember when that wasn't early at all. Well, I made it to the next line. I'm so proud of myself. That's sarcasm, by the way. I wonder if I was suppose to right this thing as a narrative. Oh well too late now. Time for me to head out. Until next time, good bye and good luck. I don't know. \\
\bottomrule
\end{tabular}
\caption{An example of a PsychoGenerator and Big Five Essay Dataset}
\label{tab:psychogenerator_bigfive_essay_example}
\end{table*}

\section{Bias \& Hallucination Discussion}
\label{appn:bias_hallucination}

Upon qualitative investigation, we observe that personality extremes (e.g. high neuroticism and low agreeableness) can induce hallucinations within LLM-generated dialogues. In this context, ``hallucination'' refers to the creation of unfounded details, such as overblown catastrophic outcomes or attributed malevolent intentions, that are not substantiated by the surrounding discourse. These fabrications introduce bias by distorting perceived reality, thereby amplifying potential issues or misrepresenting motives without adequate justification. Some picked examples are shown in \cref{tab:combined_examples}.

Conversely, dialogues characterized by personality traits, specifically, low neuroticism, high agreeableness, and moderate extraversion tend to maintain alignment with factual bases. These responses promote constructive dialogue through measured and factual communication, reducing the likelihood of escalating tensions or misinterpreting benign situations.

Other traits such as extraversion, conscientiousness, and openness, predominantly influence the tone and creativity of responses. For example, high extraversion can result in more energetic and optimistic exchanges, whereas high openness may lead to the generation of creative or unconventional ideas. However, unlike the extreme traits of neuroticism or agreeableness, these do not typically lead to the severe distortions or biases akin to hallucinations.

\section{Clarification of the Dataset Construction Pipeline}

\subsection{Why Is Multi-Step Training Needed?}
PsychGenerator consists of single-post monologues with minimal conversational structure, making it too dissimilar to our intended use cases. We trained a smaller “expert generator” model on PsychGenerator so it would learn to produce text reflecting distinct Big Five traits (high/low). Although our classifier evaluations verified that this generator indeed produces text strongly aligned with each personality trait, the outputs tended to be monotonous or incoherent across broader conversational contexts. Here are some example outputs of our generator:

\vspace{-0.5em}
\begin{itemize}
  \item \textit{do not pass go, do not collect \$200.}
  \item \textit{rusted root at the lilac festival}
  \item \textit{hold me now, oh hold me now}
  \item \textit{crap, thinks janey as she looks at the clock.}
  \item \textit{i like it..so far, i like it.}
\end{itemize}
\vspace{-0.5em}

Because purely monologue-based fine-tuning cannot capture dynamic social nuance, we use SODA scenarios to supply diverse prompts and social backdrops. Specifically, for each SODA scenario, the expert generator’s trait-specific outputs “steer” a large model, thus producing single-turn dialogue responses that align with each Big Five trait. Even though the final responses are single-turn, they are generated within varied, realistic conversation setups, thereby bringing contextual richness beyond typical monologue data.

By merging PsychGenerator’s personality signals and SODA’s scenario prompts, we create BIG5-CHAT—a large set of single-turn responses that reflect specific Big Five traits in many different conversation contexts. This ensures each response is consistent with the social situation described by SODA while maintaining the intended personality style captured by the smaller expert generator.

\subsection{Why Is Steering Needed?}

Steering is key to aligning the large backbone model with the distinct, high/low trait signals learned by the smaller generator. If we were to rely on a single generator or attempt to fine-tune a 70B model directly on PsychGenerator, we would risk monologue-style, repetitive outputs. By integrating the smaller, specialized expert’s logits with the large model’s generation process, steering leverages the backbone model’s linguistic consistency to avoid the repeated, single-theme outputs that often arise from a full fine-tune on PsychGenerator alone. In doing so, it cost-effectively “pulls” responses toward the desired personality attributes while preserving the large model’s broader language fluency. Additionally, steering enables us to generate curated single-turn dialogues for BIG5-CHAT, which we use to fine-tune the 70B model using LoRA, ensuring robust, context-responsive trait expressions without sacrificing coherence or adaptability.

\subsection{Dataset Construction Strategy}

\paragraph{Necessity of specific dataset construction approach}
Collecting human-grounded dialogues alone is insufficient for our research goal, which requires both validated personality traits and rich conversational context. PsychGenerator provides the former but lacks dialogue structure; SODA offers diverse scenarios but lacks personality annotations. By integrating the two-steering trait-specific language from PsychGenerator into SODA prompts, we generate single-turn dialogues that are both personality-expressive and contextually grounded. This hybrid approach ensures controlled, diverse, and realistic personality modeling, which purely human-curated or synthetic datasets alone cannot fully achieve.

\paragraph{Necessity of dataset construction}
\cref{tab:personality_test_results_full} demonstrates that our fine-tuned model, trained on our dataset, achieves more precise personality emulation than prompting alone when compared against state-of-the-art proprietary models (GPT-4o). Notably, the prompt used for prompting (Prompt-Inst) is identical to the instruction used during fine-tuning, in Prompt-Demo, each run involves randomly sampling 10 examples from the full BIG5-CHAT dataset.

\section{Deeper linguistic analysis}

To evaluate whether \bigfivechat{} induced personality traits align with human linguistic patterns, we conducted comparative term frequency analyses, statistical tests of trait correlations, and a detailed LIWC-based analysis, confirming strong validity and alignment with human psycholinguistic patterns to deepen the psycholinguistic insight.

Using LIWC2015, we examine the use of positive and negative emotion categories, normalized by token count, and compare these values with corresponding examples from the SODA dataset \citep{kim2022soda}. The results align closely with established psycholinguistic findings \citep{spitzley2022linguistic, mairesse2007using}: conscientiousness, extraversion, and agreeableness are positively associated with positive emotion (and negatively with negative emotion), whereas neuroticism shows the opposite pattern. These findings suggest that Big5-Chat effectively captures emotional variation associated with personality traits. Furthermore, in comparison with SODA, we observe significant distributional shifts across traits and levels, highlighting the effectiveness of personality steering. Welch's t-test reveals statistically significant differences (p=0.05) between high and low trait levels for all the columns whose p-values are not explicitly reported in the \cref{tab:liwc_emotions_personality_big5chat_soda}.

\begin{table*}[htbp]
  \centering
  \resizebox{\textwidth}{!}{%
  \begin{tabular}{l r *{10}{r}}
    \toprule
    & \multirow{2}{*}{\textbf{SODA}}
    & \multicolumn{2}{c}{\textbf{Openness}}
    & \multicolumn{2}{c}{\textbf{Conscientiousness}}
    & \multicolumn{2}{c}{\textbf{Extraversion}}
    & \multicolumn{2}{c}{\textbf{Agreeableness}}
    & \multicolumn{2}{c}{\textbf{Neuroticism}} \\
    &  & \textbf{High} & \textbf{Low}
       & \textbf{High} & \textbf{Low}
       & \textbf{High} & \textbf{Low}
       & \textbf{High} & \textbf{Low}
       & \textbf{High} & \textbf{Low}\\
    \midrule
    Positive Emotions
        & 731.65
        & 657.37 & 547.63
        & 638.44 & 544.92
        & 729.29 & 697.73
        & 874.77 & 578.95
        & 421.76 & 818.35 \\[2pt]
    Negative Emotions
        & 217.31
        & 137.94 & 205.26
        & 142.36 & 214.56
        & 141.44 & 203.29
        & 167.29 & 219.12
        & 426.43 & 149.37 \\
    \bottomrule
  \end{tabular}}
  \caption{LIWC Emotion Word Frequencies: Comparison Between SODA Baseline and Big Five Personality Extremes (Unit: Word Frequency)}
  \label{tab:liwc_emotions_personality_big5chat_soda}
\end{table*}

We then assess whether more concrete categories emerge similar correlations in our dataset. In the \cref{tab:liwc_big5_all}, ``\textbf{(+)}'' indicates a positive correlation and ``\textbf{(--)}'' indicates a negative one, as reported across several psycholinguistic studies\,\citep{spitzley2022linguistic,mairesse2007using,jiang2023personallm}.  
Taken together, the results demonstrate a strong linguistic alignment between Big5-Chat and patterns of human personality expression.

\begin{table*}[htbp]
\centering
\scriptsize
\setlength{\tabcolsep}{4pt}
\begin{minipage}{\linewidth}\centering
\begin{tabular}{lrrr}
\toprule
\multicolumn{4}{c}{\textbf{Openness}}\\\midrule
               & insight (+) & friend (+) & assent (--)\,\\\midrule
High Openness  & 371.17 & 28.59 & 58.11\\
Low Openness   & 282.25 & 16.64 & 85.04\\
SODA           & 342.44 & 24.07 & 156.49\\\bottomrule
\end{tabular}
\end{minipage}\par\vspace{4pt}

\begin{minipage}{\linewidth}\centering
\begin{tabular}{lrrrrr}
\toprule
\multicolumn{6}{c}{\textbf{Conscientiousness}}\\\midrule
                          & certain (+) & discrep (+) & body (+) &
                          function (+) & ingest (--)\,\\\midrule
High Conscientiousness    & 190.91 & 197.97 & 23.78 & 5191.21 & 36.64\\
Low Conscientiousness     & 161.51 & 156.64 & 29.00 & 5221.24 & 89.83\\
SODA                      & 171.77 & 181.49 & 18.22 & 5466.15 & 34.39\\\bottomrule
\end{tabular}
\end{minipage}\par\vspace{4pt}

\resizebox{\linewidth}{!}{%
\begin{tabular}{lrrrrrrrrrrrrr}
\toprule
\multicolumn{14}{c}{\textbf{Extraversion}}\\\midrule
                    & posemo (+) & affiliation (+) & affect (+) & money (+) &
                    social (+) & friend (+) & percept (+) & see (+) & feel (+) &
                    reward (+) & space (+) & home (+) & relig (+)\\\midrule
High Extraversion   & 729.29 & 497.57 & 881.44 & 34.02 & 1200.29 & 56.89 &
                      262.45 & 75.69 & 91.45 & 269.90 & 559.03 & 19.13 & 7.10\\
Low Extraversion    & 697.73 & 249.09 & 923.52 & 22.91 &  940.51 & 14.87 &
                      239.79 & 80.90 & 72.04 & 183.57 & 450.65 & 18.84 & 2.62\\
SODA                & 731.65 & 236.41 & 962.37 & 29.91 & 1128.54 & 24.07 &
                      234.58 & 80.20 & 81.66 & 223.35 & 401.08 & 23.33 & 3.72\\\bottomrule
\end{tabular}}%
\par\vspace{4pt}

\begin{minipage}{\linewidth}\centering
\begin{tabular}{lrrrrrrr}
\toprule
\multicolumn{8}{c}{\textbf{Agreeableness}}\\\midrule
                     & affiliation (+) & social (+) & death (--)\, &
                       female (--)\, & she/he (--)\, & we (--)\,
                       & negemo (--)\,\\\midrule
High Agreeableness   & 476.47 & 1355.80 & 1.54 & 17.95 & 17.88 & 184.11 & 167.29\\
Low Agreeableness    & 142.94 & 1306.10 & 3.93 & 14.52 &  8.70 &  32.19 & 219.12\\
SODA                 & 236.41 & 1128.54 & 3.88 & 24.12 & 28.37 &  77.70 & 217.31\\\bottomrule
\end{tabular}
\end{minipage}\par\vspace{4pt}

\resizebox{\linewidth}{!}{%
\begin{tabular}{lrrrrrrrrr}
\toprule
\multicolumn{10}{c}{\textbf{Neuroticism}}\\\midrule
                    & risk (+) & affiliation (+) & drives (+) & sad (+) &
                      anger (+) & feel (+) & negemo (+) & money (--)\,
                      & cogproc (--)\,\\\midrule
High Neuroticism    & 118.41 & 168.54 & 714.19 & 122.63 & 46.63 &
                      108.38 & 426.43 & 25.08 & 1526.10\\
Low Neuroticism     &  53.67 & 381.35 & 999.21 &  35.87 & 28.92 &
                       79.99 & 149.37 & 31.25 & 1271.52\\
SODA                &  77.89 & 236.41 & 732.50 &  70.82 & 28.81 &
                       81.66 & 217.31 & 29.91 & 1280.99\\\bottomrule
\end{tabular}}%
\caption{LIWC category frequencies for SODA baseline and Big‑Five personality.
\label{tab:liwc_big5_all}
“\textbf{(+)}” denotes a positive correlation, “\textbf{(--)}” a negative correlation.}
\end{table*}